\PassOptionsToPackage{table}{xcolor}
\documentclass[runningheads]{llncs}
\usepackage{pdflscape}
\usepackage{graphicx}
\usepackage{colortbl}
\usepackage{multicol}
\usepackage{colortbl}
\usepackage{xspace}
\usepackage{hyperref}
\usepackage{enumitem,kantlipsum}
\usepackage{amsmath}
\usepackage{pifont}
\usepackage{lscape}
\usepackage{rotating}
\usepackage{lipsum} 
\usepackage{amssymb}
\usepackage{amsfonts}
\usepackage{amstext}
\usepackage{tikz}
\usepackage{booktabs}
\usepackage{multirow}
\usepackage{array}
\usepackage{hyperref}
\usepackage{tcolorbox}
\usepackage{etoolbox}

\newcommand{\CC}{\cellcolor{gray!15}}
\newcommand{\cmark}{\raisebox{-0.1em}[0pt][0pt]{\ding{51}}}
\newcommand{\xmark}{\raisebox{-0.1em}[0pt][0pt]{\ding{55}}}
\newcommand{\pmark}{\raisebox{-0.1em}[0pt][0pt]{\ding{70}}}

\newcommand{\yes}{\cellcolor{yescolor!30}\cmark}
\newcommand{\no}{\cellcolor{nocolor!30}\xmark}
\newcommand{\partially}{\cellcolor{partialcolor!30}\pmark}

\newcommand{\STAB}[2]{\begin{tabular}{@{}c@{}}#1\end{tabular}}
\newcommand{\st}{\textsc{st}\xspace}
\newcommand{\lt}{\textsc{lt}\xspace}
\newcommand{\ltplus}{\textsc{lt}$^{\raisebox{.4\height}{\scalebox{.6}{+}}}$\xspace}

\newlength{\maxlen}

\definecolor{yescolor}{cmyk}{0.66, 0, 0.35, 0.09}
\definecolor{nocolor}{cmyk}{0,0.87,0.68,0.32}
\definecolor{lowregimecolor}{cmyk}{1,0,0.07,0.6}
\definecolor{highregimecolor}{cmyk}{1,0.25,0,0.13}
\definecolor{partialcolor}{rgb}{0.302,0.302,0.302}
\definecolor{questionmarkcolor}{cmyk}{0.1,0.5,0.4,0.3}

\usepackage{glossaries}
\setacronymstyle{long-short}
\newacronym{nlp}{NLP}{natural language processing}
\newacronym{sp}{SP}{sound processing}
\newacronym{nn}{NNs}{neural networks}
\newacronym{dl}{DL}{deep learning}
\newacronym{dnns}{DNNs}{deep neural networks}
\newacronym{ml}{ML}{machine learning}
\newacronym{tl}{TL}{transfer learning}
\newacronym{ws}{WS}{weak supervision}
\newacronym{al}{AL}{active learning}
\newacronym{dal}{DAL}{deep active learning}
\newacronym{mlp}{MLP}{multi-layer perceptron}
\newacronym{dnn}{DNN}{deep neural network}
\newacronym{sota}{SOTA}{state-of-the-art}
\newacronym{cv}{CV}{computer vision}
\newacronym{plm}{PLMs}{pre-trained language models}

\patchcmd{\paragraph}{\itshape}{\bfseries\boldmath}{}{}

\begin{document}

\title{ActiveGLAE: A Benchmark for Deep Active Learning with Transformers}
\titlerunning{Active General Language Adaption Evaluation}

\author{Lukas Rauch\inst{1} \and Matthias Aßenmacher \inst{2}\inst{3} \and  Denis Huseljic\inst{1} \and Moritz Wirth\inst{1} \and Bernd Bischl \inst{2}\inst{3} \and Bernhard Sick\inst{1}}

\authorrunning{L. Rauch et al.}

\institute{University of Kassel, Wilhelmshöher Allee 73, 34121 Kassel, Germany \\
\email{\{lukas.rauch, dhuseljic, moritz.wirth, bsick\}@uni-kassel.de}
\and
Department of Statistics, LMU Munich, Ludwigstr. 33, D-80539 Munich, Germany
\and 
Munich Center for Machine Learning (MCML), LMU Munich, Germany
\email{\{matthias, bernd.bischl\}@stat.uni-muenchen.de}}

\maketitle              

\begin{abstract}
Deep active learning (DAL) seeks to reduce annotation costs by enabling the model to actively query instance annotations from which it expects to learn the most. Despite extensive research, there is currently no standardized evaluation protocol for transformer-based language models in the field of DAL. Diverse experimental settings lead to difficulties in comparing research and deriving recommendations for practitioners. To tackle this challenge, we propose the \textsc{ActiveGLAE} benchmark, a comprehensive collection of data sets and evaluation guidelines for assessing DAL. Our benchmark aims to facilitate and streamline the evaluation process of novel DAL strategies. Additionally, we provide an extensive overview of current practice in DAL with transformer-based language models. We identify three key challenges - data set selection, model training, and DAL settings - that pose difficulties in comparing query strategies. We establish baseline results through an extensive set of experiments as a reference point for evaluating future work. Based on our findings, we provide guidelines for researchers and practitioners.

\end{abstract}
\section{Introduction}
\label{intro}
Transformer-based \gls*{plm} have exhibited \gls*{sota} performance in various \gls*{nlp} applications, including supervised fine-tuning \cite{devlin-etal-2019-bert} and few-shot learning \cite{brown2020language,chatgpt}. The commonality of these \gls*{dnns} is their ability of general language understanding acquired through self-supervised pre-training \cite{gpt2,raffel2020exploring,gao2020pile}. While pre-training reduces the need for annotated data for a downstream task, obtaining annotations (e.g., class labels) from humans is still time-intensive and costly in practice \cite{rauch2022,ein-dor2020}. Additionally, real-world applications require reliable models that can quickly adapt to new data and learn efficiently with few annotated instances \cite{tran2022}. Active Learning aims to minimize annotation cost by allowing the model to query annotations for instances which it expects to yield the highest performance gains \cite{settles2010,herde2021}.
\begin{figure*}[!tb]
\centering
\includegraphics[width=0.89\columnwidth]{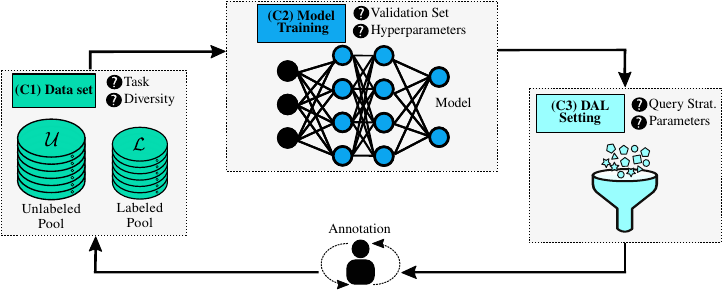} 
\caption{DAL cycle with three essential challenges that influence the evaluation protocol.}
\label{fig:abstract}
\end{figure*}
\setlength{\textfloatsep}{14pt}
 However, in the context of deep learning, evaluating \gls*{dal} is challenging due to several reasons \cite{ji2023,ren2020}. First, to ensure the practical applicability of query strategies, it is essential to have a wide range of diverse data sets (i.e., tasks). Second, the iterative fine-tuning of \gls*{plm} in each cycle iteration with multiple influential factors (i.e., model hyperparameters and \gls*{dal} settings) results in substantial runtime overhead. Third, in a realistic scenario, \gls*{dal} poses a one-time learning problem with no validation set for hyperparameter optimization, requiring careful consideration during evaluation \cite{kottke2017}. These challenges lead to researchers simplifying their experimental design, compromising the comparability of results and practical recommendations \cite{ji2023}. Thus, the benefits of applying \gls*{dal} in a realistic scenario are still ambiguous \cite{lüth2023}. For example, it remains unclear whether employing \gls*{dal} query strategies yields any benefits compared to randomly querying instances. Despite current efforts to enhance comparability of \gls*{dal} research in the vision domain \cite{lüth2023,ji2023,li2022,lang2022,beck2021}, a standardized evaluation protocol or a widely-accepted benchmark for \gls*{dal} in the \gls*{nlp} domain with \gls*{plm} is lacking.

To overcome these challenges, we propose the \textbf{Active} \textbf{G}eneral \textbf{L}anguage \textbf{A}daption \textbf{E}valuation (\textsc{ActiveGLAE}) benchmark. \textsc{ActiveGLAE} comprises a wide range of \gls*{nlp} classification tasks, along with guidelines for a realistic and comparative evaluation of \gls*{dal}. We aim to encourage research to employ a more comparable experimental design of \gls*{dal} with \gls*{plm}, hopefully enabling the identification of best practices for real-world scenarios. Following our guidelines, we provide an extensive experimental study, which yields baseline results for three \gls*{sota} \gls*{plm} utilizing various popular \gls*{dal} strategies. Our main \textbf{contributions} can be summarized as follows: 
\begin{enumerate}
    \item We highlight current practice and limitations in existing research in \gls*{dal} with transformer-based \gls*{plm} through an extensive literature analysis revealing three critical challenges for a comparative and realistic evaluation. More specifically, the evaluation of a \gls*{dal} process is heavily influenced by the selection of data sets (C1), the model training (C2), and the \gls*{dal} setting (C3). Figure~\ref{fig:abstract} illustrates these challenges along with their influential underlying factors.
    \item We propose the \textsc{ActiveGLAE} benchmark comprising ten \gls*{nlp} classification tasks that cover a diverse range of text genres, data set sizes, class cardinalities, and degrees of difficulty. Additionally, we provide guidelines to enable a standardized evaluation protocol. To streamline the evaluation process in \gls*{dal}, we will contribute the \textsc{ActiveGLAE} benchmark to Huggingface Datasets \cite{lhoest-etal-2021-datasets}.
    \item We conduct an extensive empirical study that provides baseline results for the \textsc{ActiveGLAE} benchmark. These serve as a reference for assessing novel \gls*{dal} processes and establish a minimum level of performance query strategies must exceed to be considered effective. Moreover, analyzing these baseline results and additional ablations regarding the identified challenges allows us to derive best practices for researchers and guidelines for practitioners.   
    \item We provide implementations\footnote{\href{https://github.com/dhuseljic/dal-toolbox/tree/main/experiments/aglae}{Github repository}} for all experiments to facilitate further research. The implementations are based on Huggingface Transformers \cite{wolf-etal-2020-transformers} to improve reusability. By providing each experiment's queried instances, we offer insights into the underlying \gls*{dal} process to ensure reproducibility. Additionally, all experimental results are publicly available\footnote{\href{https://wandb.ai/dal-nlp/active-glae}{Weights and Biases project}} through Weights and Biases \cite{wandb2020}.
\end{enumerate}


\section{Problem Setting}
\label{sec:problem}
We consider text classification problems where a $D$-dimensional instance is mapped to a feature vector $\mathbf{x}\in\mathcal{X}$ with the feature space $\mathcal{X} = \mathbb{R}^D$. An instance $\mathbf{x}$ is linked to a ground truth class label $y \in \mathcal{Y}$ with $\mathcal{Y}=\{1,...,C\}$  as the space of $C$ classes. We denote a model at cycle iteration $t$ through its parameters $\boldsymbol{\theta}_{t}$, equipped with a pre-trained encoder backbone and a sequence classification head. The model $f^{\boldsymbol{\theta}_{t}}: \mathcal{X} \to \mathbb{R}^C$ maps an instance $\mathbf{x}$ to a vector of class probabilities $\hat{\mathbf{p}} = f^{\boldsymbol{\theta}_{t}}(\textbf{x})$ corresponding to a prediction of the categorical class distribution. We investigate a pool-based \gls*{dal} scenario with an unlabeled pool data set ${\mathcal{U}(t) \subseteq \mathcal{X}}$ and a labeled pool data set ${\mathcal{L}(t) \subseteq \mathcal{X} \times \mathcal{Y}}$. We initialize the DAL process at $t=0$ with a randomly sampled set of annotated instances. At each cycle iteration $t$, the \gls*{dal} query strategy aggregates the most-useful instances in a batch ${\mathcal{B}(t) \subset \mathcal{U}(t)}$ with the size $b$. We denote an annotated batch as $\mathcal{B}^*(t) \in \mathcal{X} \times \mathcal{Y}$. We update the unlabeled pool $\mathcal{U}(t{+}1) = \mathcal{U}(t) \setminus \mathcal{B}(t)$ and the labeled pool $\mathcal{L}(t{+}1) = \mathcal{L}(t) \cup \mathcal{B}^*(t)$ with the annotated batch. At each cycle iteration $t$, the model $\boldsymbol{\theta}_{t}$ can be retrained from scratch (model cold-stat) or initialized with model parameters from the previous iteration (model warm-start). This leads to an update of the model parameters $\boldsymbol{\theta}_{t+1}$. The DAL process ends upon depletion of the budget $B$, representing the maximum number of queries.

\section{Current Practice of Evaluating Deep Active Learning}
\label{sec:current}
The current trend towards data-centric methods \cite{zha2023} and the adaptive capabilities of \gls*{dnns} \cite{tran2022} has led to numerous studies on \gls*{dal} with transformer-based \gls*{plm}. However, comparing \gls*{dal} results is a complex challenge \cite{munjal2022}, as indicated by varying evaluation protocols in current work. This lack of standardization makes it difficult to determine a given task's most effective query strategy. We identified three challenges that researchers need to address when designing a \gls*{dal} process (cf. Fig. \ref{fig:abstract}). In the following, we analyze the current practice concerning these challenges and briefly describe our suggested benchmark approach. While we focus on the \gls*{nlp} domain, we also include insights from \gls*{cv} where we observe a trend towards unified evaluation protocols \cite{ji2023,lang2022,lüth2023,li2022,munjal2022,beck2021}.

\subsection{(C1) Data Set Selection}
Developing robust \gls*{dal} processes that can be applied out-of-the-box (or at least task-specifically) is critical since they cannot be tested beforehand, and models only have one attempt to learn in a practical setting \cite{kottke2017}. Additionally, a lack of variation in the selection of data sets may lead to biased and non-generalizable results. Thus, to ensure generalizable results, a \gls*{dal} process needs to be evaluated on data sets covering a diverse range of text genres, pool sizes, class cardinalities, task difficulties, and label distributions. At the same time, a diverse data set selection across different publications leads to difficulties in comparing results. 

\begin{table*}[ht!]
\centering
    \caption{Data sets employed in current work with transformers in \gls*{dal}}
    \label{tab:datasets_related}
    \resizebox{\textwidth}{!}{%
    \setlength{\tabcolsep}{1.0pt}
    \begin{tabular}{l | c c c c c c c c c c c c c c c c c c}
        \toprule
                              & AGN  & B77 & DBP  & CR   & COLA & FNC1& IMDB& MNLI& MR  & PUBM& QNLI& QQP & SUBJ& SST2& SST5&TREC6& WIK & YELP5\\
        \midrule
        \cite{ein-dor2020}    & \yes & \no & \no  & \yes & \no  & \no & \no & \no & \yes& \no & \no & \no & \yes& \no & \no & \yes& \yes& \no \\ 
        \cite{lu2020}         & \yes & \no & \yes & \no  & \no  & \no & \no & \no & \yes& \no & \no & \no & \yes& \no & \no & \no & \no & \no \\  
        \cite{yuan2020}       & \yes & \no & \no  & \no  & \no  & \no & \yes& \no & \no & \no & \no & \no & \no & \yes& \no & \no & \no & \no \\
        \cite{ru2020}         & \yes & \no & \no  & \no  & \no  & \no & \no & \no & \no & \no & \no & \no & \no & \yes&\no  & \no & \no & \no \\
        \cite{prabhu2021}     & \yes & \no & \no  & \no  & \no  & \no & \no & \no & \no & \no & \no & \no & \no & \no & \no & \yes& \no & \no \\
        \cite{margatina2021a} & \yes & \no & \yes & \no  & \no  & \no & \yes& \no & \no & \no & \no & \no & \no & \yes& \no & \yes& \no & \no \\
        \cite{margatina2021}  & \yes & \no & \yes & \no  & \no  & \no & \yes& \no & \no & \yes& \yes& \yes& \no & \yes& \no & \no & \no & \no \\
        \cite{tan2021}        & \yes & \no & \no  & \no  & \no  & \no & \yes& \no & \no & \yes& \no & \no & \no & \no & \no & \no & \no & \no \\
        \cite{darcy2022}      & \yes & \no & \yes & \no  & \no  & \no & \no & \no & \yes& \no & \no & \no & \yes& \no & \no & \no & \no & \no \\
        \cite{schröder2022}   & \yes & \no & \no  & \yes & \no  & \no & \no & \no & \yes& \no & \no & \no & \yes& \no & \no & \yes& \no & \no \\
        \cite{gonsior2022}    & \yes & \no & \no  & \no  & \yes & \no & \yes& \no & \yes& \no & \no & \no & \yes& \yes& \no & \yes& \no & \no \\
        \cite{seo2022}        & \yes & \no & \no  & \no  & \no  & \no & \no & \no & \no & \yes& \no & \no & \no & \no & \no & \no & \no & \no \\
        \cite{yu2022a}        & \yes & \no & \yes & \no  & \no  & \no & \no & \no & \no & \yes& \no & \no & \no & \yes& \no & \yes& \no & \no \\
        \cite{zhang2022a}     & \yes & \no & \no  & \no  & \no  & \no & \yes& \no & \no & \no & \yes& \yes& \no & \no & \no & \yes& \no & \no \\ 
        \cite{jukić2022}      & \yes & \no & \no  & \no  & \no  & \no & \no & \no & \no & \no & \no & \no & \yes& \yes& \no & \yes& \no & \no \\
        \cite{kwak2022}       & \no  & \no & \no & \no  & \no   & \no & \no & \no & \yes& \no & \no & \no & \no & \yes& \no & \yes& \no & \no \\
        \cite{margatina2022}  & \yes & \no & \yes & \no  & \no  & \no & \yes& \no & \no & \no & \no & \no & \no & \yes& \no & \yes& \no & \no \\
        \cite{yu2022cold}     & \yes & \no & \yes & \no  & \no  & \no & \yes& \no & \no & \no & \no & \no & \no & \no & \no & \yes& \no & \no \\
        glae                  & \yes & \yes& \yes & \no  & \no  & \yes& \yes& \yes& \yes& \no & \no & \no & \no & \yes& \no & \yes& \yes& \yes\\
    \bottomrule
\end{tabular}
    }
\end{table*}
\paragraph{Current Practice.} There is an abundance of benchmark classification data sets in \gls*{nlp} leading to a disjoint selection between publications \cite{schröder2020}. Table \ref{tab:datasets_related} provides an overview of data sets employed in related work. While some specific data sets (e.g., AGN, SST2, TREC6) are used more frequently, the research landscape needs to be more cohesive. This lack of consensus reduces the comparability of results \cite{assenmacher2020comparability}. Current work reports diverse and contradictory findings across data sets, stressing the importance of further investigating the benefits of applying query strategies compared to randomly selecting instances in practice. Additionally, we found that related studies often employ data sets covering similar tasks, training data volumes, and task difficulties. We assume data sets are often bypassed due to a large unlabeled pool, which can result in high query times. 

\paragraph{Our study.} We propose the \textsc{ActiveGLAE} benchmark, a selection of data sets as a benchmark suite similar to (Super)GLUE \cite{wang2018glue,wang2020superglue}. We consider a wide variety of real-world tasks, aiming at spurring the general applicability of \gls*{dal}. This diverse set of tasks allows us to highlight task-specific challenges and derive evaluation guidelines. We intend to improve the generalizability and comparability of \gls*{dal} research.   

\subsection{(C2) Model Training}
\label{subsection:trainstrat}
Conventional hyperparameter-tuning cannot be performed when simulating a real-world setting in a \gls*{dal} process for several reasons \cite{kottke2017}: First, the existence of a validation data set contradicts the purpose of \gls*{dal} to reduce annotation effort (i.e., validation paradox \cite{lüth2023}). Second, model hyperparameters (e.g., number of learning steps and learning rate) can only be determined once at the beginning of the cycle. Third, the iterative nature of \gls*{dal} would necessitate hyperparameter optimization at each cycle iteration, leading to unwanted runtime overhead. Overcoming these challenges is crucial for successfully employing \gls*{dal} in practice \cite{munjal2022}. 

\paragraph{Current Practice.} Table \ref{tab:trainstrats} in Appendix \ref{appendix:related_work} depicts the adjustable model hyperparameters and the current practice in related work. In general, we identify three validation settings: (1) assuming the availability of all or parts of the validation data \cite{margatina2021,schröder2022}, (2) (dynamically) sample a validation set from the labeled pool \cite{tan2021}, or (3) omitting the validation set entirely \cite{lu2020,jukić2022}. All approaches with a validation set employ early stopping (or selecting the model with the best validation results) and consider the number of epochs as an optimizable hyperparameter. Ji et al.~\cite{ji2023} report that while early stopping can speed up the training process, it introduces randomness and decreases comparability. They recommend a fixed number of epochs suitable for the model architecture and data set. Recent studies suggest that model training is more important than the choice of a query strategy \cite{margatina2021a,jukić2022}, as fine-tuning \gls*{plm} on small datasets can suffer from training instability \cite{schick2021_exploit,schick2021_size,gao2021_few}. A real-world setting without a validation set is similar to few-shot learning, particularly at the beginning of a \gls*{dal} process with a small labeled pool. \cite{schick2021_exploit} and \cite{schick2021_size} omit the validation set entirely and deploy fixed hyperparameters in a practical few-shot learning setting. \cite{perez2021} find out that the presence of a validation set led to a significant overestimation of the few-shot ability of language models. In parallel, we consider this a major problem of deploying \gls*{dal} in practice. Jukić and Šnajder~\cite{jukić2022} adress this by introducing an early stopping technique utilizing the representation smoothness of \gls*{plm} layers from training. Training strategies usually follow the standard training procedure from \cite{devlin-etal-2019-bert}, employing a fixed number of training epochs (3 to 15), the AdamW \cite{loshchilov2017decoupled} optimizer with a learning rate between 2e-5 and 5e-5 and a learning rate scheduler with warmup (5-10\% of the steps). Researchers apply a fixed and well-established training strategy to a particular model to focus on the results of query strategies. However, the lack of established benchmarks in \gls*{dal} and partly incomplete hyperparameter specifications result in diverse training approaches (cf. Tab. \ref{tab:trainstrats}), that decrease comparability across publications.  

\paragraph{Our study.} Our benchmark study adopts a similar approach to previous work on few-shot learning \cite{schick2021_exploit,schick2021_size,perez2021}. We simulate a real-world \gls*{dal} process by entirely omitting the validation set without early stopping \cite{lüth2023}. We employ various fixed model hyperparameters (e.g., epochs and learning rates) that are dataset-agnostic with different \gls*{plm} as baselines to highlight their impact on resulting model performance. Therefore, our study provides reference points for further research by presenting results for various hyperparameter configurations. 

\subsection{(C3) Deep Active Learning Setting}
Evaluating \gls*{dal} requires determining a DAL setting which includes key factors such as the choice of query strategy, query size, and budget. These factors are often set as fixed without having established default values from comparable benchmark studies or meaningful baselines serving as a reference point. Query strategies can be sorted into uncertainty, diversity, and hybrid sampling. Uncertainty sampling identifies the most-uncertain instances in the hypothesis space, while diversity sampling focuses on diversity in the feature space \cite{yuan2020}. Hybrid approaches combine uncertainty and diversity sampling. 

\paragraph{Current Practice.} In Appendix~\ref{appendix:related_work}, Tab.~\ref{tab:dalregimes} presents an overview of \gls*{dal} settings and query strategies used in related work. Our benchmark study analyzes factors that influence the resulting model performance of query strategies in a \gls*{dal} process. These factors include:

\begin{itemize}[wide, labelindent=0pt]
\item{\textit{Initialization and update}: All prior studies use a model cold-start approach, where the model parameters are initialized from scratch at each cycle iteration (cf. Tab.~\ref{tab:dalregimes}). In contrast, a model warm-start initializes the model with parameters from the previous iteration. While a model warm-start may lead to faster convergence, it could likewise cause performance degradation due to potential bias towards the initial labeled pool \cite{hu2019,ash2020}. In contrast, Lang et al.~\cite{lang2022} and Ji et al.~\cite{ji2023} find that a model warm-start stabilizes the learning curve in \gls*{cv}. Uncertainty-based query strategies rely on the model's predictive uncertainty, which may require prior model training with task-specific information \cite{yuan2020,yu2022cold}. We refer to this as a data warm-start. Alternatively, if no initial labeled pool is available for model training, we refer to it as data cold-start. While related work commonly employs a data warm-start with a fixed initial pool size, Yuan et al.~\cite{yuan2020} and Yu et al.~\cite{yu2022cold} aim to leverage the pre-trained knowledge of a transformer-based \gls*{plm} focusing on data cold-start in a \gls*{dal} process.}

\item{\textit{Stopping criterion}: To ensure comparability, current studies usually use a fixed budget with a maximum limit of 2000 annotations as the stopping criterion. Alternatively, the budget may be data set-specific (e.g., 15\% of the available unlabeled pool) \cite{margatina2022,zhang2022a}. Tran et al.~\cite{tran2022} and Hacohen et al.~\cite{hacohen2022} determine the budget based on the complexity of the learning task (e.g., with the number of classes). Once the budget depletes, the \gls*{dal} process stops. Recent work in \gls*{cv} differentiates between budget sizes to simulate a diverse set of real-world scenarios. Hacohen et al. report that uncertainty-based strategies perform better with higher budgets. Note that \cite{hacohen2022} use the terms budget and warm-start interchangeably and refer to a high-data regime when a sizeable initial set is available (i.e., data warm-start) and vice versa. At the current state, there are no detailed investigations concerning the budget in \gls*{nlp} with transformer-based \gls*{plm}}.  
\item{\textit{Query size}: The query size is a crucial since it affects the number of cycle iterations given a specific annotation budget. For example, a larger query size leads to fewer model updates and may affect model performance and runtime. While Lüth et al.~\cite{lüth2023} and Lang et al.~\cite{lang2022} report better results with smaller query sizes in the \gls*{cv} domain, D'Arcy and Downey~\cite{darcy2022} demonstrate no difference between a query size of 12 and 25 with transformer-based \gls*{plm}. Current work mostly uses a fixed query size of 100 annotations or less \cite{ein-dor2020,yu2022cold}, with some studies also applying a relative size, such as 2\% of the unlabeled pool \cite{margatina2021a,margatina2022}.}

\item{\textit{Pool subset}: Since the size of the unlabeled pool significantly impacts query time, researchers often avoid large data sets or introduce an unlabeled subset from which they query annotations \cite{schröder2022}. For instance, \cite{margatina2021a} subsamples DBPedia \cite{lehmann2015} via stratified sampling once at the beginning of the \gls*{dal} process to maintain the initial label distribution. However, \cite{ji2023} suggest avoiding sub-sampling as it can potentially alter the ranking of \gls*{dal} query strategies.}
\end{itemize}
\vspace{-0.4cm}
\paragraph{Our study.} We present baseline results for low and high-budget settings alongside two query sizes on \textsc{ActiveGLAE}. We concentrate on a fixed number of initial instances (i.e., data warm-start) and use the term budget size independent of the initial pool. We contend that the number of classes cannot easily determine the task complexity since other factors like class imbalance or pool size play a crucial role. Therefore, to ensure comparability across experiments and draw conclusions on the influence of the data set complexity, we employ a fixed \gls*{dal} setting that is independent of the data set. We also explore the influence of model warm-start and model cold-start. While iteratively subsampling at each cycle iteration may introduce randomness \cite{ji2023}, it significantly reduces experiment runtime. Thus, we examine the effect of employing a pool subset on model performance across query strategies. 

\section{ActiveGLAE - Data Sets and Tasks}
\label{sec:activeglae}

We aim at creating a representative benchmark collection of real-world tasks enabling a standardized and realistic evaluation of \gls*{dal} strategies. Thus, we carefully select a multitude of data sets (cf. Tab. \ref{tab:overviewdatasets} for a comprehensive overview) exhibiting different characteristics relevant to practical applications of \gls*{dal}. Our selection covers balanced as well as imbalanced data sets and binary as well as multi-class settings. We choose class sets of low (3 to 6 classes), medium (14 classes), and high (77 classes) cardinality. Since class imbalance is a common problem in real-world scenarios, we consider naturally imbalanced data sets, rather than artificially introducing imbalance, for a more realistic representation. The data sets differ in size (10k - 650k examples) and encompass various classification tasks.

\begin{table}[ht]
\caption{Overview of \textsc{ActiveGLAE} data sets and tasks. When no test set is available (MNLI, QNLI, SST2), we use the validation set as a replacement. All data sets are available on Huggingface Datasets and respective links are added to the table.}
\label{tab:overviewdatasets}
\centering
\resizebox{\textwidth}{!}{%
\begin{tabular}{@{}m{5em}lll>{\centering\arraybackslash}m{3em}cl@{}}
\toprule
\textbf{Corpus}         & $\vert$\textbf{Train}$\vert$ & $\vert$\textbf{Test}$\vert$   & \textbf{Task}                      & \textbf{\#cls} & \textbf{Balanced}  & \textbf{Text Source}        \\ \midrule
\href{https://huggingface.co/datasets/ag_news}{AG's News}      & 120k                & 7600                 & news classification       & 4         & \yes      & news articles       \\
\href{https://huggingface.co/datasets/banking77}{Banking77}      & 10k                 & 3000                 & conversational language   & 77        & \yes  & banking intents     \\
\href{https://huggingface.co/datasets/dbpedia_14}{DBPedia}       & 560k                & 5000                 & ontology classification   & 14        & \yes      & wikipedia articles  \\
\href{https://huggingface.co/datasets/nid989/FNC-1}{FNC-1}          & 40k                 & 4998                 & stance detection          & 4         & \no       & news articles       \\
\href{https://huggingface.co/datasets/multi_nli}{MNLI}           & 390k                & 9815                 & textual entailment        & 3         & \yes      & miscellaneous        \\
\href{https://huggingface.co/datasets/glue/viewer/qnli/test}{QNLI}           & 104k                & 5463                 & question answering        & 2         & \yes      & wikipedia articles    \\
\href{https://huggingface.co/datasets/sst2}{SST-2}          & 67k                 & 872                  & sentiment classification  & 2         & \yes      & movie reviews         \\
\href{https://huggingface.co/datasets/trec}{TREC-6}         & 5452                & 500                  & question classification   & 6         & \yes      & miscellaneous         \\
\href{https://huggingface.co/datasets/jigsaw_toxicity_pred}{Wiki Talk}      & 159k                & 64k                  & toxic comment detection   & 2         & \no       & wikipedia talk pages  \\
\href{https://huggingface.co/datasets/yelp_review_full}{Yelp-5}         & 650k                & 50k                  & sentiment classification  & 5         & \yes      & yelp reviews          \\ \bottomrule
\end{tabular}}
\end{table}
\vspace{-0.5cm}
\setlength{\textfloatsep}{4pt}

\textbf{AG's News \protect\cite{zhang2015character}} is one of the larger data sets for multi-class news classification, with a target variable consisting of four classes and 120k observations in the training set. Further, it is currently the most popular one for \gls*{dal} with transformers (cf. Tab. \ref{tab:datasets_related}). \textbf{Banking77 \protect\cite{casanueva2020}} comes with the task of conversational language understanding alongside intent detection. We include it due to its yet unprecedented high-class cardinality for \gls*{dal} applications. The small unlabeled pool of 10k observations makes this data set even more challenging. \textbf{DBPedia \protect\cite{lehmann2015}} is representative of medium class cardinality (14 classes). It poses a rather easy multi-class (ontology) classification problem indicated by a high passive baseline performance, leaving much room for \gls*{dal} strategies to prove their effectiveness. \textbf{FNC-1 \protect\cite{pomerleau2017}} was created as a subset of the Emergent data set \cite{ferreira2016emergent} for the 2017 Fake News Challenge. Samples consist of news headlines and articles, with the stance detection task being to predict their relationship. It contains long texts and has a strongly imbalanced class distribution ($> 70\%$ unrelated stance), posing an interesting and challenging task for \gls*{dal}. \textbf{MNLI \protect\cite{williams2018}, QNLI \protect\cite{wang2018glue} and SST-2 \protect\cite{socher2013}} are taken from the popular GLUE benchmark. While the former two are sentence-pair inference tasks for textual entailment (MNLI) and question answering (QNLI), the latter is a single-sentence sentiment classification task (SST-2). Both question answering and sentiment classification represent important practical challenges, while entailment is a linguistically meaningful task. \textbf{TREC-6 \protect\cite{li2002}} presents a multi-class classification task requiring question language understanding. It is a valuable addition to \textsc{ActiveGLAE} due to its short average text length ($\sim$10 words) and the small training pool of 5.5k questions. \textbf{Wikipedia Talk \protect\cite{wulczyn2017}} is a large-scale data set of Wikipedia discussion comments, potentially containing toxic content. The binary classification task to detect toxicity in these comments is imbalanced (only $\sim$10\% toxic) and comes with a rather large training set size ($\sim$159k samples). \textbf{Yelp-5 \protect\cite{zhang2015character}} was introduced for (multi-class) sentiment detection and constitutes a good counterpart to the binary sentiment classification task (SST-2). 

\section{Experimental Setup}
\label{sec:exp}
\paragraph{Baselines.} To obtain the baseline results, we follow our insights from Section \ref{subsection:trainstrat} and simulate a real-world \gls*{dal} scenario without a validation set. We employ BERT~\cite{devlin-etal-2019-bert}, DistilBERT~\cite{sanh2020}, and RoBERTa~\cite{liu2019} as they are among the most popular contemporary models. We resort to a model for sequence classification from the Huggingface Transformers code base \cite{wolf-etal-2020-transformers} to ensure reproducibility. We adopt the model hyperparameters from \cite{jukić2022}: a short training (\st) with 5 epochs \cite{devlin-etal-2019-bert} and a long training (\textsc{lt}) with 15 epochs \cite{yu2022cold}. We use AdamW \cite{loshchilov2017decoupled} with a learning rate of 5e-5 and a linear scheduler with a warmup of 5\% \cite{yu2022cold}. 

To establish baseline performance values, we implement popular query strategies that cover uncertainty-based (\textsc{entropy}), diversity-based (\textsc{coreset}), and hybrid approaches (\textsc{badge}, \textsc{cal}). We employ random sampling as the baseline query strategy. The model's \textsc{[cls]} token embedding corresponds to the embedding of an instance $\mathbf{x}$. For \textsc{entropy}, and \textsc{cal}, we aggregate a batch $\mathcal{B}(t)$ by greedily selecting the most-useful instances until we reach the desired query size $b$. For \textsc{badge} and \textsc{coreset}, $b$ instances are selected in a batch $\mathcal{B}(t)$ (cf. Appendix \ref{appendix:query_strats} for further details). We focus on a data warm-start and begin each \gls*{dal} experiment by initializing 100 randomly selected instances (fixed per seed) from the unlabeled pool. We assess the impact of low and high budgets by setting them to 500 and 1600, respectively (c.f. Tab. \ref{tab:dalregimes}). For each cycle iteration $t$, we select a query size $b$ of 100 and randomly query a subset of 10,000 instances from the entire unlabeled pool $\mathcal{U}(t)$. This enables us to reduce the query times and substantially increase the number of experiments. To ensure a fair comparison, we maintain fixed hyperparameters and \gls*{dal} settings across all tasks. For a comprehensive overview, please refer to Tab. \ref{tab:hyperparameters} in Appendix \ref{appendix:trainin settings}.

\paragraph{Ablations.} In ablation studies, we only investigate BERT with \st and \lt to reduce computational runtime. More specifically, we examine the impact on model performance across query strategies by (1) using a model warm-start instead of a model cold-start in each cycle iteration, (2) querying the entire unlabeled pool without a subset, and (3) reducing the query size to 25 with the same budget resulting in 60 cycle iterations. Additionally, we adopt the learning strategy from \cite{mosbach2021}, who recommend a lower learning rate (2e-5), an increased number of epochs (20) warmup ratio of 10\% (\ltplus) to increase training stability on small datasets.  
\vspace{-0.2cm}
\paragraph{Evaluation.}
Following related work, we visualize the performance of \gls*{dal} query strategies with learning curves for all experiments \cite{schröder2022,yu2022cold}. For balanced data sets, we use accuracy, while for imbalanced data sets (FNC-1, WikiTalk), we report the balanced accuracy \cite{brodersen2010} as metrics to report test performance after each cycle iteration. To facilitate comparability, we additionally present a single metric for each query strategy's model performance on the test set - the final (balanced) accuracy (FAC) and the area under the learning curve (AUC). The AUC score is normalized to ensure comparability \cite{schröder2022}. In addition to reporting the results for each dataset individually, we compute an aggregate performance score by taking the average across the entire \textsc{ActiveGLAE} benchmark, with equal weighting for each dataset, following the methodology of GLUE. We repeat each experiment with five random seeds.
\vspace{-0.2cm}
\section{Results}
\label{sec:results}
\begin{figure}[!ht]
\begin{picture}(300, 205)
    \put(35,200){\includegraphics[width=0.8\textwidth]{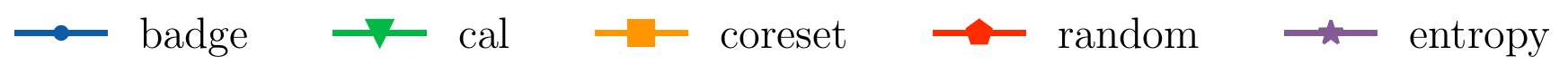}}

    \put(0,125){\includegraphics[width=0.33\textwidth]{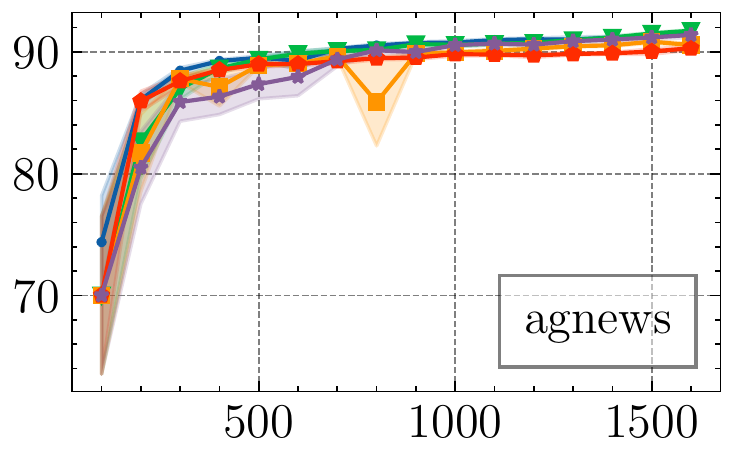}}
    \put(115,125){\includegraphics[width=0.33\textwidth]{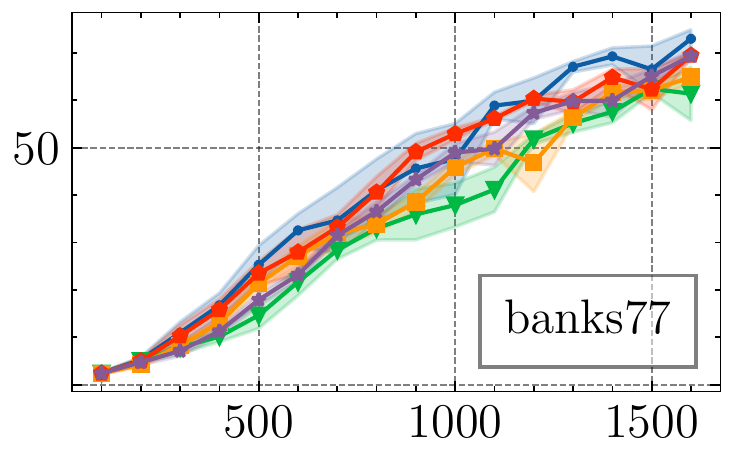}}
    \put(230,125){\includegraphics[width=0.33\textwidth]{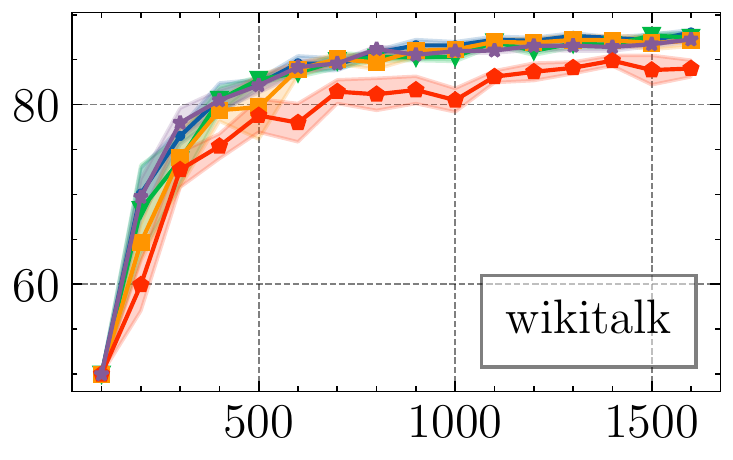}}

    \put(0,63){\includegraphics[width=0.33\textwidth]{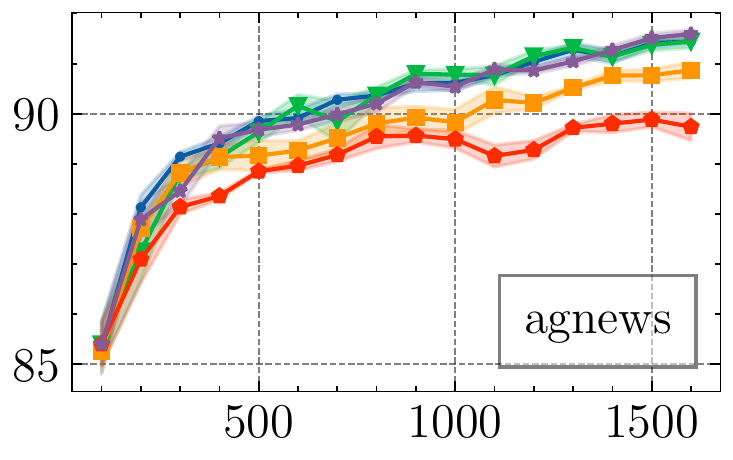}}
    \put(115,63){\includegraphics[width=0.33\textwidth]{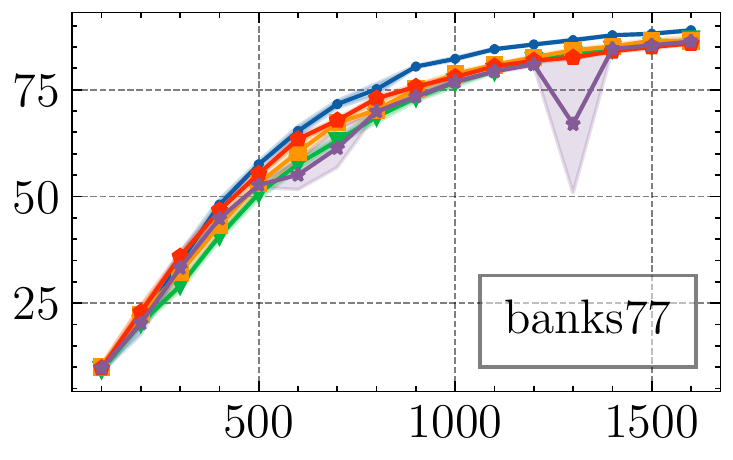}}
    \put(230,63){\includegraphics[width=0.33\textwidth]{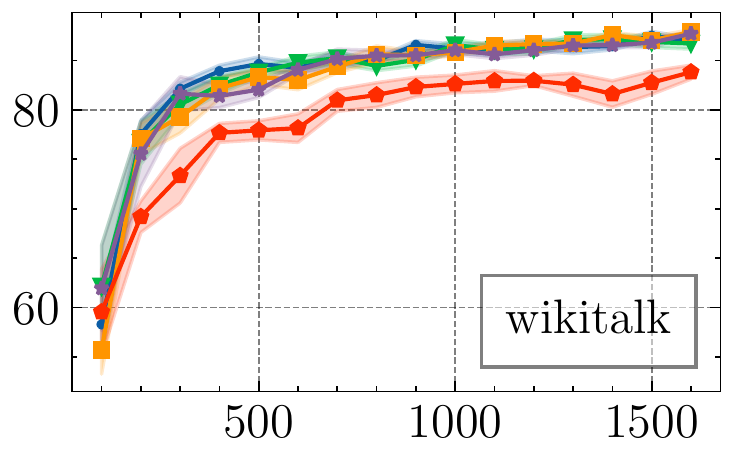}}

    \put(0,0){\includegraphics[width=0.33\textwidth]{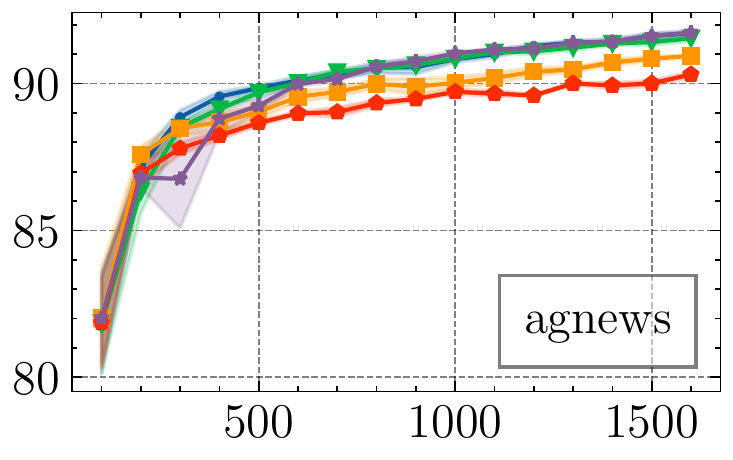}}
    \put(115,0){\includegraphics[width=0.33\textwidth]{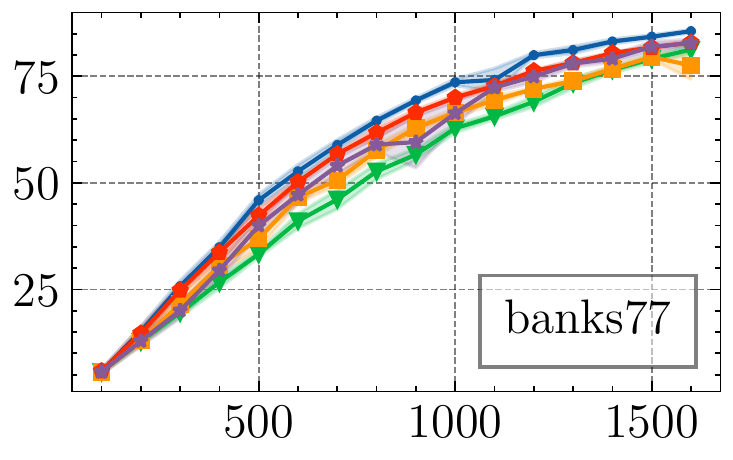}}
    \put(230,0){\includegraphics[width=0.33\textwidth]{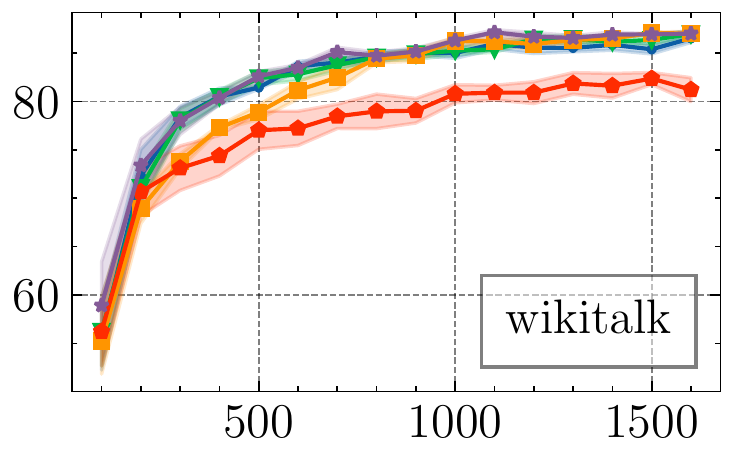}}

    \put(60,144.5){(\footnotesize{\st})}
    \put(60,82.5){(\footnotesize{\lt})}
    \put(55,20.5){(\footnotesize{\ltplus})}

    \put(172,144.5){(\footnotesize{\st})}
    \put(172,82.5){(\footnotesize{\lt})}
    \put(167,20.5){(\footnotesize{\ltplus})}

    \put(288,144.5){(\footnotesize{\st})}
    \put(288,82.5){(\footnotesize{\lt})}
    \put(283,20.5){(\footnotesize{\ltplus})}

\end{picture}
\caption{Selected learning curves for BERT reporting test accuracy with \protect\st, \protect\lt and \protect\ltplus.}
\label{fig:bertbaseline}
\end{figure}
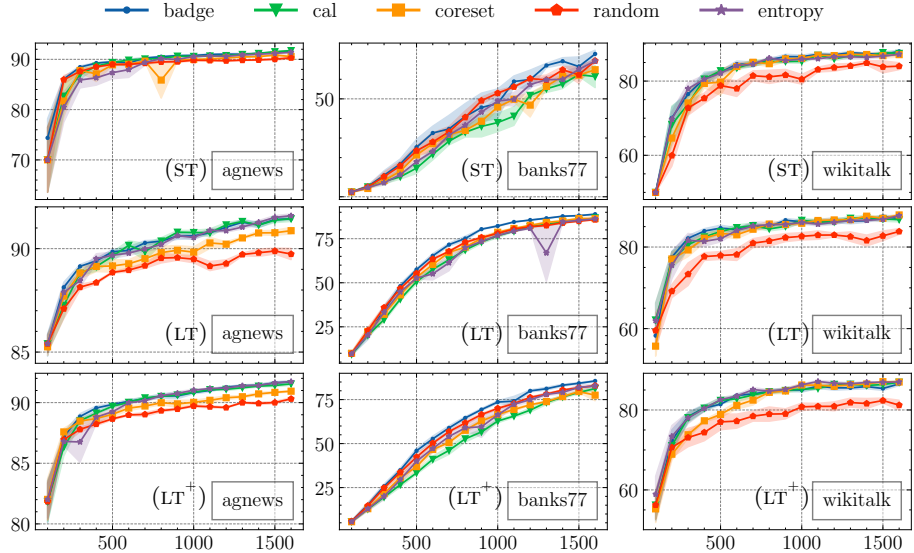

\vspace{-0.3cm}
In this section, we present our main baseline results as well as the ablations based on the three challenges - data set selection (C1), model training (C2), and DAL settings (C3) - we identified in the design of a \gls*{dal} process in Sec.~\ref{sec:current}. For each challenge, we provide a main takeaway summarizing our key findings. Tab. \ref{tab:AUC Bert Results} reports the AUC results of our extensive benchmark study on \textsc{ActiveGLAE} with BERT. Fig. \ref{fig:bertbaseline} shows a selection of learning curves with \textsc{st},\textsc{lt}, and \textsc{lt}$^+$ that supply the first two challenges (C1, C2). Additionally, we depict selected learning curves for our ablations to address the third challenge (C3). For additional details and the complete set of results, refer to Appendix \ref{appendix: baseline results} and \ref{appendix:ablations}.
\vspace{-0.2cm}
\paragraph{(C1) Data sets.}
Our results suggest that the effectiveness of a \gls*{dal} query strategy crucially depends on the data set and the accompanying task, as the results vary notably (cf. Tab.~\ref{fig:bert_15_improvement_to_random} in Appendix \ref{appendix: baseline results}). We see in Tab.~\ref{tab:AUC Bert Results} that model performance across query strategies hinges on data set characteristics, including task difficulty, task type, class cardinality, and class balance. For instance, we observe in Fig. \ref{fig:bertbaseline} that the performance differences are most noticeable in the imbalanced task (Wikitalk), and the query strategies' effectiveness vary across tasks. In fact, we record the highest model performance gains of \gls*{dal} imbalanced data sets (Banks77 and Wikitalk) across all models. This parallels results shown in related work \cite{lüth2023,yi2022,ein-dor2020}.
\mbox \\\begin{tcolorbox}[arc=0pt, boxrule=0.5pt, left=2pt, top=2pt, bottom=2pt]
\textbf{Takeaway.} The results of query strategies vary notably across data sets in our benchmark suite. We observe the greatest gains in model performance with \gls*{dal} in class-imbalance scenarios. To evaluate the robustness of query strategies and model training, using a diverse collection of datasets with varying tasks, difficulties, and class imbalances is crucial.
\end{tcolorbox}

\setlength{\textfloatsep}{15pt}
\paragraph{(C2) Model training.}
Our benchmark study emphasizes model training in \gls*{dal} \cite{margatina2021a} and highlights the importance of the number of epochs as a hyperparameter when no validation set is available. Employing a short training \st with only five epochs leads to consistently worse model performance regardless of the query strategy or task. Especially when applying \gls*{dal} with a low budget, we observe substantial disparities between \st and \lt in the overall AUC (Tab. \ref{tab:AUC Bert Results}) and FAC (Tab.\ref{tab:FinalACC Bert Results}, Appendix\ref{appendix: baseline results}) benchmark scores. This is prominently shown on Banks77 in Tab. \ref{tab:AUC Bert Results}, where the differences are as high as~22\%. Although the differences diminish with a higher budget, extensive training time still leads to better model performance across all query strategies. We report similar results for DistilBERT and RoBERTa in Tab. \ref{tab:AUC Distilbert Results} and \ref{tab:AUC Roberta Results} in Appendix \ref{appendix: baseline results}. 
Table \ref{tab:AUC Bert Results} also illustrates that the ranking of query strategies is heavily influenced by model training, except for \textsc{badge}, which consistently outperforms the other strategies. Additionally, the ablation \ltplus with a lower learning rate (2e-5) and more epochs (20) performs better than \st but generally worse than \lt. However, we see in Fig. \ref{fig:bertbaseline} that \ltplus further stabilizes the learning curves with a lower learning rate and more epochs, which follows the findings of \cite{mosbach2021}. 
\vspace{-0.4cm}
\mbox \\
\begin{tcolorbox}[arc=0pt, boxrule=0.5pt, left=2pt, top=2pt, bottom=2pt]
\textbf{Takeaway.} Increasing the number of epochs leads to improved model performance regardless of the budget and query strategy. While the performance boosts in the low-budget setting are larger, higher budget sizes also profit from longer training times (15-20 epochs). Model hyperparameters impact the overall performance more than the query strategy. The rankings of query strategies vary depending on the model hyperparameters, except for \textsc{badge}, which consistently outperforms all other strategies.
\end{tcolorbox}

\paragraph{(C3) Deep active learning setting.} 
The ablations provide additional baseline results to serve as a reference point for future work. In the following, we examine the results based on key factors of a \gls*{dal} setting that may influence model performance.
\begin{itemize}[wide, labelindent=0pt]
\item{\textit{Model initialization and update.}}
Surprisingly, the results in Tab. \ref{tab:auc_bert_warm} reveal that model warm-start improves overall model performance compared to model-cold-start regardless of the query strategy. The difference is especially pronounced when using a low budget with \textsc{st} and for more complicated tasks like Banks77 or FNC-1. Although less pronounced, we also observe improvements for \lt and in a high-budget setting. Interestingly, the performance disparities between \st and \lt are drastically reduced across query strategies when employing a model warm-start. This is further illustrated in Fig. \ref{figure:selectedwarmstart}, where we can see that using \st for model warm-start can achieve even higher model performance than with \lt in Banks77 (cf. Tab. \ref{tab:auc_bert_warm}, \ref{appendix:ablations}).

\begin{figure}[h]
\begin{picture}(300, 80)
    \put(40,75){\includegraphics[width=0.8\textwidth]{figures/resultPlots/legend.pdf}}

    \put(0,0){\includegraphics[width=0.347\textwidth]{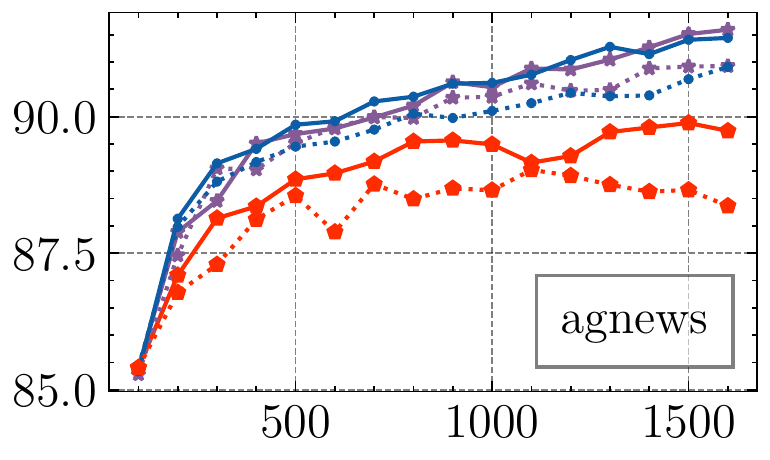}}
    \put(120,0){\includegraphics[width=0.33\textwidth]{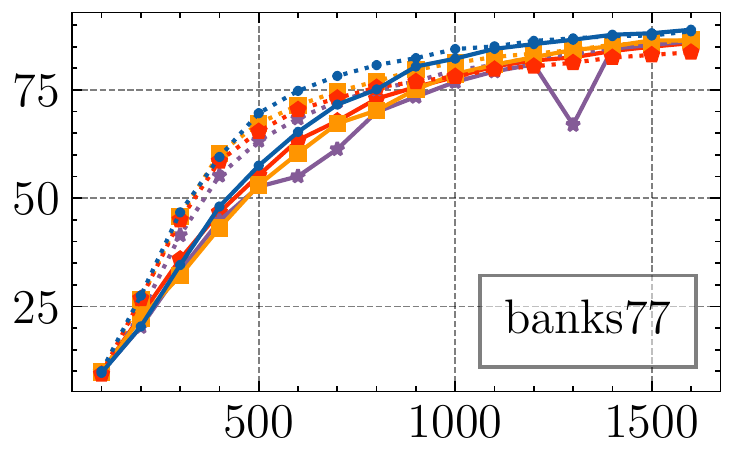}}
    \put(235,0){\includegraphics[width=0.33\textwidth]{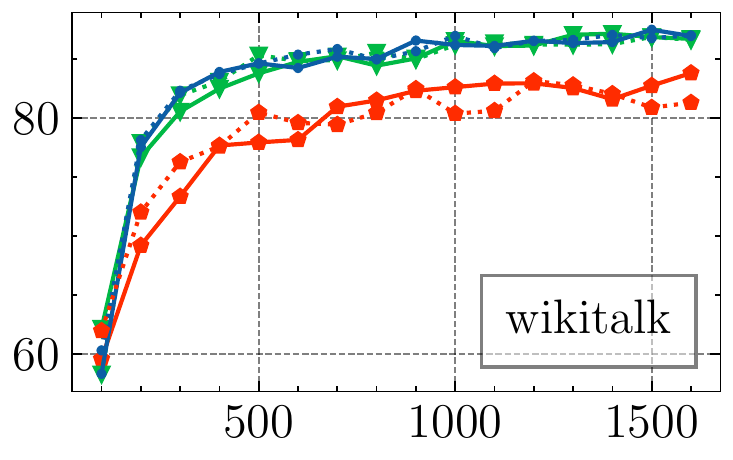}}
 
    \put(65,19){(\footnotesize{\lt})}
    \put(174,19){(\footnotesize{\lt})}
    \put(290,19){(\footnotesize{\lt})}
\end{picture}
\caption{Learning curves for BERT with {\lt} to compare test accuracy between the baseline with \textbf{model cold-start} (straight lines) and the ablation with model warm-start (dotted lines).}
\label{figure:selectedwarmstart}
\end{figure}
\vspace{-0.5cm}
\item{\textit{Stopping criterion:} We consider the budget an essential parameter governing the resulting model performance of a query strategy. In the low-budget setting, only \textsc{badge} and \textsc{coreset} can outperform \textsc{random} while \textsc{cal} performs worse and \textsc{entropy} has only a minimal improvement (cf. Tab. \ref{tab:AUC Bert Results}). This is in line with the findings of \cite{hacohen2022}, who report \textsc{random} outperforming most \gls*{dal} query strategies in low-budget settings. With a higher budget, the differences become more prevalent, and all query strategies outperform \textsc{random}. Interestingly, when considering the FAC in Tab. \ref{tab:FinalACC Bert Results} in Appendix  \ref{appendix: baseline results}, almost all query strategies outperform \textsc{random}. While we report similar results for DistilBERT (cf. Tab.~\ref{tab:AUC Distilbert Results}, \ref{appendix: baseline results}), the overall model performance for RoBERTa diminishes with a query strategy in a high-budget setting (cf. Tab. \ref{tab:AUC Roberta Results}, \ref{appendix: baseline results}). Additionally, we see in Tab. \ref{tab:AUC Bert Results} that the ranking of query strategies change with different budgets, except for \textsc{badge}.}

\item{\textit{Query strategy:} While no query strategy clearly outperforms all other strategies, \textsc{badge} a exhibits consistently better performance. The AUC and the FAC are often superior for the low and the high data budgets compared to other query strategies. Fig.~\ref{fig:bert_15_improvement_to_random} in Appendix \ref{appendix: baseline results} further highlights this by showing nearly constant positive improvements upon \textsc{random} performance on our \textsc{ActiveGLAE} benchmark. Moreover, it emphasizes the variability in the results across tasks.}

\item{\textit{Pool subset:} Interestingly, we can see no meaningful drawback in the effectiveness of a query strategy when employing a dynamic pool subset (cf. Tab.\ref{tab:auc_bert_subset} \ref{appendix:ablations}). Note that the pool is redrawn in each cycle iteration rather than fixed once, increasing the diversity of the queried pool. By reducing the pool size, the query time per cycle iteration is heavily reduced, which allows an extensive set of experiments and reduces the computational time accordingly. Without a subset, our baseline results exhibit an approximately 10-fold increase in average query time. Notably, uncertainty-based sampling (\textsc{entropy}) even seems to improve by using the more diverse pool subsets, which can be seen in Fig.~\ref{fig:selectednosubset}. }
\begin{figure}[!ht]

\begin{picture}(300, 80)
    \put(40,75){\includegraphics[width=0.8\textwidth]{figures/resultPlots/legend.pdf}}
    \put(0,0){\includegraphics[width=0.347\textwidth]{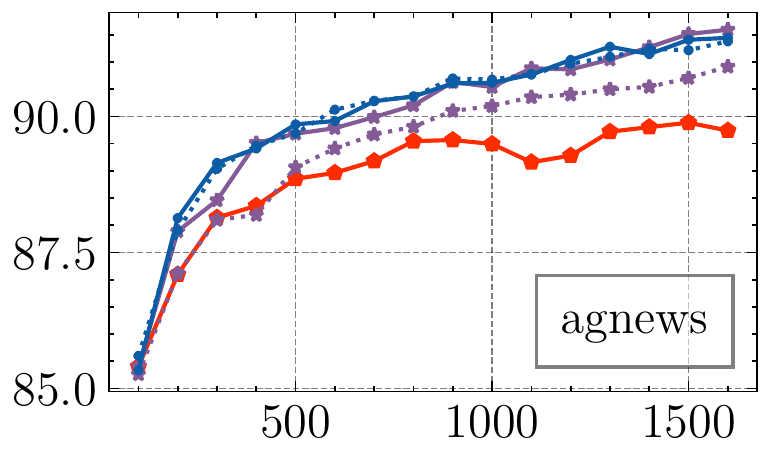}}
    \put(120,0){\includegraphics[width=0.33\textwidth]{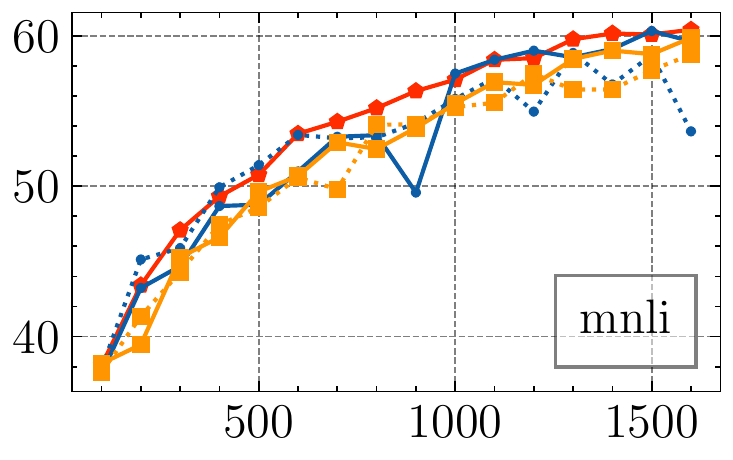}}
    \put(235,0){\includegraphics[width=0.33\textwidth]{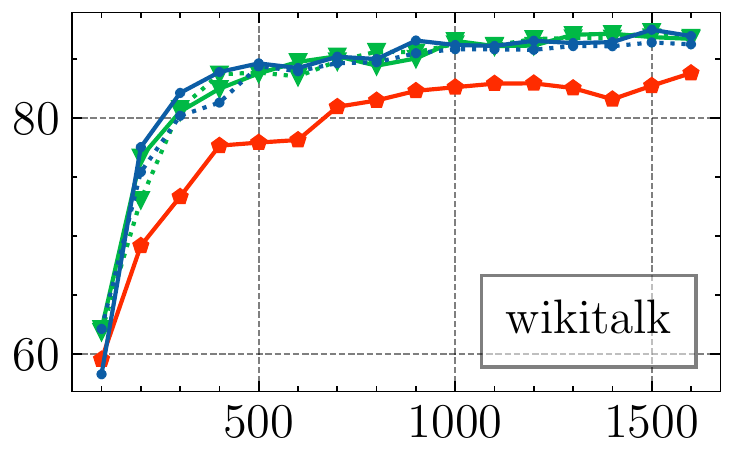}}
    \put(65,19){(\footnotesize{\lt})}
    \put(187,19){(\footnotesize{\lt})}
    \put(290,19){(\footnotesize{\lt})}
\end{picture}
\caption{Learning curves for BERT with {\lt} to compare test accuracy between the baseline with a \textbf{pool subset} (straight lines) and the ablation with no subset (dotted lines).}
\label{fig:selectednosubset}
\end{figure}
\vspace{-0.5cm}
\item{\textit{Query size:}  Comparing the ablation results with the baseline results, we notice a minimal impact on model performance across query strategies (cf. Tab.\ref{tab:auc_bert_querybatch}, \ref{appendix:ablations}). These findings align with the results of \cite{beck2021} from \gls*{cv} but differ from those of \cite{lüth2023} in the \gls*{cv} domain. However, the investigated query sizes in \gls*{cv} are much larger (\textgreater1000), while we investigate in a smaller setting between 25 and 100. Additionally, in Fig.~\ref{fig:selected2560}, we observe that more re-trainings (60 instead of 25) also increase the noise in the learning curve.}
\begin{figure}[!ht]

\begin{picture}(300, 80)
    \put(40,75){\includegraphics[width=0.8\textwidth]{figures/resultPlots/legend.pdf}}

    \put(0,0){\includegraphics[width=0.347\textwidth]{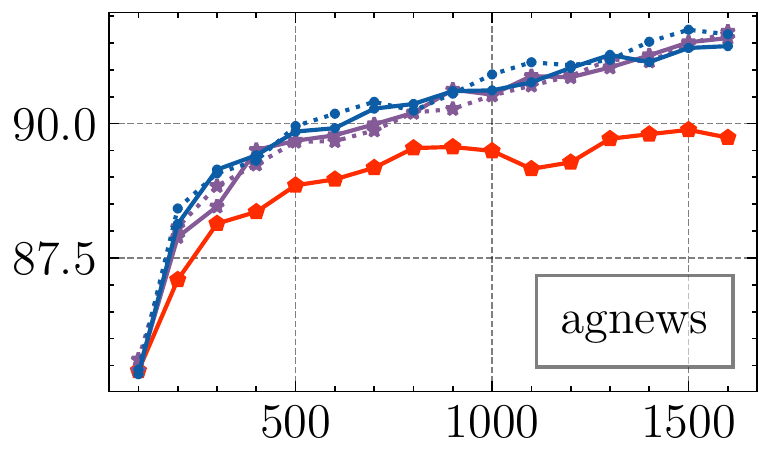}}

    \put(120,0){\includegraphics[width=0.33\textwidth]{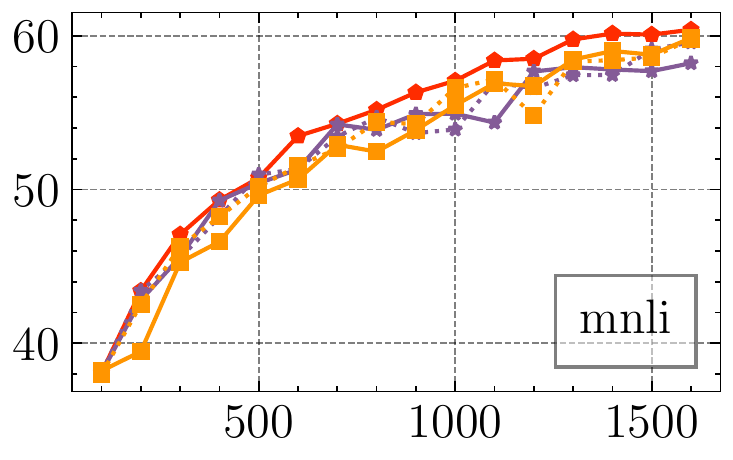}}
    \put(235,0){\includegraphics[width=0.33\textwidth]{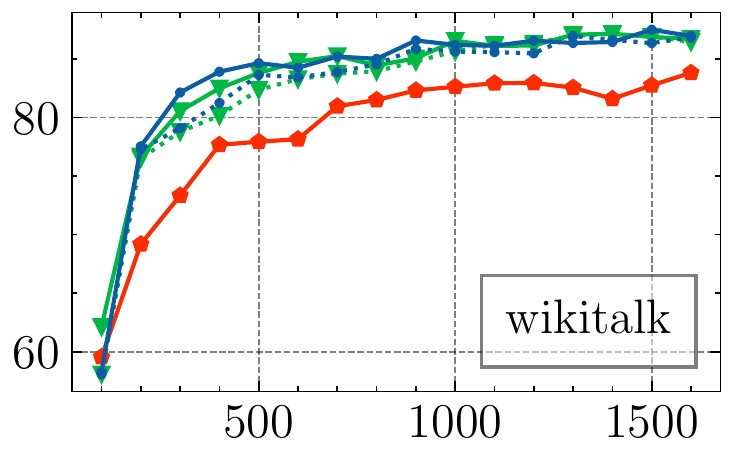}}

    \put(65,19){(\footnotesize{\lt})}
    \put(187,19){(\footnotesize{\lt})}
    \put(290,19){(\footnotesize{\lt})}
\end{picture}
\caption{Learning curves for BERT with \lt to compare test accuracy between the baseline with an \textbf{query size} of 100 (straight lines) and the ablation with 25 (dotted lines). }
\label{fig:selected2560}
\end{figure}
\setlength{\textfloatsep}{2pt}
\vspace{-1.5cm}
\end{itemize}

\mbox \\
\begin{tcolorbox}[arc=0pt, boxrule=0.5pt, left=2pt, top=2pt, bottom=2pt]
\textbf{Takeaway:} Model warm-start improves performance across query strategies while mitigating the impact of model hyperparameters on the results. Utilizing a dynamically sampled pool subset does not adversly affect performance but substantially reduces query time. Minor adjustments to the query size do not have a notable influence on resulting model performance.  

\end{tcolorbox}
\begin{table*}[t!]
\centering
    \caption{Baseline \textbf{BERT AUC results} on \textsc{ActiveGLAE} with \protect\st, \protect\lt, \protect\ltplus, two budget sizes (500, 1600) and 5 repetitions ($\pm$standard deviation). \textbf{Best} and \underline{second} best results are highlighted for each dataset. $\color{blue}\uparrow$ and $\color{red}\downarrow$ demonstrate improvements over \textsc{random}. \textbf{R}anking indicates the placement of a query strategy.}
    \label{tab:AUC Bert Results}
    \resizebox{0.97\textwidth}{!}{%
    \begin{tabular}{l | c | c | c | c | c | c | c | c | c | c | c || c c} 
        \toprule
        \multicolumn{14}{c}         {{\cellcolor{yescolor!10}
        \textbf{Low Data Budget: $100+400$}}} \\
        \midrule \rowcolor{white}
               &  & \textbf{AGN}  & \textbf{B77}  & \textbf{DBP} & \textbf{FNC1}&  \textbf{MNLI}& \textbf{QNLI}& \textbf{SST2}& \textbf{TREC6}& \textbf{WIKI} & \textbf{YELP5} & \textbf{Average} & \textbf{R}\\ 
        \specialrule{0.2pt}{0.5pt}{0.5pt}
        
        \multirow{3}{*}{\STAB{\rotatebox[origin=c]{90}{\scriptsize{\textbf{\textsc{random}}}}}}  
            & \multirow{1}*{\textsc{st}}   & 85.4$\pm{1.8}$    & 11.0$\pm{3.1}$    & 79.5$\pm{4.2}$    & 31.7$\pm{4.6}$    & 41.2$\pm{0.9}$    & 63.6$\pm{2.0}$    & 81.9$\pm{1.9}$    & 84.2$\pm{3.8}$    & 68.1$\pm{1.4}$    & 43.5$\pm{2.0}$    & 59.2 $\leftarrow$ & 2\\ [0.15em] 
                                 
            & \multirow{1}*{\textsc{lt}}   & 87.7$\pm{0.3}$    & \textbf{34.5}$\pm{1.6}$    & \underline{96.6}$\pm{0.4}$    & 43.92$\pm{4.8}$    & \textbf{46.1}$\pm{1.6}$    & 67.6$\pm{1.9}$    & 82.1$\pm{2.3}$    & 90.7$\pm{0.9}$    & 72.2$\pm{3.6}$    & \textbf{48.8}$\pm{1.6}$    & 67.3 $\leftarrow$ & 4\\  [0.15em]  

            & \multirow{1}*{\textsc{lt}$^+$}   & 87.1$\pm{0.7}$    & 24.3$\pm{2.3}$    & 93.6$\pm{1.2}$    & 42.6$\pm{4.3}$    & 45.0$\pm{2.3}$    & 66.5$\pm{2.0}$    & 82.2$\pm{2.3}$    & 89.8$\pm{2.4}$    & 71.2$\pm{4.2}$    & 46.9$\pm{2.2}$    & 64.93 $\leftarrow$ & 2\\  [0.15em]
        \cline{1-14} 
        \midrule
        
        \multirow{3}{*}{\STAB{\rotatebox[origin=cb]{90}{{\scriptsize\textbf{entropy}}}}} 
            & \multirow{1}*{\textsc{st}}   & 82.8$\pm{3.6}$    & 8.3$\pm{1.4}$    & 74.2$\pm{6.7}$    & 32.7$\pm{2.5}$    & 42.0$\pm{1.8}$    & 62.9$\pm{2.7}$    & 82.2$\pm{1.6}$    & 80.3$\pm{5.3}$    & 73.5$\pm{1.7}$    & 42.6$\pm{2.9}$    & 58.18 $\color{red}\downarrow$\color{red}\scriptsize{0.83} & 5 \\ [0.15em]
                                 
            & \multirow{1}*{\textsc{lt}}   & \underline{88.3}$\pm{0.3}$    & 32.3$\pm{1.4}$    & 94.9$\pm{2.2}$    & \underline{46.9}$\pm{3.1}$    & 45.5$\pm{0.8}$    & \textbf{67.81}$\pm{2.1}$    & 82.3$\pm{2.1}$    & 90.5$\pm{2.2}$    & 77.6$\pm{2.6}$    & 46.5$\pm{1.5}$    & 67.28 $\color{blue}\uparrow$\color{blue}\scriptsize{0.26} & 3\\  [0.15em]  

            & \multirow{1}*{\textsc{lt}$^+$}  & 87.0$\pm{1.5}$    & 21.3$\pm{2.8}$    & 92.7$\pm{2.5}$    & 38.8$\pm{1.9}$    & 45.7$\pm{1.2}$    & 65.0$\pm{2.9}$    & \underline{83.1}$\pm{1.2}$    & 89.0$\pm{2.3}$    & 75.6$\pm{2.5}$    & 45.0$\pm{2.2}$    & 64.34 $\color{red}\downarrow$\color{red}\scriptsize{0.59} & 4\\  [0.15em]
        \cline{1-14}  
        
        \multirow{3}{*}{\STAB{\rotatebox[origin=cb]{90}{{\scriptsize\textbf{badge}}}}}  
        
            & \multirow{1}*{\textsc{st}}   & 86.4$\pm{1.2}$    & 11.7$\pm{3.6}$    & 84.4$\pm{1.8}$    & 33.1$\pm{3.7}$    & 42.1$\pm{2.9}$    & 65.6$\pm{2.1}$    & 81.9$\pm{1.6}$    & 85.5$\pm{3.4}$    & 73.3$\pm{2.1}$    & 45.4$\pm{1.1}$    & 60.96 $\color{blue}\uparrow$\color{blue}\scriptsize{1.95} & 1\\ [0.15em] 
                            
            & \CC \multirow{1}{*}\textsc{lt}   &\CC\textbf{88.5}$\pm{0.2}$    &\CC \underline{34.1}$\pm{2.6}$    & \CC\textbf{96.9}$\pm{0.6}$    &\CC 46.0$\pm{4.5}$    &\CC 44.9$\pm{1.7}$    & \CC66.5$\pm{2.6}$    & \CC\textbf{83.2}$\pm{0.8}$    & \CC\textbf{92.3}$\pm{1.0}$    & \CC\textbf{78.7}$\pm{1.3}$    & \CC\underline{48.4}$\pm{1.6}$    & \CC\textbf{67.97} $\color{blue}\uparrow$\color{blue}\scriptsize{0.95} & \CC1 \\  

            & \multirow{1}*{\textsc{lt}$^+$} & 87.8$\pm{0.6}$    & 25.4$\pm{2.0}$    & 94.9$\pm{1.0}$    & 42.6$\pm{3.6}$    & 44.6$\pm{1.3}$    & 66.5$\pm{2.7}$    & 82.7$\pm{2.1}$    & 91$\pm{2.1}$    & 74.9$\pm{2.4}$    & 46.7$\pm{2.4}$    & 65.73 $\color{blue} \uparrow$\color{blue}\scriptsize{0.80} & 1\\  [0.15em]
        \cline{1-14} 
        
        \multirow{3}{*}{\STAB{\rotatebox[origin=cb]{90}{{\scriptsize\textbf{coreset}}}}}  
            & \multirow{1}*{\textsc{st}}   & 84.0$\pm{1.8}$    & 9.3$\pm{1.1}$    & 74.6$\pm{4.2}$    & 34.2$\pm{3.3}$    & 40.1$\pm{2.1}$    & 62.9$\pm{0.6}$    & 78.6$\pm{5.1}$    & 82.7$\pm{3.5}$    & 70.7$\pm{2.6}$    & 43.7$\pm{1.3}$    & 58.12 $\color{red} \downarrow$\color{red}\scriptsize{0.89} & 4 \\ [0.15em] 
                                 
            & \multirow{1}*{\textsc{lt}}& 88.2$\pm{0.4}$    & 32.3$\pm{3.9}$    & 96.3$\pm{1.1}$    & \textbf{49.2}$\pm{4.0}$    & 43.8$\pm{1.0}$    & 66.6$\pm{1.8}$    & 82.3$\pm{2.1}$    & 91.5$\pm{1.8}$    & 77.0$\pm{1.3}$    & 47.0$\pm{1.0}$    & \underline{67.42} $\color{blue}\uparrow$\color{blue}\scriptsize{0.40} & 2 \\  [0.15em]  

            & \multirow{1}*{\textsc{lt}$^+$}  & 87.5$\pm{0.3}$    & 21.6$\pm{2.5}$    & 94.7$\pm{0.7}$    & 44.4$\pm{3.0}$    & 45.9$\pm{1.5}$    & 64.8$\pm{2.9}$    & 80.4$\pm{2.4}$    & 90.9$\pm{1.8}$    & 71.7$\pm{1.4}$    & 44.8$\pm{2.0}$    & 64.71 $\color{red} \downarrow$\color{red}\scriptsize{0.22} & 3 \\  [0.15em]
        \cline{1-14} 
        
        \multirow{3}{*}{\STAB{\rotatebox[origin=cb]{90}{{\scriptsize\textbf{cal}}}}}  
            & \multirow{1}*{\textsc{st}}   & 84.5$\pm{3.7}$    & 7.9$\pm{1.6}$    & 71.3$\pm{4.4}$    & 32.9$\pm{3.3}$    & 42.1$\pm{1.9}$    & 64.7$\pm{1.6}$    & 81.9$\pm{2.8}$    & 80.9$\pm{5.2}$    & 72.3$\pm{4.5}$    & 40.0$\pm{2.9}$    & 57.86 $\color{red}\downarrow$\color{red}\scriptsize{1.15} & 5 \\ [0.15em]
                                 
            & \multirow{1}*{\textsc{lt}}  & 88.2$\pm{0.3}$    & 30.0$\pm{1.5}$    & 94.5$\pm{1.6}$    & 43.2$\pm{4.0}$    & \underline{46.0}$\pm{2.1}$    & \underline{67.7}$\pm{2.1}$    & 82.4$\pm{2.4}$    & \underline{91.7}$\pm{1.1}$    & \underline{78.2}$\pm{1.6}$    & 47.5$\pm{1.2}$    & 66.94 $\color{red}\downarrow$\color{red}\scriptsize{0.08} & 5\\  [0.15em]  

            & \multirow{1}*{\textsc{lt}$^+$} & 87.44$\pm{0.8}$    & 19.62$\pm{1.3}$    & 91.91$\pm{1.9}$    & 40.72$\pm{2.9}$    & 45.24$\pm{2.9}$    & 65.99$\pm{3.2}$    & 82.20$\pm{2.1}$    & 90.44$\pm{1.8}$    & 74.73$\pm{1.9}$    & 44.52$\pm{1.5}$    & 64.28 $\color{red}\downarrow$\color{red}\scriptsize{0.65}& 5 \\  [0.15em]
    \midrule
    \multicolumn{14}{c}
    {\cellcolor{yescolor!40}
    {\textbf{High Data Budget: $100+1500$}}}\\
\specialrule{0.2pt}{0.5pt}{0.5pt}
        
        \multirow{3}{*}{\STAB{\rotatebox[origin=c]{90}{\scriptsize{\textbf{random}}}}}  
            & \multirow{1}*{\textsc{st}} & 88.5$\pm{0.5}$    & 39.8$\pm{3.2}$    & 93.6$\pm{1.1}$    & 43.8$\pm{5.6}$    & 52.6$\pm{0.8}$    & 72.8$\pm{1.2}$    & 85.6$\pm{0.6}$    & 92.1$\pm{0.8}$    & 78.4$\pm{1.4}$    & 52.5$\pm{0.5}$    & 70.0 $\leftarrow$ & 3\\ [0.15em] 
                                 
            & \multirow{1}*{\textsc{lt}}& 89.0$\pm{0.2}$    & \underline{65.3}$\pm{0.7}$    & 98.1$\pm{0.1}$    & \textbf{54.9}$\pm{2.2}$    & \textbf{54.2}$\pm{1.0}$    & 74.3$\pm{0.8}$    & 85.6$\pm{0.6}$    & 94.3$\pm{0.4}$    & 79.2$\pm{1.6}$    & \textbf{53.4}$\pm{0.6}$    & 74.83 $\leftarrow$ & 5\\  [0.15em]  

            & \multirow{1}*{\textsc{lt}$^+$}& 88.90$\pm{0.2}$    & 56.96$\pm{1.3}$    & 97.35$\pm{0.3}$    & 53.97$\pm{1.9}$    & \underline{53.80}$\pm{1.3}$    & 73.71$\pm{1.4}$    & 85.95$\pm{0.7}$    & 94.06$\pm{0.8}$    & 77.73$\pm{2.2}$    & 53.18$\pm{0.7}$    & 73.56 $\leftarrow$ & 5\\  [0.15em]
        \cline{1-14}  
        \midrule
        \multirow{3}{*}{\STAB{\rotatebox[origin=cb]{90}{{\scriptsize\textbf{entropy}}}}} 
            & \multirow{1}*{\textsc{st}} & 88.2$\pm{1.4}$    & 36.8$\pm{2.5}$    & 92.2$\pm{1.8}$    & 49.5$\pm{1.2}$    & 51.4$\pm{1.5}$    & 72.2$\pm{1.5}$    & 86.2$\pm{0.8}$    & 92.2$\pm{1.4}$    & 82.5$\pm{0.6}$    & 51.0$\pm{0.6}$    & 70.22 $\color{blue}\uparrow$\color{blue}\scriptsize{0.23} &2 \\ [0.15em] 
                                 
            & \multirow{1}*{\textsc{lt}}& 90.05$\pm{0.1}$    & 62.2$\pm{3.7}$    & 97.9$\pm{0.6}$    & 61.2$\pm{1.8}$    & 52.7$\pm{0.9}$    & \underline{74.6}$\pm{1.2}$    & 86.1$\pm{0.5}$    & 95.1$\pm{0.6}$    & 83.5$\pm{0.5}$    & 51.1$\pm{1.0}$    & 75.44 $\color{blue}\uparrow$\color{blue}\scriptsize{0.61} & 4\\  [0.15em]  

            & \multirow{1}*{\textsc{lt}$^+$}& 89.8$\pm{0.4}$    & 54.6$\pm{3.4}$    & 97.3$\pm{0.7}$    & 58.0$\pm{1.0}$    & 53.1$\pm{0.7}$    & 73.7$\pm{1.5}$    & \textbf{86.7}$\pm{0.7}$    & 94.6$\pm{0.7}$    & 83.1$\pm{0.5}$    & 51.4$\pm{1.1}$    & 74.24 $\color{blue}\uparrow$\color{blue}\scriptsize{0.68} & 2\\  [0.15em]
        \cline{1-14}  
        
        \multirow{3}{*}{\STAB{\rotatebox[origin=cb]{90}{{\scriptsize\textbf{badge}}}}}  
            & \multirow{1}*{\textsc{st}} & 89.5$\pm{0.3}$    & 41.3$\pm{7.2}$    & 95.1$\pm{0.5}$    & 50.1$\pm{3.9}$    & 52.7$\pm{1.2}$    & 73.2$\pm{2.2}$    & 86.1$\pm{0.6}$    & 93.6$\pm{1.0}$    & 82.9$\pm{0.6}$    & 53.1$\pm{0.6}$    & 71.76 $\color{blue}\uparrow$\color{blue}\scriptsize{1.77} & 1 \\ [0.15em] 
                                 
            & \CC\multirow{1}*{\textsc{lt}}  & \CC\textbf{90.1}$\pm{0.1}$    & \CC\textbf{67.8}$\pm{0.9}$    & \CC\textbf{98.5}$\pm{0.2}$    & \CC\underline{61.3}$\pm{2.1}$    & \CC52.93$\pm{0.6}$    & \CC74.1$\pm{0.9}$    & \CC86.4$\pm{0.5}$    & \CC\textbf{95.7}$\pm{0.4}$    & \CC\textbf{84.1}$\pm{0.7}$    & \CC53.1$\pm{0.7}$    & \CC\textbf{76.41} $\color{blue}\uparrow$\color{blue}\scriptsize{1.58} & \CC1\\  [0.15em]  

            & \multirow{1}*{\textsc{lt}$^+$} & \underline{90.1}$\pm{0.3}$    & 59.3$\pm{1.6}$    & 97.9$\pm{0.3}$    & 58.6$\pm{1.2}$    & 53.3$\pm{0.9}$    & 74.2$\pm{1.4}$    & \underline{86.6}$\pm{0.6}$    & 95.5$\pm{0.5}$    & 82.3$\pm{1.0}$    & \underline{53.3}$\pm{0.8}$    & 75.11 $\color{blue}\uparrow$\color{blue}\scriptsize{1.55} & 1\\  [0.15em]
        \cline{1-14} 
        
        \multirow{3}{*}{\STAB{\rotatebox[origin=cb]{90}{{\scriptsize\textbf{coreset}}}}}  
            & \multirow{1}*{\textsc{st}} & 88.1$\pm{0.8}$    & 35.6$\pm{1.9}$    & 92.3$\pm{1.2}$    & 49.6$\pm{2.5}$    & 50.5$\pm{2.0}$    & 72.0$\pm{0.7}$    & 83.9$\pm{1.8}$    & 92.7$\pm{0.9}$    & 81.8$\pm{1.2}$    & 51.2$\pm{0.3}$    & 69.79 $\color{red}\downarrow$\color{red}\scriptsize{0.20} & 4 \\ [0.15em] 
                                 
            & \multirow{1}*{\textsc{lt}}  & 89.6$\pm{0.3}$    & 64.6$\pm{1.8}$    & \underline{98.3}$\pm{0.3}$    & \textbf{62.4}$\pm{1.3}$    & 52.3$\pm{0.9}$    & 73.8$\pm{0.9}$    & 85.6$\pm{0.9}$    & 95.4$\pm{0.5}$    & 83.5$\pm{0.4}$    & 50.0$\pm{0.5}$    & 75.57 $\color{blue}\uparrow$\color{blue}\scriptsize{0.74} & 3\\  [0.15em]  

            & \multirow{1}*{\textsc{lt}$^+$} & 89.5$\pm{0.1}$    & 53.3$\pm{1.9}$    & 97.9$\pm{0.2}$    & 60.3$\pm{1.4}$    & 53.1$\pm{0.7}$    & 72.9$\pm{1.6}$    & 85.0$\pm{0.9}$    & 95.1$\pm{0.5}$    & 81.4$\pm{0.5}$    & 50.7$\pm{0.5}$    & 73.92 $\color{blue}\uparrow$\color{blue}\scriptsize{0.36} & 3\\  [0.15em]
        \cline{1-14} 
        
        \multirow{3}{*}{\STAB{\rotatebox[origin=cb]{90}{{\scriptsize\textbf{cal}}}}}  
            & \multirow{1}*{\textsc{st}}  & 89.0$\pm{1.1}$    & 32.9$\pm{3.7}$    & 91.3$\pm{1.1}$    & 47.7$\pm{3.3}$    & 51.8$\pm{1.4}$    & 73.4$\pm{1.0}$    & 86.3$\pm{0.8}$    & 92.4$\pm{1.4}$    & 82.2$\pm{1.3}$    & 49.9$\pm{1.0}$    & 69.71 $\color{red}\downarrow$\color{red}\scriptsize{0.28} & 5\\ [0.15em] 
                                 
            & \multirow{1}*{\textsc{lt}} & 90.0$\pm{0.1}$    & 62.8$\pm{1.1}$    & 97.4$\pm{0.6}$    & 59.8$\pm{1.7}$    & 53.6$\pm{1.4}$    & \textbf{74.6}$\pm{1.0}$    & 86.2$\pm{1.0}$    & \underline{95.7}$\pm{0.4}$    & \underline{83.8}$\pm{0.8}$    & 52.3$\pm{0.8}$    & \underline{75.63} $\color{blue}\uparrow$\color{blue}\scriptsize{0.80} & 2\\  [0.15em]  

            & \multirow{1}*{\textsc{lt}$^+$} & 89.9$\pm{0.3}$    & 50.6$\pm{1.2}$    & 97.0$\pm{0.5}$    & 58.1$\pm{0.8}$    & 53.2$\pm{1.4}$    & 74.1$\pm{1.6}$    & 86.6$\pm{0.8}$    & 95.2$\pm{0.6}$    & 82.4$\pm{0.6}$    & 51.9$\pm{1.1}$    & 73.90 $\color{blue}\uparrow$\color{blue}\scriptsize{0.34} & 4\\  [0.15em]
    \bottomrule
\end{tabular}

    } 
\end{table*}

\section{Conclusion}
\setlength{\textfloatsep}{10pt}
\label{sec:concl}
\vspace{-0.2cm}
In this article, we proposed the \textsc{ActiveGLAE} benchmark, which consists of various NLP classification tasks along with a robust evaluation protocol to foster comparability of \gls*{dal} with \gls*{plm}. We analyzed the scientific environment and identified three crucial challenges when designing evaluation protocols in \gls*{dal}: data set selection, model training, and DAL setting. Additionally, we conducted a comprehensive set of experiments to provide baseline results as a reference point for future work and derived guidelines for practitioners within the challenges. We found that the effectiveness of query strategies varies across tasks, making it essential to ensure their robustness to diverse NLP classification tasks to enable a real-world deployment. We confirmed the importance of model training in DAL and recommend an extended training period to improve overall model performance across query strategies. Additionally, we observed that model warm-start improves and stabilizes performance, and employing a pool subset reduces query time with minimal impact on model performance. We reported \textsc{badge} to perform consistently better than other query strategies on \textsc{ActiveGLAE}. 



\clearpage
\newpage
\newpage
\section*{Ethics statement}

\paragraph{Limitations} This work represents a snapshot of the current practice of applying DAL to the NLP domain. This comes with the limitation that we do not explicitly go into detail about other fields, like e.g. Computer Vision or Speech Processing. Especially in the former of these two fields, an active research community is working on evaluating DAL \cite{lüth2023,ji2023,li2022,lang2022,beck2021}, and the difference in modalities make a comparison across fields highly intriguing.
Further, the experimental outcome of our work is not exhaustive. We tested a limited number of models and query strategies on a given number of data sets, which we considered representative, controlling for as much exogenous influence as possible. This should be seen as a blueprint for the experimental setup rather than a definite statement about the SOTA. \\

\paragraph{Ethical considerations} To the best of our knowledge, no ethical considerations are implied by our work. There are only two aspects that are affected in a broader sense. First, the environmental impact of the computationally expensive experiments that come with evaluating \gls*{dal} strategies. Given the ever-increasing model sizes and the already controversial debate around this topic, this is a crucial aspect to consider. The second point to be addressed is substituting human labor for labeling data sets by \gls*{dal}. Especially when it comes to labeling toxic or explicit content, suitable \gls*{dal} strategies might be one way to limit human exposure to such data.
\newpage

%
%
%
%
\bibliographystyle{splncs04}
\bibliography{literatur}

\newpage
\appendix

\section{Related Work}
\label{appendix:related_work}

This section further details the current work in \gls*{dal} with transformers in the \gls*{nlp} domain. Tab. \ref{tab:trainstrats} shows the deployed training strategies and models. Tab. \ref{tab:dalregimes} gives a detailed overview of the parameters of related work in the respective \gls*{dal} setting along with selected query strategies. We classify the publications into three types: (i) comparison studies that empirically evaluate DAL, (ii) strategy papers that propose new query strategies, and (iii) training-focused articles that specifically focus on model training in the context of DAL.

\begin{table*}[ht!]
\centering
    \caption{Overview of the model hyperparameters, optimization approaches, and selected models in current work for \gls*{dal} with transformer-based \gls*{plm}. }
    \label{tab:trainstrats}
    \resizebox{1.0\textwidth}{!}{%
    
\begin{tabular}{l | c c c | c c c | c c l }
        \toprule
        \multicolumn{1}{c|}{\multirow{2}{*}{}} &
        \multicolumn{3}{c|}{\textbf{Hyperparameters}} &
        \multicolumn{3}{c|}{\textbf{Model}} & 
        \multicolumn{3}{c}{\textbf{Optimization}}\\
                              & epochs   & learning rate  & scheduler     & BERT & RoBERTa & Distiled         & Validation  & Source                   & \multicolumn{1}{c}{Approach}           \\
        \midrule
        \cite{ein-dor2020}    & 5        & 5e-5           & --            & \yes & \yes    & \no               & \yes       & complete              & best model from validation results  \\
        \cite{lu2020}         & 15       & 1e-5           & --            & \yes & \yes    & \no               & \no        & --                    & best model from train results  \\
        \cite{yuan2020}       & 3        & 2e-5           & linear        & \yes & \no     & \no               & \yes       & --                    & preliminary experiments on validation set \\
        \cite{ru2020}         & --       & --             & --            & \yes & \no     & \no               & \yes       & --                    & --  \\
        \cite{prabhu2021}     & 1        & 3e-5           & --            & \no  & \no     & \yes              & \yes       & complete              & --  \\
        \cite{margatina2021a} & 20       & 2e-5           & 10\% linear   & \yes & \no     & \no               & \yes       & 5\%,10\% train set    & early stopping with validation results  \\ 
        \cite{margatina2021}  & 3        & 2e-5           & --            & \yes & \no     & \no               & \yes       & 5\%,10\% train set    & best model from validation results  \\ 
        \cite{tan2021}        & 30       & 2e-5           & --            & \no  & \no     & \yes              & \yes       & dynamic 30\% labeled  & early stopping with validation results \\
        \cite{darcy2022}      & --       & 1e-5           & 150           & \yes & \yes    & \no               & \yes       & --                    & early stopping with validation results   \\ 
        \cite{schröder2022}   & 15       & 1e-5           & --            & \yes & \no     & \yes              & \yes       & dynamic 10\% labeled  & early stopping with validation results  \\ 
        \cite{gonsior2022}    & --       & --             & --            & \yes & \yes    & \no               & --         & --                    & --  \\ 
        \cite{seo2022}        & --       & 2e-5           & --            & \yes & \no     & \no               & \yes       & --                    & --  \\ 
        \cite{yu2022a}        & 15       & 1,2e-5         & 10\% linear   & \no  & \yes    & \no               & \yes       & 500,1000              & best model from validation results \\
        \cite{zhang2022a}     & ?        & 2e-5           & --            & \yes & \no     & \no               & \yes       & 5\%,10\% train set    & --  \\ 
        \cite{jukić2022}      & 15       & 1e-5           & --            & \yes & \no     & \no               & \no        & --                    & besov early stopping from train results   \\
        \cite{kwak2022}       & 10       & 2e-5           & --            & \yes & \no     & \no               & \yes       & 10\% train set        & --  \\
        \cite{margatina2022}  & 20       & 2e-5           & --            & \yes & \no     & \no               & \yes       & 5\%,10\% train set    & best model from validation results  \\
        \cite{yu2022cold}     & 15       & 1,2,5e-5       & --            & \no  & \yes    & \no               & \yes       & 32 trains set         & best model from validation results  \\
    \bottomrule
\end{tabular}

    }
\end{table*}

\begin{table*}[ht!]
\centering
    \caption{Overview of DAL parameters and query strategies in current work for \gls*{dal} with \gls*{plm}. Type refers to the approach of the respective work. Note that we do not include all query strategies and only focus on the most prominent ones. If a work introduces a novel strategy and we do not include it in the table, it is marked with *. }
    \label{tab:dalregimes}
    \resizebox{1.0\textwidth}{!}{%
        \begin{tabular}{l | c | c c c c c c c c c c| c c c c c}
        \toprule
        \multicolumn{1}{c|}{\multirow{2}{*}{}} &
        \multicolumn{1}{c|}{\multirow{2}{*}{}} &
        \multicolumn{10}{c|}{\textbf{Query strategy}} &
        \multicolumn{5}{c}{\textbf{DAL parameters}} \\
                              &   Type      & Random & Entropy & LC   & Coreset & BADGE   & BALD & CAL & BERT-KM & ALPS & DAL   & \#init  & \#query      & \#budget       & model cold & data cold  \\  \midrule
        \cite{ein-dor2020}    & Comp.       & \yes   & \no     &\yes  & \no     & \yes    & \no  & \no & \no     & \no  & \yes  &  100    & 50           & 350            & \yes          & \no\\  
        \cite{lu2020}         & Comp.       & \yes   & \yes    &\no   & \no     & \no     & \no  & \no & \no     & \no  & \no   &  10     & 10           & 1000           & \yes & \no\\
        \cite{yuan2020}       & Strat.      & \yes   & \yes    & \no  & \no     & \yes    & \no  & \no & \yes    & \yes & \no   &  0      & 100          & 1000           & \yes & \yes\\ 
        \cite{ru2020}         & Strat.$^*$  & \yes   & \yes    &\no   & \no     & \no     & \no  & \no & \no     & \no  & \no   &  0.1\%  & 32           &2,4,6,8,10\%    & \yes & \no\\
        \cite{prabhu2021}     & Comp.       & \yes   & \yes    &\no   & \yes    & \no     & \no  & \no & \no     & \no  & \yes  &  100    & 100          & 2000           & \yes & \no\\
        \cite{margatina2021a} & Train       & \yes   & \yes    &\no   & \no     & \yes    & \no  & \no & \yes    & \yes & \no   &  1\%    & 1\%          & 15\%           & \yes & \no\\
        \cite{margatina2021}  & Strat.      & \yes   & \yes    &\no   & \no     & \yes    & \no  & \yes& \yes    & \yes & \no   &  1\%    & 2\%          & 15\%           & \yes & \no\\
        \cite{tan2021}        & Strat.$^*$  & \yes   & \yes    &\no   & \no     & \no     &\yes  & \no & \no     & \no  & \no   &  20     & 1,5,10,50,100& 500            & \yes & \no\\
        \cite{darcy2022}      & Comp.       & \yes   & \no     &\no   & \yes    & \no     &\yes  & \no &\no      & \no  & \no   &  25     & 12,25,50     & 1000           & \yes & \no\\
        \cite{schröder2022}   & Comp.       & \yes   & \yes    &\yes  & \no     & \no     &\no   &\yes &\no      &\no   & \no   &  25     & 25           & 525            & \yes & \no\\
        \cite{gonsior2022}    & Train.      & \yes   & \yes    &\yes  & \no     & \no     &\no   &\no  &\no      &\no   & \no   &  25     & 25           & 525            & \yes & \no\\
        \cite{seo2022}        & Strat.      & \yes   & \yes    &\yes  & \yes    & \yes    &\yes  &\no  &\no      &\yes  & \yes  &  100    & 100          & 1000           & \yes & \no\\
        \cite{yu2022a}        & Strat.$^*$  & \yes   & \yes    &\no   & \no     & \yes    &\yes  &\yes &\no      &\yes  & \no   &  100    & 100          & 1000           & \yes & \no\\
        \cite{zhang2022a}     & Strat.$^*$  & \yes   & \yes    &\no   & \no     & \yes    &\no   &\yes &\no      &\no   & \no   &  0.1\%  & 50           & 15\%           & \yes & \no\\
        \cite{jukić2022}      & Train       & \yes   & \yes    &\no   & \yes    & \no     &\no   &\no  &\no      &\no   & \yes  &  50     & 50           & 1000,2000      & \yes & \no\\
        \cite{kwak2022}       & Train       & \yes   & \no     &\yes  & \yes    & \yes    &\no   &\no  &\no      &\no   & \no   &  2\%    & 2\%          & 100\%          & \yes & \no\\
        \cite{margatina2022}  & Train       & \yes   & \yes    &\yes  & \no     & \yes    &\yes  &\no  &\yes     &\yes   & \no  &  1\%    & 1\%          & 15\%           & \yes & \no\\
        \cite{yu2022cold}     & Strat.$^*$  & \yes   & \yes    &\no   & \yes    & \no     &\no   &\yes &\yes     &\yes   & \no  &  0      & 64           & 512            & \yes & \yes\\
        glae                  & Bench.      & \yes   & \yes    &\no      & \yes    & \yes    & \no  & \yes&  \no    & \no   & \no  & 100    & 100,25      & 500,1600      & \partially & \no\\
    \bottomrule
\end{tabular}




    }
\end{table*}

\newpage

\section{Training Settings}
\label{appendix:trainin settings}

Tab. \ref{tab:hyperparameters} shows the parameters we use for our experiments to obtain baseline performance values for BERT, DistilBERT, and RoBERTa. This table includes hyperparameters of model training and the parameters of the DAL setting. Due to computational reasons, we limit our ablations studies to BERT. 

\begin{table}[!ht]
\caption{Hyperparameter configurations for model training and parameters of the DAL setting. \underline{Ablations} are highlighted.}
\centering
\resizebox{0.9\textwidth}{!}{%
\begin{tabular}{@{}m{10.5em}|m{7.5em}m{7.5em}m{7.5em}@{}}
\toprule
 \multicolumn{4}{c}
    {\textbf{Model Training}}\\ 
    \hline \\[-0.7em]
\textbf{Parameter} & \textbf{\href{https://huggingface.co/bert-base-cased}{BERT}} & \textbf{\href{https://huggingface.co/distilbert-base-cased}{DistilBERT}} & \textbf{\href{https://huggingface.co/roberta-base}{RoBERTa}}\\\midrule
Model Parameters        & HF Sequence   & HF Sequence         & HF Sequence      \\
Maximum Tokens          & 512           & 512                 & 512              \\
Epochs                  & $5$, $15$, \underline{20}     & 5, 15               & 5, 15            \\
Learning Rate           & 5e-5, \underline{2e-5}    & 5e-5                & 5e-5             \\
Scheduler               & Linear Warmup & Linear Warmup       & Linear Warmup    \\
Warmup Ratio            & 0.05, \underline{0.1}     & 0.05                & 0.05             \\
Batch Size              & 20            & 20                  & 20               \\
AdamW $\epsilon$        & 0.9           & 0.9                 & 0.9              \\
AdamW $\beta_{1}$       & 0.9           & 0.9                 & 0.9              \\
AdamW $\beta_{2}$       & 0.999         & 0.999               & 0.999            \\
AdamW Bias Corr.   &\multicolumn{3}{c}{\yes}            \\
AdamW Weight Decay      & 0.01          & 0.01                & 0.01             \\
Validation Set          &\multicolumn{3}{c}{\no}             \\
\midrule
 \multicolumn{4}{c}
    {\textbf{Deep Active Learning Setting}}\\
    \hline \\[-0.7em]
Model Cold-Start        & True, \underline{False}   & \multicolumn{2}{c}{\yes} \\
Data Cold-Start         &\multicolumn{3}{c}{\no}            \\
\# Initialization       & 100                       & 100                 & 100       \\
\# Query                & 100, \underline{25}       & 100                 & 100       \\
\# Budget               & 500, 1600                 & 500, 1600           & 500, 1600 \\
\# Seeds                & 5                         & 5                   & 5         \\
\# Pool Subset          & 10000, \underline{False}              & 10000               & 10000     \\
\bottomrule
\end{tabular}}
\label{tab:hyperparameters}
\end{table}
\newpage 

\section{Query Strategies}
\label{appendix:query_strats}
We choose the most popular selection strategies for uncertainty-based, diversity-based, and hybrid sampling. Note that we omit query strategies that rely on the epistemic uncertainty of a model. 

\begin{enumerate}
    \item \textbf{Random}: Random sampling is considered the baseline in AL research, where instances are drawn randomly from the unlabeled data set $\mathcal{U}(t)$. The basic goal of al \gls*{dal} query strategies is to achieve better results than random sampling. 
    \item \textbf{Entropy \cite{settles2008}}: Entropy sampling approximated the utility of instances from the unlabeled pool $\mathcal{U}(t)$ based on the entropy \cite{lewis1994} of the predicted probability vector $\mathbf{\hat{p}}$. At cycle iteration $t$, we greedily select the instance with maximum entropy. In our study on \textsc{ActiveGLAE}, we report an average query time of 37 seconds per cycle iteration. 
    \item \textbf{Coreset \cite{sener2018}}: Coreset sampling selects a subset in the unlabeled pool $\mathcal{U}(t)$ that is representative of the entire data set and minimizes the bound between the average loss over any given subset and the remaining data points. At each iteration $t$, $b$ instances $\mathbf{x}^*$ are selected greedily in a batch $\mathcal{B}(t) \subset \mathcal{U}(t)$ to minimize the largest distance between an instance and its nearest center. In our study on \textsc{ActiveGLAE}, we report an average query time of 52 seconds per cycle iteration.
    \item \textbf{CAL \cite{margatina2021}}: Contrastive Active Learning selects instances that are similar in the embedding space, but the predicted probability vector $\mathbf{\hat{p}}$ is strongly contrastive to their neighbors. CAL calculates the similarity with the maximum mean Kullback-Leibler divergence. In our study on \textsc{ActiveGLAE}, we report an average query time of 180 seconds per cycle iteration.
    \item \textbf{BADGE \cite{ash2020a}}: Batch Active Learning by Diverse Gradient Embeddings is a hybrid query strategy that combines the predictive uncertainty and the diversity of instances. BADGE treats the magnitude of the gradients in the model's final output layer as the predictive uncertainty. In our study on \textsc{ActiveGLAE}, we report an average query time of 58 seconds per cycle iteration.
\end{enumerate}

\section{Further Baseline Results}
\label{appendix: baseline results}

\begin{figure}[!ht]
\begin{picture}(300, 270)
    \put(35,260){\includegraphics[width=0.8\textwidth]{figures/resultPlots/legend.pdf}}
    \put(0,187){\includegraphics[width=0.33\textwidth]{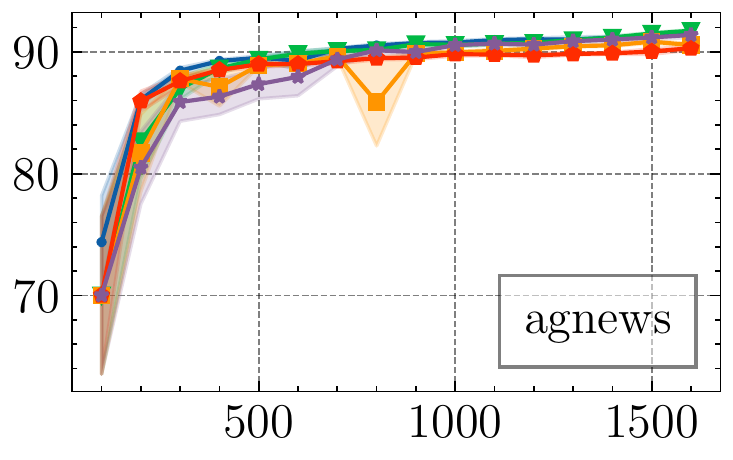}}
    \put(115,187){\includegraphics[width=0.33\textwidth]{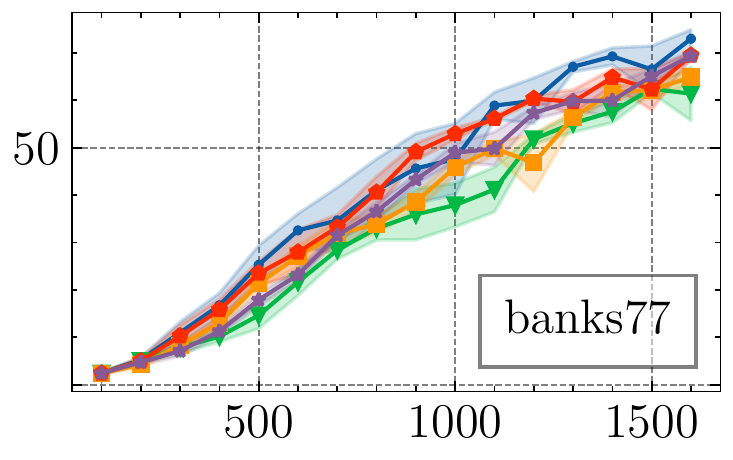}}
    \put(230,187){\includegraphics[width=0.33\textwidth]{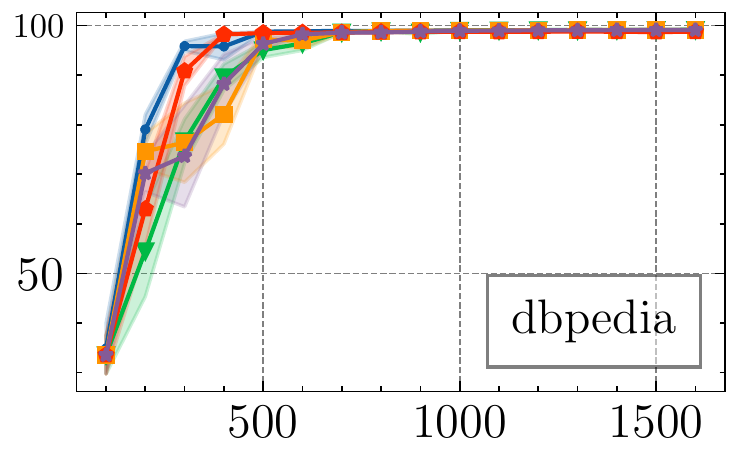}}

    \put(0,125){\includegraphics[width=0.33\textwidth]{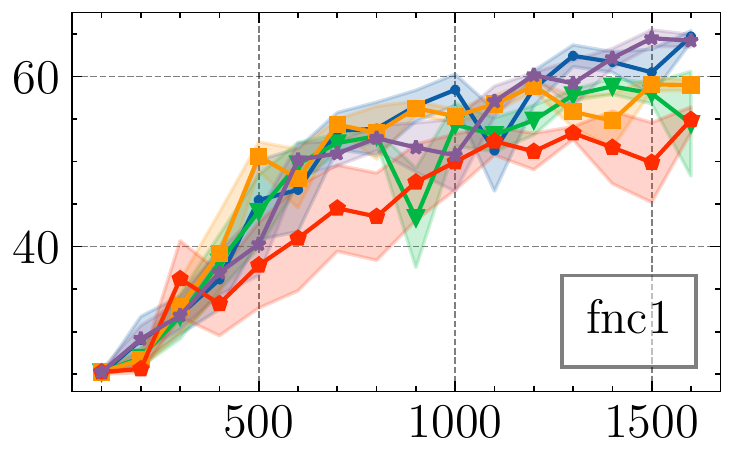}}
    \put(115,125){\includegraphics[width=0.33\textwidth]{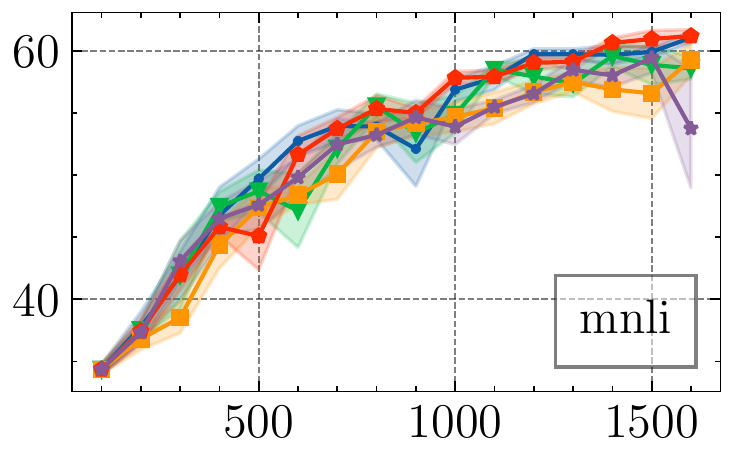}}
    \put(230,125){\includegraphics[width=0.33\textwidth]{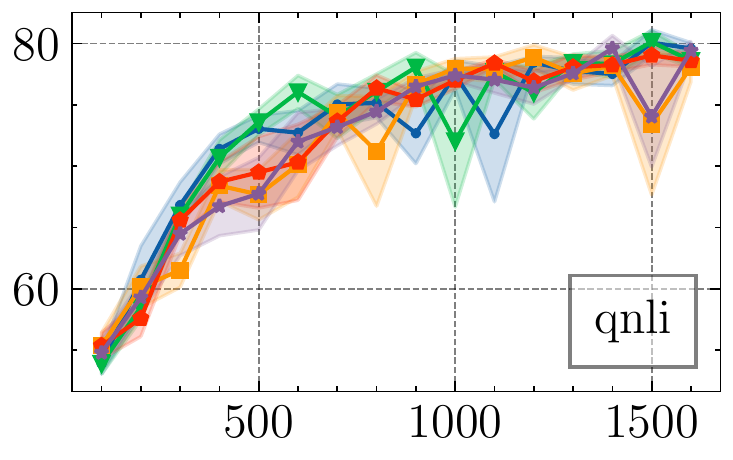}}

    \put(0.5,62){\includegraphics[width=0.327\textwidth]{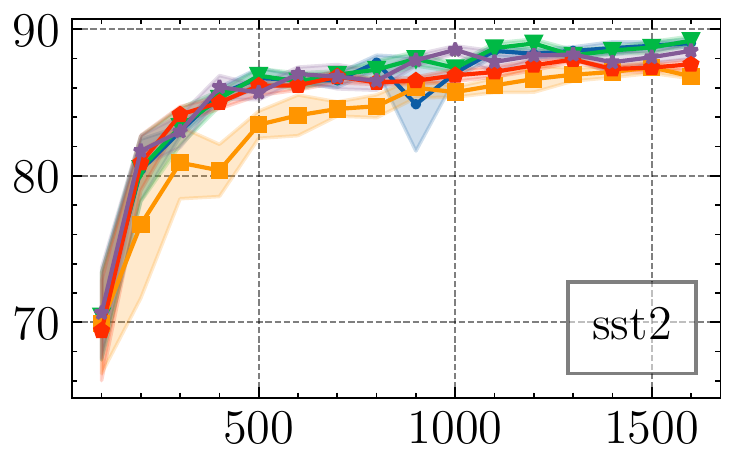}}
    \put(115,62.7){\includegraphics[width=0.33\textwidth]{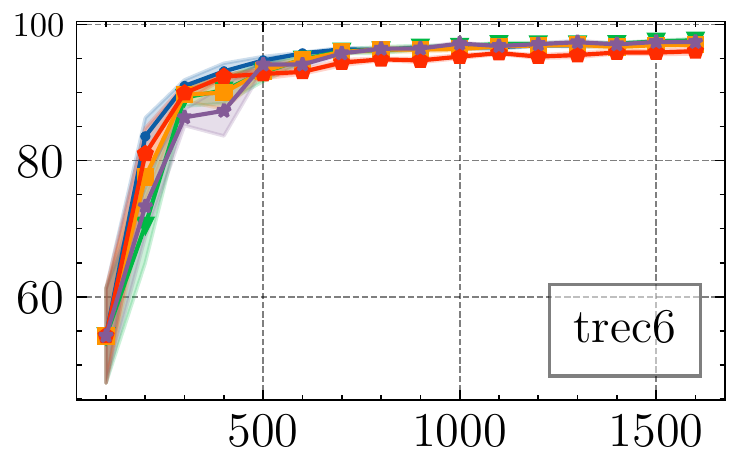}}
    \put(230,63){\includegraphics[width=0.33\textwidth]{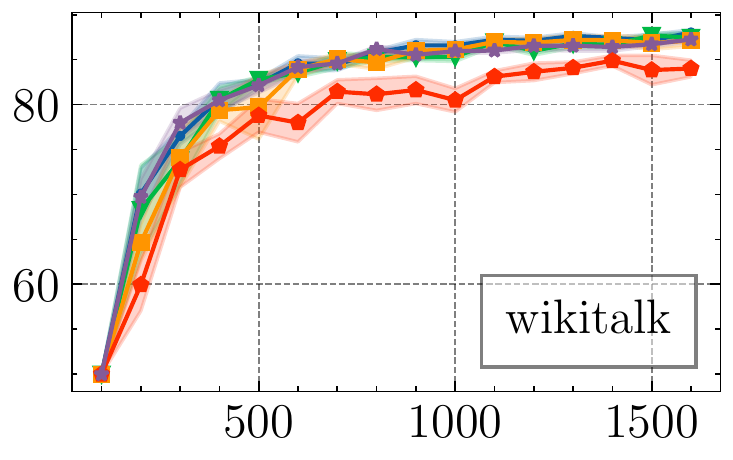}}

    \put(115,0){\includegraphics[width=0.33\textwidth]{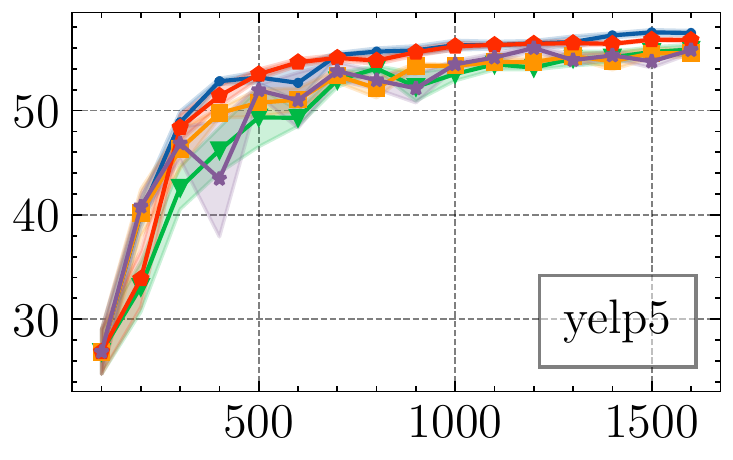}}
\end{picture}
\caption{Baseline \textbf{learning curves for BERT} on \textsc{ActiveGLAE} reporting test accuracy with \protect\st, a learning rate of 5e-5 and 5 repetitions. The shaded area marks the standard error.}
\label{fig:bertbaseline_complete_5ep}
\end{figure}

\begin{figure}[!ht]
\begin{picture}(300, 270)
    \put(35,260){\includegraphics[width=0.8\textwidth]{figures/resultPlots/legend.pdf}}
    \put(0,187){\includegraphics[width=0.33\textwidth]{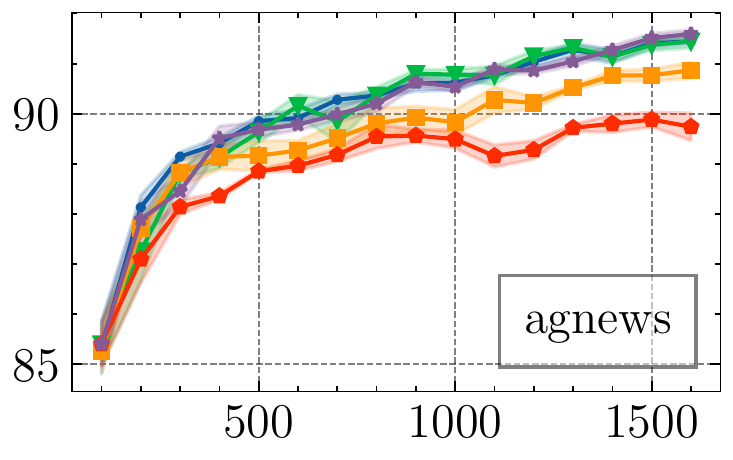}}
    \put(115,187){\includegraphics[width=0.33\textwidth]{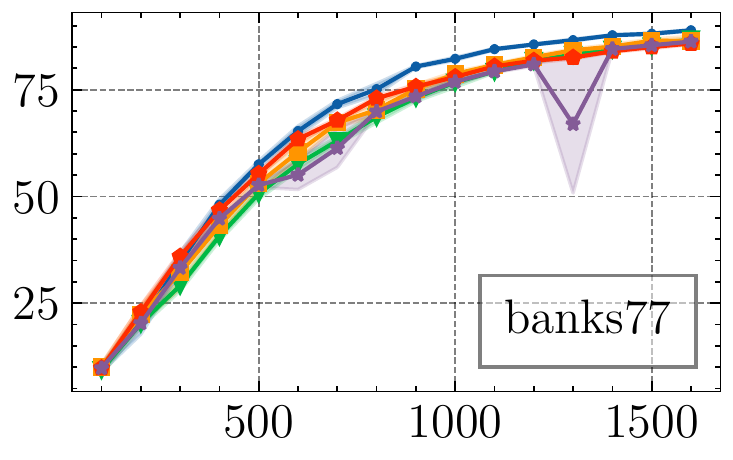}}
    \put(230,187){\includegraphics[width=0.33\textwidth]{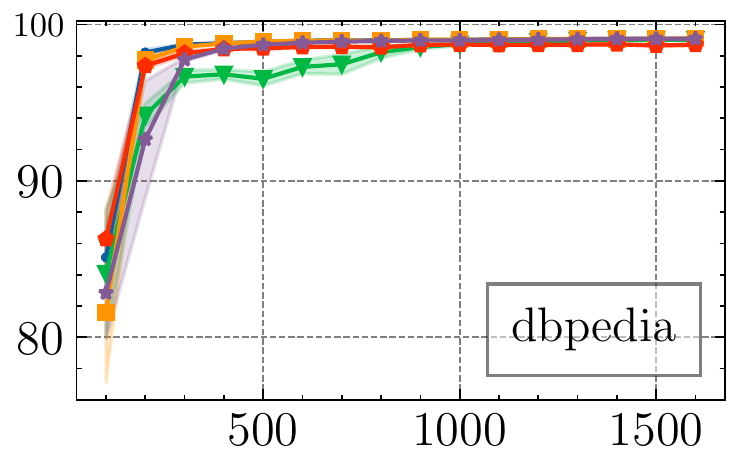}}

    \put(0,125){\includegraphics[width=0.33\textwidth]{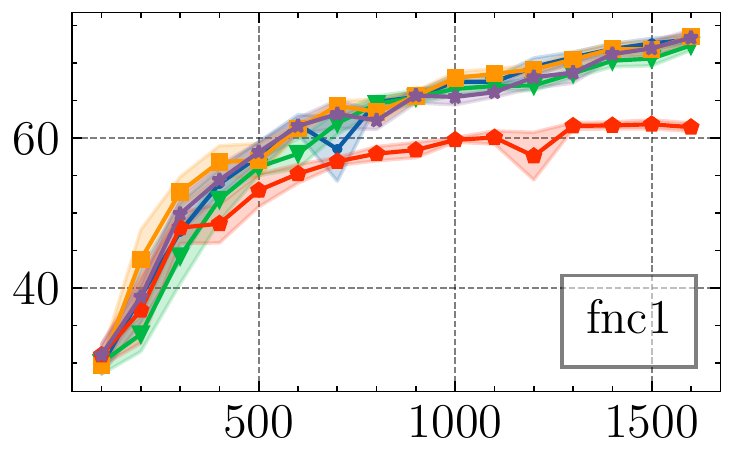}}
    \put(115,125){\includegraphics[width=0.33\textwidth]{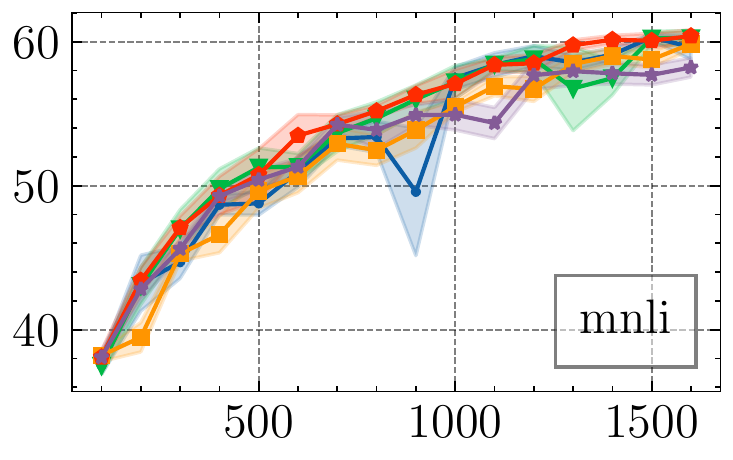}}
    \put(230,125){\includegraphics[width=0.33\textwidth]{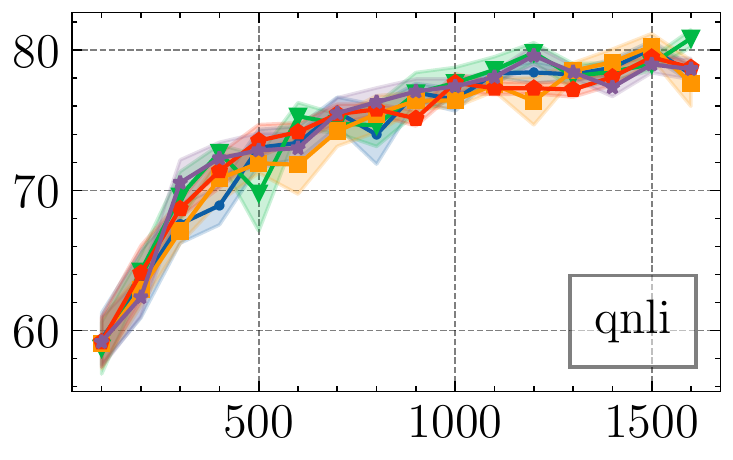}}

    \put(0.5,62){\includegraphics[width=0.327\textwidth]{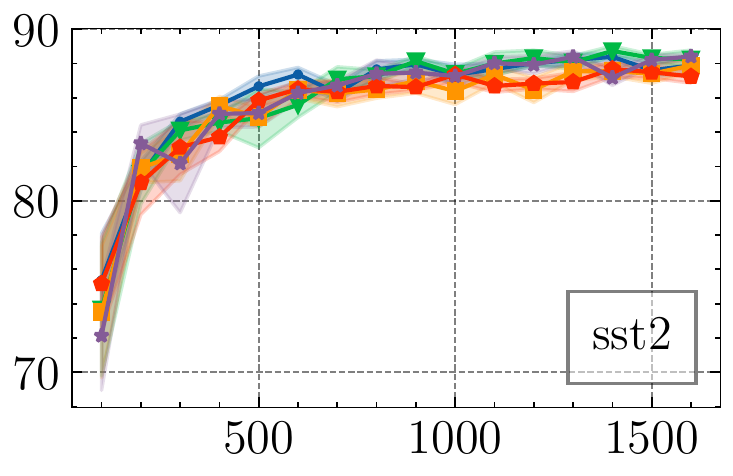}}
    \put(115,63){\includegraphics[width=0.33\textwidth]{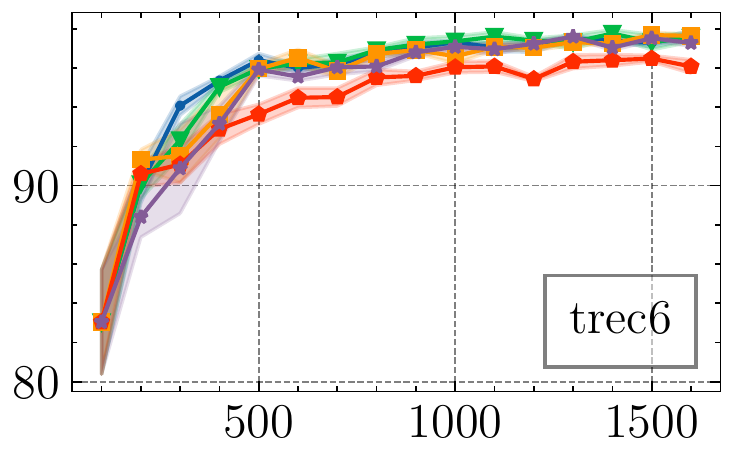}}
    \put(230,63){\includegraphics[width=0.33\textwidth]{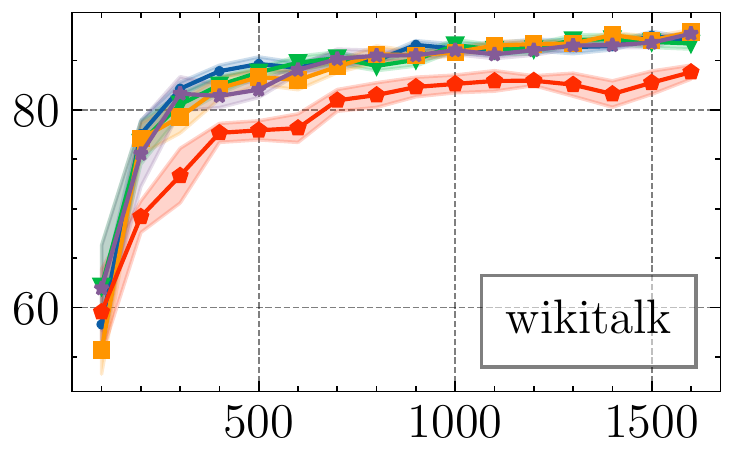}}

    \put(115,0){\includegraphics[width=0.33\textwidth]{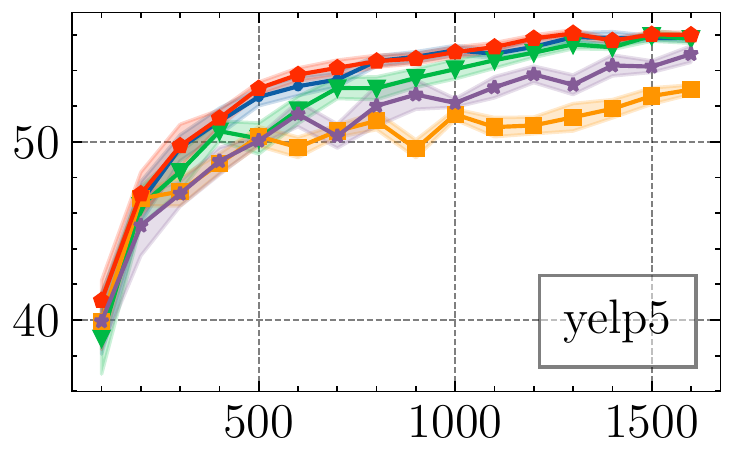}}
\end{picture}
\caption{Baseline \textbf{learning curves for BERT} on \textsc{ActiveGLAE} reporting test accuracy with \protect\lt, learning rate of 5e-5 and 5 repetitions. The shaded area marks the standard deviation.}
\label{fig:bertbaseline_complete_15ep}
\end{figure}

\begin{figure}[!ht]
\begin{picture}(300, 270)
    \put(35,260){\includegraphics[width=0.8\textwidth]{figures/resultPlots/legend.pdf}}
    \put(0,187){\includegraphics[width=0.33\textwidth]{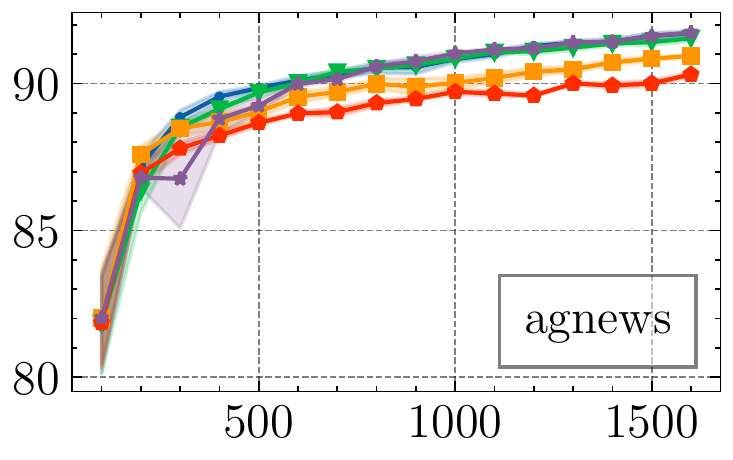}}
    \put(115,187){\includegraphics[width=0.33\textwidth]{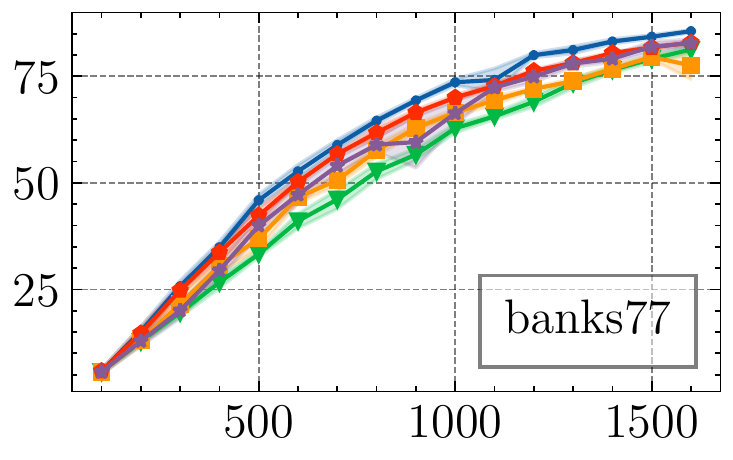}}
    \put(230,187){\includegraphics[width=0.33\textwidth]{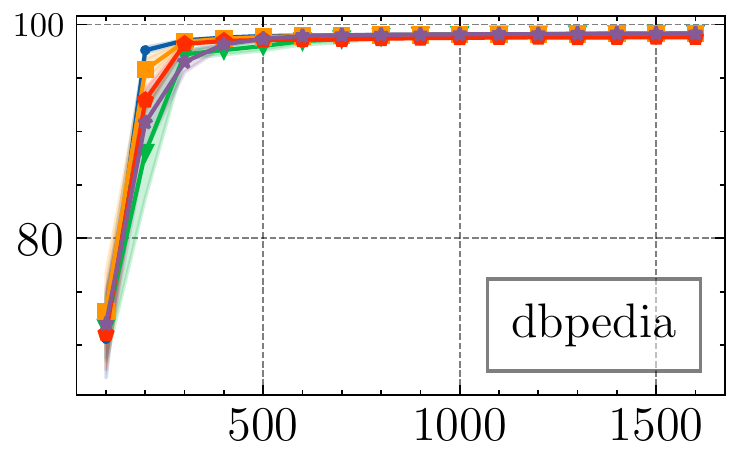}}

    \put(0,125){\includegraphics[width=0.33\textwidth]{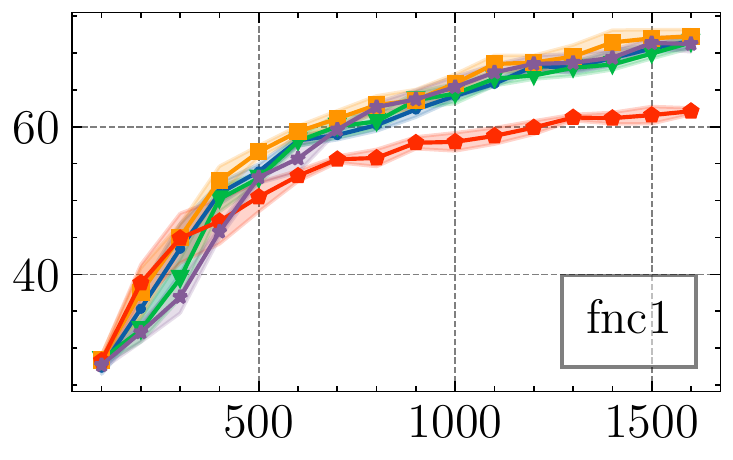}}
    \put(115,125){\includegraphics[width=0.33\textwidth]{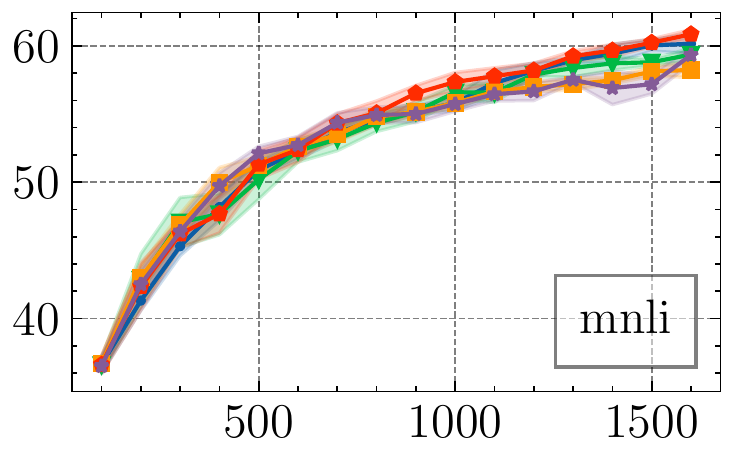}}
    \put(230,125){\includegraphics[width=0.33\textwidth]{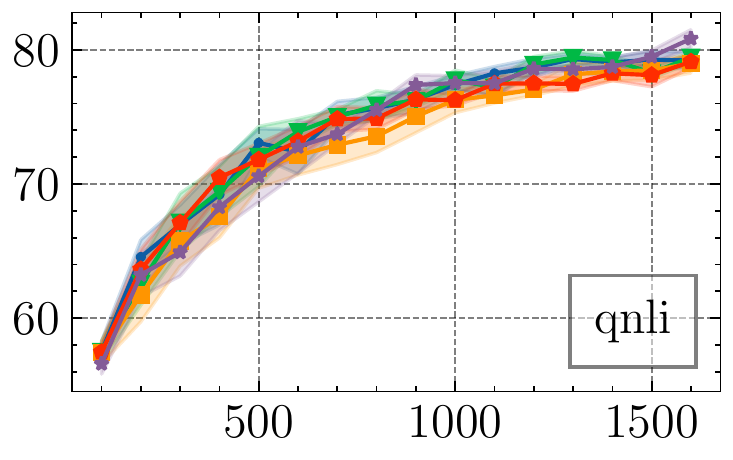}}

    \put(0,61.5){\includegraphics[width=0.33\textwidth]{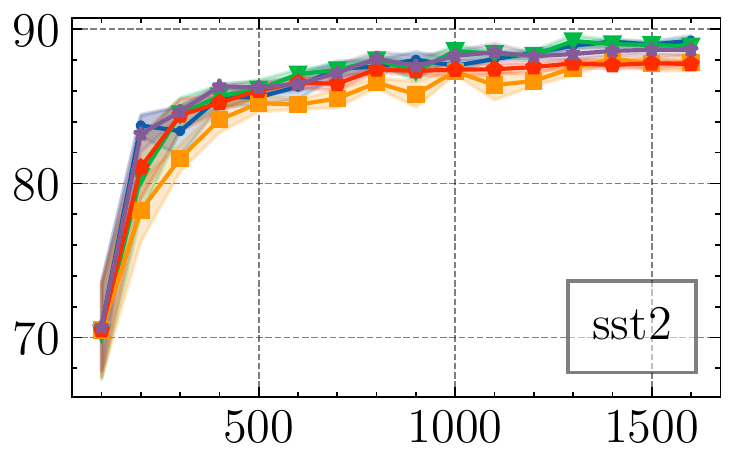}}
    \put(115,63){\includegraphics[width=0.33\textwidth]{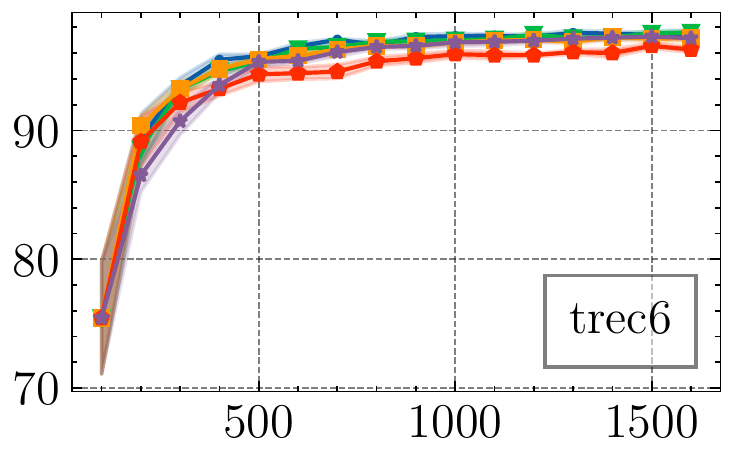}}
    \put(230,63){\includegraphics[width=0.33\textwidth]{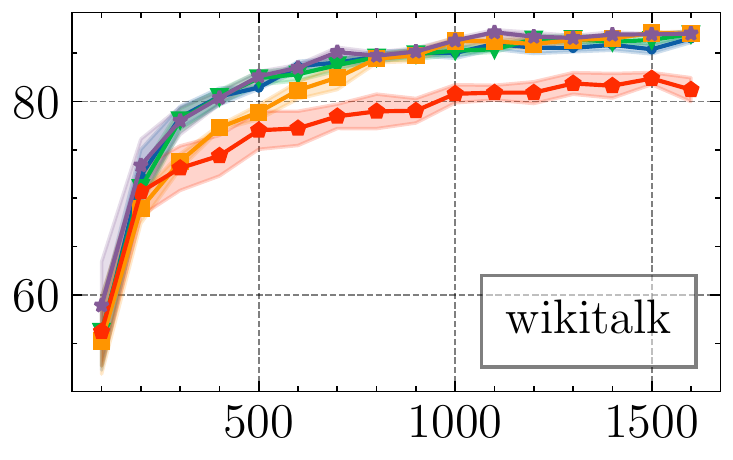}}

    \put(115,0){\includegraphics[width=0.33\textwidth]{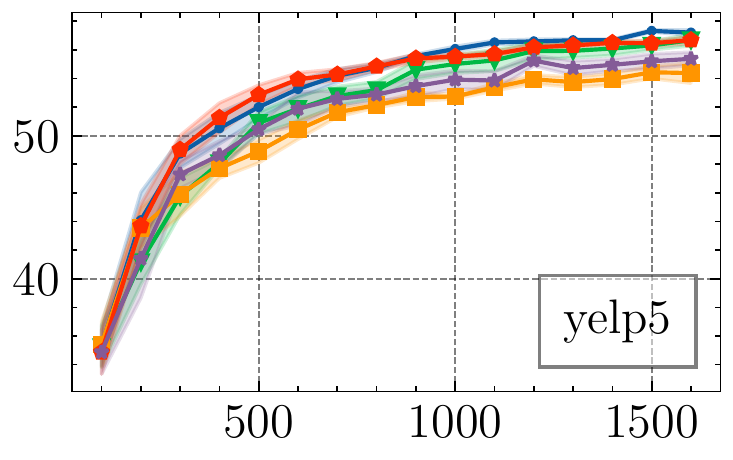}}
\end{picture}
\caption{Baseline \textbf{learning curves for BERT} on \textsc{ActiveGLAE} reporting test accuracy with \protect\ltplus, learning rate of 2e-5 and 5 repetitions. The shaded area marks the standard deviation.}
\label{fig:bertbaseline_complete_20ep}
\end{figure}

\begin{figure}[!ht]
\begin{picture}(300, 260)
    \put(35,260){\includegraphics[width=0.8\textwidth]{figures/resultPlots/legend.pdf}}

    \put(10,180){\includegraphics[width=0.29\textwidth]{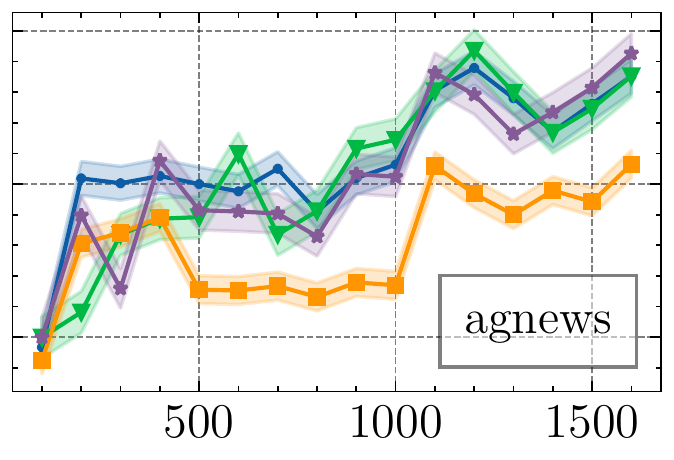}}
    \put(6,196.5){\scriptsize{0}}
    \put(6,219){\scriptsize{1}}
    \put(6,242){\scriptsize{2}}
    
    \put(125,180){\includegraphics[width=0.29\textwidth]{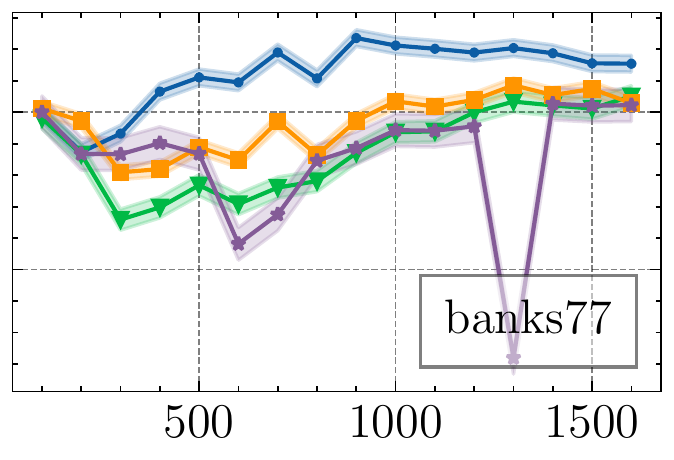}}
    \put(121,230){\scriptsize{0}}
    \put(114.5,206.5){\scriptsize{-10}}
    
    \put(240,180){\includegraphics[width=0.29\textwidth]{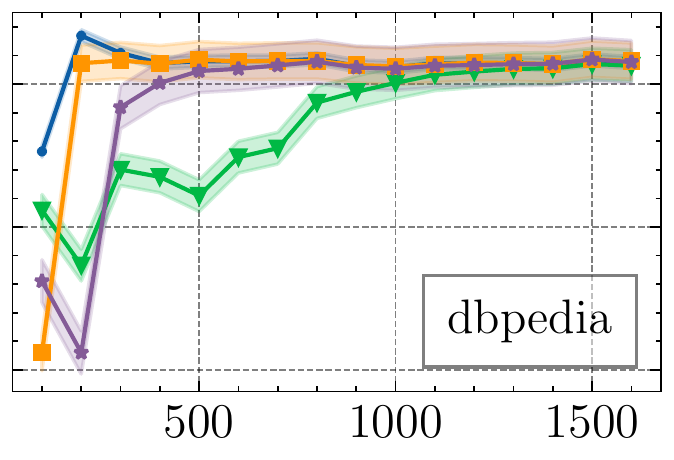}}
    \put(236  ,233){\scriptsize{0}}
    \put(227.5,213){\scriptsize{-2.5}}
    \put(233  ,193){\scriptsize{-5}}

    \put(10,120){\includegraphics[width=0.29\textwidth]{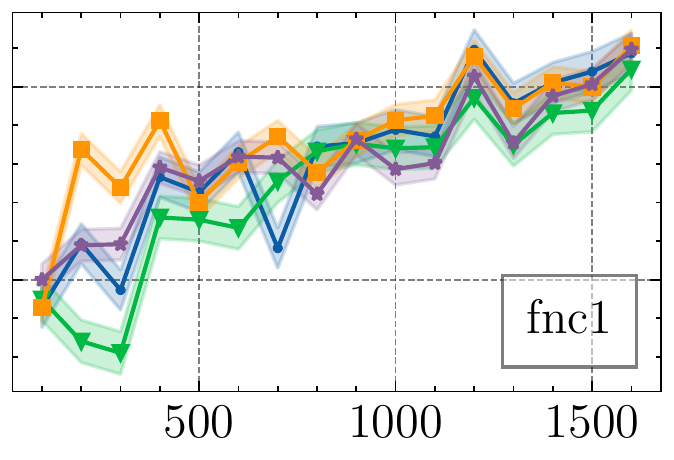}}
    \put(2,173.5){\scriptsize{10}}
    \put(6,145){\scriptsize{0}}
    
    \put(125,120){\includegraphics[width=0.29\textwidth]{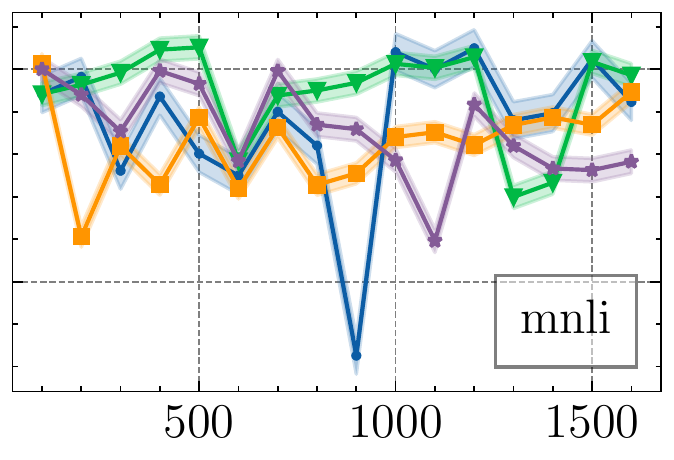}}
    \put(121,175.5){\scriptsize{0}}
    \put(118,145){\scriptsize{-5}}
    
    \put(240,120){\includegraphics[width=0.29\textwidth]{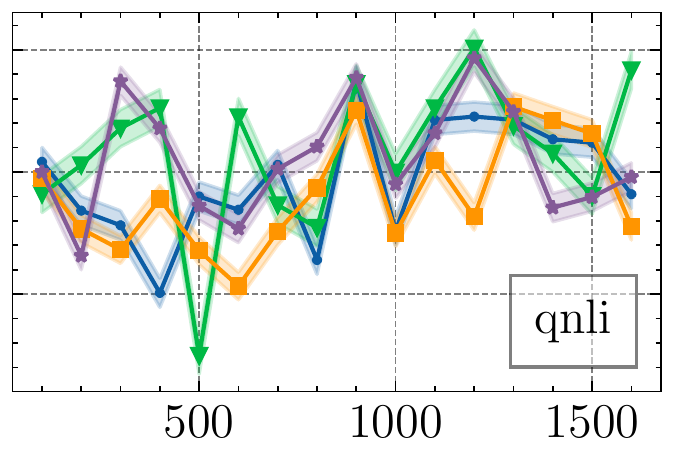}}
    \put(230  ,178.5){\scriptsize{2.5}}
    \put(236  ,161){\scriptsize{0}}
    \put(227.5,142.5){\scriptsize{-2.5}}

    \put(10,60){\includegraphics[width=0.29\textwidth]{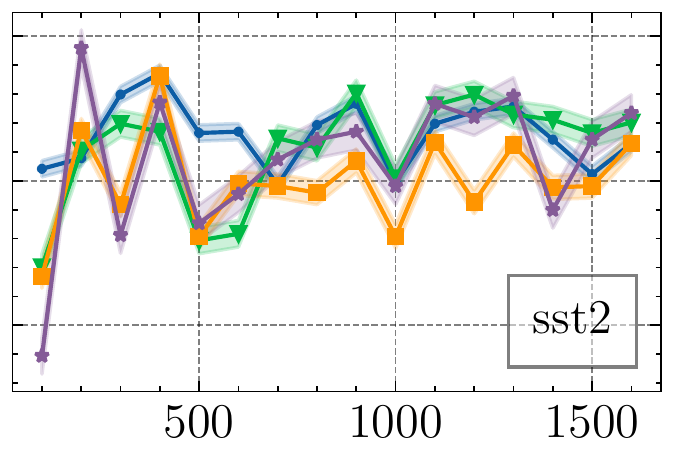}}
    \put(0,120.3){\scriptsize{2.5}}
    \put(6,100){\scriptsize{0}}
    \put(-4,77.5){\scriptsize{-2.5}}

    \put(125,60){\includegraphics[width=0.29\textwidth]{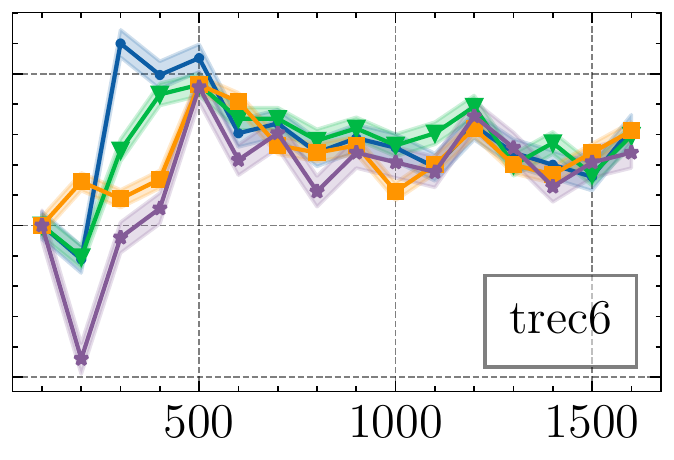}}
    \put(115  ,115){\scriptsize{2.5}}
    \put(121  ,93){\scriptsize{0}}
    \put(112.5,70){\scriptsize{-2.5}}
    
    \put(240,60){\includegraphics[width=0.29\textwidth]{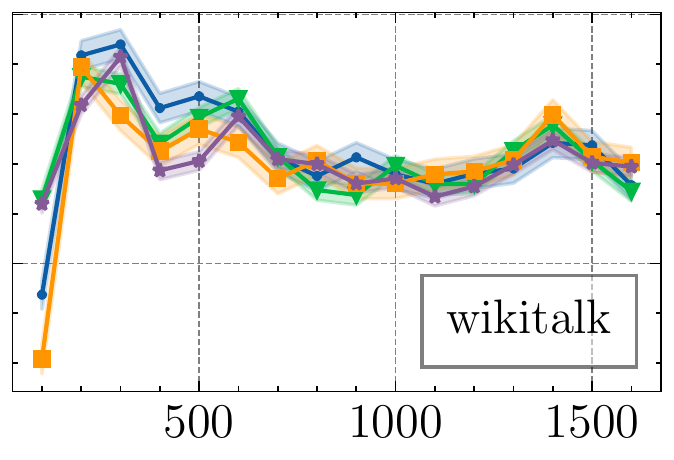}}
    \put(232  ,122.5){\scriptsize{10}}
    \put(236  ,87){\scriptsize{0}}

    \put(10,0){\includegraphics[width=0.29\textwidth]{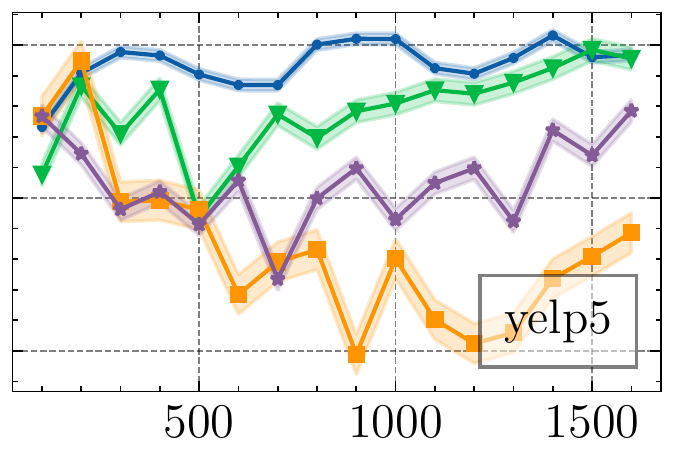}}
    \put(6  ,59.5){\scriptsize{0}}
    \put(-2.5,36.5){\scriptsize{-2.5}}
    \put(3  ,15){\scriptsize{-5}}
    
    \put(114.5,0){\includegraphics[width=0.32\textwidth]{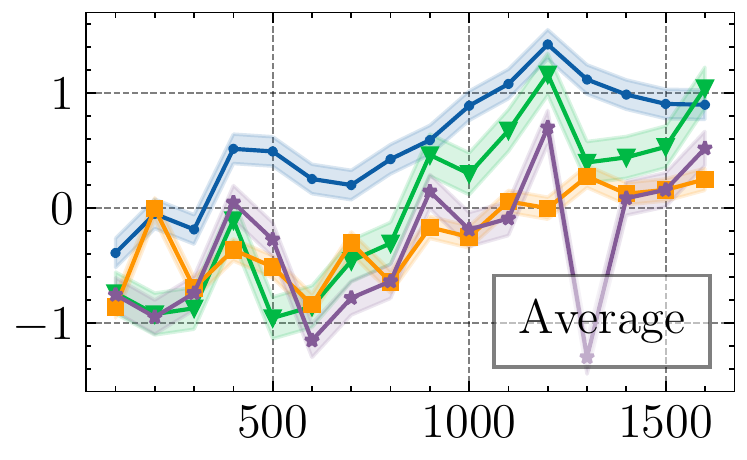}}
    
\end{picture}
\caption{Test accuracy\textbf{ improvement curve based on FAC} relative to random sampling with \protect\lt on \textsc{ActiveGLAE}.}
\label{fig:bert_15_improvement_to_random}
\end{figure}

Tab. \ref{tab:AUC Bert Results} and Tab. \ref{tab:FinalACC Bert Results} provide the more fine-grained results on the respective data sets from \textsc{ActiveGLAE}. We further depict the results by visualizing the learning curves in Figs. \ref{fig:bertbaseline_complete_5ep}, \ref{fig:bertbaseline_complete_15ep} and \ref{fig:bertbaseline_complete_20ep}. Fig. \ref{fig:bert_15_improvement_to_random} shows the improvements in the learning curve of \lt with respect to random sampling. Tab.~\ref{tab:AUC Distilbert Results} and Tab.~\ref{tab:AUC Roberta Results} display the AUC results from RoBERTa and DistilBERT. 



\begin{table*}[ht!]
\centering
    \caption{Baseline \textbf{BERT FAC results} on \textsc{ActiveGLAE} with \protect\st, \protect\lt, \protect\ltplus, two budget sizes (500, 1600) and 5 repetitions ($\pm$standard deviation). \textbf{Best} and \underline{second} best results are highlighted for each dataset. $\color{blue}\uparrow$ and $\color{red}\downarrow$ demonstrate improvements over \textsc{random}.}
    \label{tab:FinalACC Bert Results}
    \resizebox{1.0\textwidth}{!}{%
    \begin{tabular}{l | c | c | c | c | c | c | c | c | c | c | c || c}
        \toprule
        \multicolumn{13}{c} {\cellcolor{yescolor!10}
        \textbf{Low Data Budget: $100+400$}}\\
        \midrule
               &  & \textsc{AGN}     & \textsc{B77}  & \textbf{\textsc{DBP}} & \textsc{FNC1}&  \textsc{MNLI}& \textsc{QNLI}& \textsc{SST2}& \textsc{TREC6}& \textsc{WIK} & \textsc{YELP5} & Average \\ 
        \specialrule{0.2pt}{0.5pt}{0.5pt}
        
        \multirow{3}{*}{\STAB{\rotatebox[origin=c]{90}{\scriptsize{\textbf{random}}}}}  
            & \multirow{1}*{\textsc{st}} & 89.00$\pm{0.2}$    & 27.96$\pm{9.5}$    & 98.58$\pm{0.1}$    & 40.97$\pm{12.3}$   & 51.65$\pm{3.0}$    & 70.32$\pm{6.2}$    & 86.17$\pm{0.6}$    & 93.04$\pm{0.6}$    & 77.98$\pm{4.3}$    & \textbf{54.64}$\pm{0.7}$   & 69.03 $\leftarrow$\\ [0.15em] 
                                 
            & \multirow{1}*{\textsc{lt}} & 88.96$\pm{0.2}$    & \underline{63.42}$\pm{1.2}$    & 98.57$\pm{0.1}$    & 55.23$\pm{2.5}$    & \textbf{53.48}$\pm{2.9}$    & 74.17$\pm{1.3}$    & 86.51$\pm{0.8}$    & 94.48$\pm{1.0}$    & 78.15$\pm{2.9}$    & 53.76$\pm{0.8}$    & 74.67 $\leftarrow$\\  [0.15em]  

            & \multirow{1}*{\textsc{lt}$^+$} & 88.98$\pm{0.1}$    & 50.21$\pm{1.1}$    & 98.58$\pm{0.1}$    & 53.34$\pm{1.5}$    & 52.37$\pm{1.8}$    & 73.18$\pm{2.2}$    & 86.54$\pm{0.7}$    & 94.44$\pm{0.9}$    & 77.21$\pm{3.5}$    & \underline{53.94}$\pm{0.9}$    & 72.88 $\leftarrow$\\  [0.15em]
        \cline{1-13} 
        \midrule
        
        \multirow{3}{*}{\STAB{\rotatebox[origin=cb]{90}{{\scriptsize\textbf{entropy}}}}} 
            & \multirow{1}*{\textsc{st}} & 87.95$\pm{3.1}$    & 23.19$\pm{2.6}$    & 98.24$\pm{0.3}$    & 50.20$\pm{3.9}$    & 49.81$\pm{3.0}$    & 72.00$\pm{4.7}$    & 86.95$\pm{0.9}$    & 94.12$\pm{0.9}$    & 84.17$\pm{1.5}$    & 50.99$\pm{5.3}$    & 69.76 $\color{blue}\uparrow$\color{blue}\scriptsize{0.73}\\ [0.15em]
                                 
            & \multirow{1}*{\textsc{lt}} & 89.78$\pm{0.2}$    & 55.05$\pm{6.9}$    & 98.83$\pm{0.1}$    & \underline{61.61}$\pm{2.1}$    & 51.32$\pm{1.2}$    & 73.01$\pm{2.0}$    & 86.28$\pm{0.9}$    & 95.56$\pm{0.4}$    & 84.05$\pm{0.9}$    & 51.55$\pm{1.4}$    & 74.70 $\color{blue}\uparrow$\color{blue}\scriptsize{0.03}\\  [0.15em]  

            & \multirow{1}*{\textsc{lt}$^+$} & 89.99$\pm{0.2}$    & 47.21$\pm{4.2}$    & 98.92$\pm{0.1}$    & 55.69$\pm{3.7}$    & 52.72$\pm{1.3}$    & 72.77$\pm{3.9}$    & 86.44$\pm{1.0}$    & 95.40$\pm{0.4}$    & 83.48$\pm{0.6}$    & 51.89$\pm{2.4}$    & 73.45 $\color{blue}\uparrow$\color{blue}\scriptsize{0.57}\\  [0.15em]
        \cline{1-13}  
        
        \multirow{3}{*}{\STAB{\rotatebox[origin=cb]{90}{{\scriptsize\textbf{badge}}}}}  
            & \multirow{1}*{\textsc{st}} & 89.29$\pm{1.5}$    & 32.56$\pm{7.0}$    & 98.85$\pm{0.1}$    & 46.67$\pm{9.8}$    & \underline{52.74}$\pm{2.5}$    & 72.73$\pm{3.6}$    & 86.54$\pm{0.6}$    & 95.76$\pm{0.3}$    & \underline{84.58}$\pm{1.7}$    & 52.66$\pm{3.5}$    & 71.24 $\color{blue}\uparrow$\color{blue}\scriptsize{2.21}\\ [0.15em] 
                                 
            & \CC \multirow{1}*{\textsc{lt}} & \CC89.92$\pm{0.3}$    & \CC\textbf{65.31}$\pm{2.3}$    & \CC\textbf{98.97}$\pm{0.0}$    & \CC\textbf{61.83}$\pm{2.6}$    & \CC50.98$\pm{2.2}$    & \CC73.40$\pm{2.3}$    & \CC\textbf{87.36}$\pm{0.8}$    &\CC 96.00$\pm{0.3}$    &\CC 84.25$\pm{1.0}$    & \CC53.11$\pm{1.0}$    &\CC \textbf{76.11} $\color{blue}\uparrow$\color{blue}\scriptsize{1.44}\\  [0.15em]  

            & \multirow{1}*{\textsc{lt}$^+$}& \underline{90.11}$\pm{0.3}$    & 52.70$\pm{3.1}$    & 98.96$\pm{0.0}$    & 58.13$\pm{1.4}$    & 52.26$\pm{1.1}$    & 72.40$\pm{3.3}$    & 86.28$\pm{1.7}$    & \textbf{96.52}$\pm{0.5}$    & 83.63$\pm{1.0}$    & 53.27$\pm{1.2}$    & 74.43 $\color{blue}\uparrow$\color{blue}\scriptsize{1.55}\\  [0.15em]
        \cline{1-13} 
        
        \multirow{3}{*}{\STAB{\rotatebox[origin=cb]{90}{{\scriptsize\textbf{coreset}}}}}  
            & \multirow{1}*{\textsc{st}}& 89.03$\pm{0.4}$    & 27.15$\pm{4.3}$    & 96.99$\pm{3.2}$    & 47.99$\pm{6.9}$    & 48.39$\pm{1.6}$    & 70.13$\pm{5.3}$    & 84.13$\pm{2.8}$    & 94.80$\pm{0.8}$    & 83.93$\pm{1.4}$    & 51.04$\pm{2.2}$    & 69.36 $\color{blue}\uparrow$\color{blue}\scriptsize{0.33}\\ [0.15em] 
                                 
            & \multirow{1}*{\textsc{lt}}& 89.27$\pm{0.4}$    & 60.37$\pm{3.5}$    & \underline{98.96}$\pm{0.0}$    & 61.31$\pm{1.3}$    & 50.67$\pm{2.3}$    & 71.83$\pm{4.2}$    & 86.47$\pm{1.0}$    & \underline{96.52}$\pm{0.9}$    & 83.02$\pm{2.0}$    & 49.69$\pm{1.1}$    & \underline{74.81} $\color{blue}\uparrow$\color{blue}\scriptsize{0.14}\\  [0.15em]  

            & \multirow{1}*{\textsc{lt}$^+$}& 89.55$\pm{0.4}$    & 46.64$\pm{2.7}$    & 98.92$\pm{0.1}$    & 59.32$\pm{1.5}$    & 52.63$\pm{1.2}$    & 72.13$\pm{2.9}$    & 85.11$\pm{0.6}$    & 95.80$\pm{0.4}$    & 81.15$\pm{1.8}$    & 50.42$\pm{1.3}$    & 73.17 $\color{blue}\uparrow$\color{blue}\scriptsize{0.29}\\  [0.15em]
        \cline{1-13} 
        
        \multirow{3}{*}{\STAB{\rotatebox[origin=cb]{90}{{\scriptsize\textbf{cal}}}}}  
            & \multirow{1}*{\textsc{st}} & 89.84$\pm{0.4}$    & 21.68$\pm{5.7}$    & 96.42$\pm{2.8}$    & 49.72$\pm{5.1}$    & 47.13$\pm{5.9}$    & \textbf{76.05}$\pm{2.7}$    & 86.44$\pm{1.1}$    & 94.88$\pm{1.1}$    & 83.53$\pm{0.9}$    & 49.30$\pm{1.6}$    & 69.50 $\color{blue}\uparrow$\color{blue}\scriptsize{0.47}\\ [0.15em]
                                 
            & \multirow{1}*{\textsc{lt}}& \textbf{90.16}$\pm{0.4}$    & 57.58$\pm{2.9}$    & 97.29$\pm{0.9}$    & 57.92$\pm{4.0}$    & 51.32$\pm{1.8}$    & \underline{75.30}$\pm{1.9}$    & 85.60$\pm{1.5}$    & 96.24$\pm{0.4}$    & \textbf{84.76}$\pm{1.3}$    & 51.78$\pm{1.5}$    & 74.79 $\color{blue}\uparrow$\color{blue}\scriptsize{0.12}\\  [0.15em]  

            & \multirow{1}*{\textsc{lt}$^+$} & 90.03$\pm{0.2}$    & 41.06$\pm{3.4}$    & 98.44$\pm{0.7}$    & 58.03$\pm{1.4}$    & 52.28$\pm{1.6}$    & 73.89$\pm{1.9}$    & \underline{87.09}$\pm{0.6}$    & 96.32$\pm{0.4}$    & 82.80$\pm{1.3}$    & 51.89$\pm{1.8}$    & 73.18 $\color{blue}\uparrow$\color{blue}\scriptsize{0.30}\\  [0.15em]
    \midrule
    \multicolumn{13}{c}
    {\cellcolor{yescolor!40}
    {\textbf{High Data Budget: $100+1500$}}}\\
\specialrule{0.2pt}{0.5pt}{0.5pt}
        
        \multirow{3}{*}{\STAB{\rotatebox[origin=c]{90}{\scriptsize{\textbf{random}}}}}  
            & \multirow{1}*{\textsc{st}} & 90.30$\pm{0.3}$    & 69.45$\pm{1.0}$    & 98.77$\pm{0.1}$    & 54.91$\pm{3.0}$    & \textbf{61.19}$\pm{1.0}$    & 78.59$\pm{1.0}$    & 87.61$\pm{0.6}$    & 96.08$\pm{0.8}$    & 84.01$\pm{1.9}$    & 56.77$\pm{0.4}$    & 77.77  $\leftarrow$\\ [0.15em] 
                                 
            & \multirow{1}*{\textsc{lt}}& 89.74$\pm{0.5}$    & 85.82$\pm{0.5}$    & 98.71$\pm{0.1}$    & 61.45$\pm{1.1}$    & 60.39$\pm{0.9}$    & 78.77$\pm{0.7}$    & 87.25$\pm{0.8}$    & 96.08$\pm{0.7}$    & 83.81$\pm{1.4}$    & 55.99$\pm{0.3}$    & 79.80 $\leftarrow$ \\  [0.15em]  

            & \multirow{1}*{\textsc{lt}$^+$}& 90.31$\pm{0.2}$    & 82.90$\pm{0.7}$    & 98.79$\pm{0.1}$    & 62.09$\pm{0.9}$    & 60.85$\pm{0.7}$    & 79.12$\pm{0.9}$    & 87.75$\pm{0.3}$    & 96.28$\pm{0.5}$    & 81.21$\pm{2.5}$    & 56.68$\pm{0.5}$    & 79.60  $\leftarrow$\\  [0.15em]
        \cline{1-13}  
        \midrule
        \multirow{3}{*}{\STAB{\rotatebox[origin=cb]{90}{{\scriptsize\textbf{entropy}}}}} 
            & \multirow{1}*{\textsc{st}} & 91.39$\pm{0.4}$    & 69.32$\pm{2.3}$    & 99.11$\pm{0.0}$    & 64.20$\pm{1.4}$    & 53.75$\pm{9.6}$    & 79.42$\pm{1.5}$    & 88.53$\pm{1.1}$    & 97.44$\pm{0.5}$    & 87.24$\pm{0.6}$    & 55.81$\pm{0.7}$    & 78.62 $\color{blue}\uparrow$\color{blue}\scriptsize{0.85}\\ [0.15em] 
                                 
            & \multirow{1}*{\textsc{lt}}& 91.59$\pm{0.2}$    & 86.27$\pm{0.2}$    & 99.09$\pm{0.0}$    & \underline{73.37}$\pm{2.1}$    & 58.22$\pm{1.3}$    & 78.66$\pm{1.5}$    & 88.42$\pm{0.2}$    & 97.28$\pm{0.3}$    & 87.70$\pm{0.3}$    & 54.92$\pm{0.9}$    & 81.55 $\color{blue}\uparrow$\color{blue}\scriptsize{1.75}\\  [0.15em]  

            & \multirow{1}*{\textsc{lt}$^+$}& \textbf{91.74}$\pm{0.2}$    & 82.79$\pm{1.8}$    & \textbf{99.16}$\pm{0.0}$    & 71.20$\pm{2.2}$    & 59.33$\pm{1.2}$    & \underline{80.83}$\pm{1.3}$    & 88.67$\pm{0.9}$    & 97.16$\pm{0.4}$    & 87.00$\pm{0.6}$    & 55.38$\pm{0.9}$    & 81.33 $\color{blue}\uparrow$\color{blue}\scriptsize{1.73}\\  [0.15em]
        \cline{1-13}  
        
        \multirow{3}{*}{\STAB{\rotatebox[origin=cb]{90}{{\scriptsize\textbf{badge}}}}}  
            & \multirow{1}*{\textsc{st}} & 91.61$\pm{0.2}$    & 72.95$\pm{3.8}$    & 99.12$\pm{0.0}$    & 64.69$\pm{1.4}$    & \underline{61.06}$\pm{1.0}$    & 79.60$\pm{0.9}$    & 89.06$\pm{0.5}$    & 97.56$\pm{0.4}$    & \textbf{88.02}$\pm{0.6}$    & \textbf{57.44}$\pm{0.5}$    & 80.11 $\color{blue}\uparrow$\color{blue}\scriptsize{2.34}\\ [0.15em] 
                                 
            & \multirow{1}*{\textsc{lt}}& 91.44$\pm{0.1}$    & \textbf{88.91}$\pm{0.3}$    & 99.09$\pm{0.0}$    & 73.18$\pm{1.2}$    & 59.63$\pm{1.7}$    & 78.31$\pm{1.1}$    & 87.89$\pm{1.0}$    & \textbf{97.68}$\pm{0.4}$    & 86.96$\pm{0.6}$    & 55.83$\pm{0.4}$    & \underline{81.89} $\color{blue}\uparrow$\color{blue}\scriptsize{2.09}\\  [0.15em]

            & \multirow{1}*{\textsc{lt}$^+$} & 91.71$\pm{0.1}$    & 85.58$\pm{0.7}$    & 99.15$\pm{0.0}$    & 71.58$\pm{1.9}$    & 60.16$\pm{1.4}$    & 79.21$\pm{1.3}$    & \textbf{89.27}$\pm{0.6}$    & 97.52$\pm{0.6}$    & 86.51$\pm{0.8}$    & \underline{57.21}$\pm{0.4}$    & 81.79 $\color{blue}\uparrow$\color{blue}\scriptsize{2.19}\\  [0.15em]
        \cline{1-13} 
        
        \multirow{3}{*}{\STAB{\rotatebox[origin=cb]{90}{{\scriptsize\textbf{coreset}}}}}  
            & \multirow{1}*{\textsc{st}} & 90.56$\pm{0.2}$    & 64.88$\pm{2.6}$    & 99.13$\pm{0.0}$    & 58.97$\pm{1.2}$    & 59.26$\pm{2.5}$    & 78.09$\pm{2.4}$    & 86.79$\pm{0.3}$    & 97.04$\pm{0.4}$    & 87.12$\pm{1.3}$    & 55.52$\pm{0.7}$    & 77.74 $\color{red}\downarrow$\color{red}\scriptsize{0.03}\\ [0.15em] 
                                 
            & \multirow{1}*{\textsc{lt}} & 90.87$\pm{0.3}$    & 86.40$\pm{2.2}$    & 99.11$\pm{0.0}$    & \textbf{73.57}$\pm{1.2}$    & 59.86$\pm{0.8}$    & 77.65$\pm{3.3}$    & 87.89$\pm{0.8}$    & 97.64$\pm{0.1}$    & \underline{87.87}$\pm{0.5}$    & 52.92$\pm{0.5}$    & 81.38 $\color{blue}\uparrow$\color{blue}\scriptsize{1.58}\\  [0.15em]  

            & \multirow{1}*{\textsc{lt}$^+$}& 90.94$\pm{0.1}$    & 77.53$\pm{6.5}$    & \textbf{99.16}$\pm{0.0}$    & 72.23$\pm{1.8}$    & 58.24$\pm{1.1}$    & 78.98$\pm{1.4}$    & 87.87$\pm{1.1}$    & 97.16$\pm{0.3}$    & 87.00$\pm{0.5}$    & 54.38$\pm{1.4}$    & 80.35 $\color{blue}\uparrow$\color{blue}\scriptsize{0.75}\\  [0.15em]
        \cline{1-13} 
        
        \multirow{3}{*}{\STAB{\rotatebox[origin=cb]{90}{{\scriptsize\textbf{cal}}}}}  
            & \multirow{1}*{\textsc{st}} & \textbf{91.74}$\pm{0.2}$    & 61.38$\pm{11.3}$    & \textbf{99.16}$\pm{0.0}$    & 54.44$\pm{12.3}$    & 58.59$\pm{1.7}$    & 78.59$\pm{1.1}$    & \underline{89.22}$\pm{0.6}$    & \underline{97.68}$\pm{0.5}$    & 87.46$\pm{1.0}$    & 55.85$\pm{1.0}$    & 77.41 $\color{red}\downarrow$\color{red}\scriptsize{0.36}\\ [0.15em] 
                                 
            & \CC\multirow{1}*{\textsc{lt}}& 91.44$\pm{0.3}$    & \CC\underline{86.84}$\pm{0.5}$    & \CC99.03$\pm{0.0}$    & \CC72.35$\pm{1.3}$    & \CC60.25$\pm{1.0}$    & \CC\textbf{80.85}$\pm{1.2}$    & \CC88.26$\pm{1.1}$    & \CC97.56$\pm{0.5}$    & \CC86.72$\pm{1.1}$    & \CC55.77$\pm{0.5}$    & \CC\textbf{81.91} $\color{blue}\uparrow$\color{blue}\scriptsize{2.11}\\  [0.15em]  

            & \multirow{1}*{\textsc{lt}$^+$}& 91.54$\pm{0.2}$    & 81.29$\pm{2.0}$    & 99.15$\pm{0.0}$    & 71.39$\pm{1.5}$    & 59.36$\pm{0.9}$    & 79.44$\pm{0.8}$    & 88.90$\pm{0.7}$    & 97.60$\pm{0.3}$    & 86.95$\pm{0.7}$    & 56.68$\pm{0.8}$    & 81.23 $\color{blue}\uparrow$\color{blue}\scriptsize{1.63}\\  [0.15em]
    \bottomrule
\end{tabular}

    }
\end{table*}

\newpage

\begin{table*}[ht!]
\centering
    \caption{Baseline \textbf{DistilBERT AUC results} on \textsc{ActiveGLAE} with three model training approaches (\protect\st, \protect\lt), two budges sizes (500, 1600) and 5 repetitions ($\pm$standard deviation). \textbf{Best} and \underline{second} best results are highlighted for each dataset. $\color{blue}\uparrow$ and $\color{red}\downarrow$ demonstrate improvements over \textsc{random}.}
    \label{tab:AUC Distilbert Results}
    \resizebox{1.0\textwidth}{!}{%
\begin{tabular}{l | c | c | c | c | c | c | c | c | c | c | c || c}
        \toprule
        \multicolumn{13}{c} {\cellcolor{yescolor!10}
        \textbf{Low Data Budget: $100+400$}}\\
        \midrule
               &  & \textsc{AGN}     & \textsc{B77}  & \textbf{\textsc{DBP}} & \textsc{FNC1}&  \textsc{MNLI}& \textsc{QNLI}& \textsc{SST2}& \textsc{TREC6}& \textsc{WIK} & \textsc{YELP5} & Average \\ 
        \specialrule{0.2pt}{0.5pt}{0.5pt}
        
        \multirow{2}*{random} 
            & \multirow{1}*{\textsc{st}} & 85.58$\pm{1.2}$    & 14.85$\pm{1.0}$    & 86.74$\pm{1.9}$    & 25.90$\pm{1.0}$    & 37.07$\pm{1.2}$    & 62.21$\pm{1.4}$    & 79.86$\pm{0.9}$    & 80.39$\pm{2.5}$    & 68.85$\pm{0.6}$    & 44.93$\pm{1.0}$    & 58.64  $\leftarrow$\\ [0.15em] 
                                 
            & \multirow{1}*{\textsc{lt}}& 86.48$\pm{0.6}$    & 34.60$\pm{1.9}$    & 96.42$\pm{0.5}$    & 31.92$\pm{1.0}$    & 39.75$\pm{1.8}$    & 64.57$\pm{1.4}$    & 80.06$\pm{1.0}$    & 89.10$\pm{1.8}$    & 72.63$\pm{2.3}$    & \textbf{47.13}$\pm{0.9}$    & 64.27  $\leftarrow$\\  [0.15em]  
        \cline{1-13}  
        
        \multirow{2}*{entropy} 
            & \multirow{1}*{\textsc{st}}  & 85.98$\pm{0.7}$    & 9.40$\pm{1.7}$    & 84.11$\pm{4.9}$    & 27.24$\pm{1.2}$    & 36.78$\pm{2.3}$    & 61.69$\pm{0.9}$    & 79.12$\pm{2.4}$    & 78.47$\pm{1.1}$    & 69.53$\pm{4.5}$    & 42.90$\pm{1.5}$    & 57.52 $\color{red}\downarrow$\color{red}\scriptsize{1.12} \\
\bottomrule
\end{tabular}

    }
\end{table*}
\begin{table*}[ht!]
\centering
    \caption{Baseline \textbf{DistilBERT FAC results} on \textsc{ActiveGLAE} with three model training approaches (\protect\st, \protect\lt), two budges sizes (500, 1600) and 5 repetitions ($\pm$standard deviation). \textbf{Best} and \underline{second} best results are highlighted for each dataset. $\color{blue}\uparrow$ and $\color{red}\downarrow$ demonstrate improvements over \textsc{random}.}
    \label{tab:FAC Distilbert Results}
    \resizebox{1.0\textwidth}{!}{%
    \begin{tabular}{l | c | c | c | c | c | c | c | c | c | c | c || c}
        \toprule
        \multicolumn{13}{c} {\cellcolor{yescolor!10}
        \textbf{Low Data Budget: $100+400$}}\\
        \midrule
               &  & \textsc{AGN}     & \textsc{B77}  & \textbf{\textsc{DBP}} & \textsc{FNC1}&  \textsc{MNLI}& \textsc{QNLI}& \textsc{SST2}& \textsc{TREC6}& \textsc{WIK} & \textsc{YELP5} & Average \\ 
        \specialrule{0.2pt}{0.5pt}{0.5pt}
        
        \multirow{2}*{random} 
            & \multirow{1}*{\textsc{st}} & 87.87$\pm{0.2}$    & 32.15$\pm{1.5}$    & 98.29$\pm{0.1}$    & 29.48$\pm{2.4}$    & 43.53$\pm{4.4}$    & 71.32$\pm{2.5}$    & 84.36$\pm{0.6}$    & 93.44$\pm{0.9}$    & 78.59$\pm{4.2}$    & \underline{52.32}$\pm{0.6}$    & 67.13  $\leftarrow$\\ [0.15em] 
                                 
            & \multirow{1}*{\textsc{lt}}& 87.61$\pm{0.4}$    & 61.04$\pm{1.3}$    & 98.22$\pm{0.2}$    & 38.94$\pm{6.4}$    & \underline{47.08}$\pm{1.5}$    & 71.17$\pm{1.3}$    & 83.58$\pm{1.1}$    & 93.64$\pm{1.4}$    & 80.21$\pm{2.9}$    & 51.40$\pm{1.2}$    & 71.29 $\leftarrow$\\  [0.15em]  
        \cline{1-13}  
        
        \multirow{2}*{entropy} 
            & \multirow{1}*{\textsc{st}}  & 88.94$\pm{0.3}$    & 23.33$\pm{2.9}$    & 98.20$\pm{0.2}$    & 30.65$\pm{1.9}$    & 43.96$\pm{6.0}$    & 70.81$\pm{0.7}$    & 83.26$\pm{0.5}$    & 93.68$\pm{0.7}$    & 83.18$\pm{1.4}$    & 51.36$\pm{0.5}$    & 66.74 $\color{red}\downarrow$\color{red}\scriptsize{0.39}\\
\bottomrule
\end{tabular}

    }
\end{table*}

\begin{table*}[ht!]
\centering
    \caption{Baseline \textbf{RoBERTa AUC results} on \textsc{ActiveGLAE} with three model training approaches (\protect\st, \protect\lt), two budges sizes (500, 1600) and 5 repetitions ($\pm$standard deviation). \textbf{Best} and \underline{second} best results are highlighted for each dataset. $\color{blue}\uparrow$ and $\color{red}\downarrow$ demonstrate improvements over \textsc{random}.}
    \label{tab:AUC Roberta Results}
    \resizebox{1.0\textwidth}{!}{%
    \begin{tabular}{l | c | c | c | c | c | c | c | c | c | c | c || c}
        \toprule
        \multicolumn{13}{c} {\cellcolor{yescolor!10}
        \textbf{Low Data Budget: $100+400$}}\\
        \midrule
               & & \textsc{AGN}     & \textsc{B77}  & \textbf{\textsc{DBP}} & \textsc{FNC1}&  \textsc{MNLI}& \textsc{QNLI}& \textsc{SST2}& \textsc{TREC6}& \textsc{WIK} & \textsc{YELP5} & Average \\ 
        \specialrule{0.2pt}{0.5pt}{0.5pt}
        
        \multirow{2}*{random}
            & \multirow{1}*{\textsc{st}}  & 87.77$\pm{0.9}$    & 17.24$\pm{1.7}$    & 88.91$\pm{1.8}$    & 29.32$\pm{2.8}$    & 48.71$\pm{1.2}$    & 76.78$\pm{1.2}$    & 81.99$\pm{2.3}$    & 83.47$\pm{2.2}$    & 73.90$\pm{3.2}$    & 53.01$\pm{0.6}$    & 64.11  $\leftarrow$\\ [0.15em] 
                                 
            & \multirow{1}*{\textsc{lt}} & 87.94$\pm{0.3}$    & \underline{46.38}$\pm{0.9}$    & 97.13$\pm{0.4}$    & 46.70$\pm{3.7}$    & \textbf{55.13}$\pm{2.3}$    & 77.68$\pm{1.0}$    & 85.74$\pm{0.6}$    & 92.59$\pm{0.6}$    & 77.57$\pm{1.8}$    & \underline{56.78}$\pm{0.5}$    & 72.36  $\leftarrow$\\  [0.15em]  
        \cline{1-13}  
        
        \multirow{2}*{entropy} 
            & \multirow{1}*{\textsc{st}}  & 87.86$\pm{1.1}$    & 13.75$\pm{2.0}$    & 83.85$\pm{2.0}$    & 38.85$\pm{3.5}$    & 48.48$\pm{3.4}$    & 74.98$\pm{4.2}$    & 81.45$\pm{4.2}$    & 82.97$\pm{2.4}$    & 79.39$\pm{1.3}$    & 48.42$\pm{2.2}$    & 64.00 $\color{red}\downarrow$\color{red}\scriptsize{0.11}\\
\bottomrule
\end{tabular}

    }
\end{table*}
\begin{table*}[ht!]
\centering
    \caption{Baseline \textbf{RoBERTa FAC results} on \textsc{ActiveGLAE} with three model training approaches (\protect\st, \protect\lt), two budges sizes (500, 1600) and 5 repetitions ($\pm$standard deviation). \textbf{Best} and \underline{second} best results are highlighted for each dataset. $\color{blue}\uparrow$ and $\color{red}\downarrow$ demonstrate improvements over \textsc{random}.}
    \label{tab:FAC Roberta Results}
    \resizebox{1.0\textwidth}{!}{%
    %

\caption{Baseline \textbf{learning curves for DistilBERT} on \textsc{ActiveGLAE} reporting test accuracy with \protect\st, learning rate of 5e-5 and 5 repetitions. The shaded area marks the standard error.}
\label{fig:distilbertbaseline_complete_5ep}
\end{figure}

\begin{figure}[!ht]
%
\caption{Baseline \textbf{learning curves for DistilBERT} on \textsc{ActiveGLAE} reporting test accuracy with \protect\lt, learning rate of 5e-5 and 5 seeds. The shaded area marks the standard error.}
\label{fig:distilbertbaseline_complete_15ep}
\end{figure}

\begin{figure}[!ht]
%
\caption{Baseline \textbf{learning curves for RoBERTa} on \textsc{ActiveGLAE} reporting test accuracy with \protect\st, learning rate of 5e-5 and 5 repetitions. The shaded area marks the standard error.}
\label{fig:robertabaseline_complete_5ep}
\end{figure}

\begin{figure}[!ht]
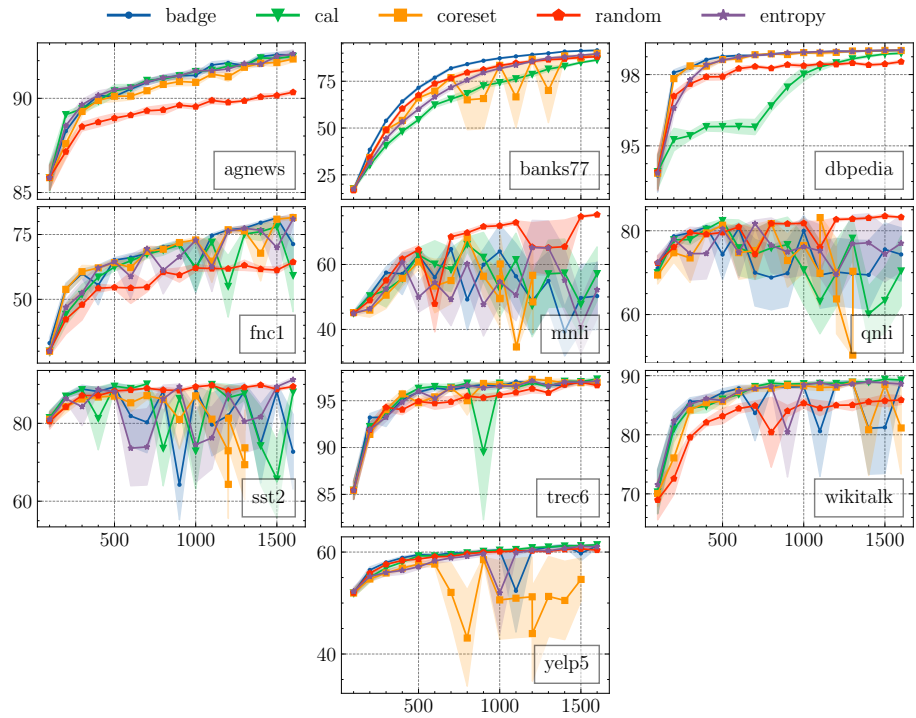

%
\caption{Baseline \textbf{learning curves for RoBERTa} on \textsc{ActiveGLAE} reporting the test accuracy with \protect\lt, a learning rate of 5e-5 and 5 repetitions. The shaded area marks the standard error.}
\label{fig:robertabaseline_complete_15ep}
\end{figure}

\clearpage
\newpage
\section{Ablations}
\label{appendix:ablations}
For our ablations, we chose selected parameters in a \gls*{dal} regime that can be important factors in the performance of query strategies and, accordingly can produce randomness in the experiment design:
\begin{itemize}
    \item \textbf{Query size}: Instead of $n{=}15$ cycle iterations and a query size of $b{=}100$, we investigate a more fine-grained setting with $n{=}60$ and $b{=}25$. While the budget does not change, it leads to more model update steps and a higher runtime.
    \item \textbf{Pool subset}: Instead of using a subset of 10k of the unlabeled pool $\mathcal{U}^*(t)$ at each cycle iteration as our baseline, we query the complete unlabeled pool $\mathcal{U}(t)$. Depending on the available unlabeled pool size, it can increase the experiment runtime significantly (approximately by a factor of 10 per query batch). 
    \item \textbf{Model warm-start}: Instead of re-training the model from scratch at each cycle iteration, we utilize the model parameters $\boldsymbol{\theta}_{t-1}$ from the previous iteration for re-training. 
\end{itemize}

We focus on BERT since it is the most popular one of the investigated \gls*{plm}. Due to computational reasons and experiment runtime, we only use six out of the ten \textsc{ActiveGLAE} data sets for the pool subset ablation. Additionally, we transfer the results based on random sampling for investigating the pool subset and the query size from our baseline results since the results are independent of the respective ablation. 

\begin{table*}[ht!]
\centering
    \caption{Ablation \textbf{BERT AUC} results on \textsc{ActiveGLAE} with \protect\st and \protect\lt, two budget sizes (500, 1600), and 5 repetitions ($\pm$ standard deviation). \textbf{Best} and \underline{second} best results are highlighted for each dataset. $\color{blue}\uparrow$ and $\color{red}\downarrow$ demonstrate improvements over \textsc{random}. This ablation deals with the \textbf{query size}.}
    \label{tab:auc_bert_querybatch}
    \resizebox{1.0\textwidth}{!}{%
\begin{tabular}{l | c | c | c | c | c | c | c | c | c | c | c || c || c}
        \toprule
        \multicolumn{14}{c} {\cellcolor{yescolor!10}
        \textbf{Low Data Budget: $100+400$}}\\
        \midrule
               &  & \textsc{AGN}     & \textsc{B77}  & \textbf{\textsc{DBP}} & \textsc{FNC1}&  \textsc{MNLI}& \textsc{QNLI}& \textsc{SST2}& \textsc{TREC6}& \textsc{WIK} & \textsc{YELP5} & Average & Baseline \\ 
        \specialrule{0.2pt}{0.5pt}{0.5pt}
        
        \multirow{2}*{random} 
            & \multirow{1}*{\textsc{st}} & 85.39$\pm{1.8}$    & 11.02$\pm{3.1}$    & 79.52$\pm{4.2}$    & 31.66$\pm{4.6}$    & 41.24$\pm{0.9}$    & 63.59$\pm{2.0}$    & 81.96$\pm{1.9}$    & 84.19$\pm{3.8}$    & 68.11$\pm{1.4}$    & 43.46$\pm{2.0}$    & 59.01  $\leftarrow$& 59.01 \\ [0.15em] 
                                 
            & \multirow{1}*{\textsc{lt}}& 87.68$\pm{0.3}$    & 34.52$\pm{1.6}$    & 96.59$\pm{0.4}$    & 43.92$\pm{4.8}$    & 46.06$\pm{1.6}$    & 67.64$\pm{1.9}$    & 82.10$\pm{2.3}$    & 90.72$\pm{0.9}$    & 72.23$\pm{3.6}$    & \underline{48.80}$\pm{1.6}$    & 67.02  $\leftarrow$& 67.02\\  [0.15em]  
        \cline{1-14}  
        
        \multirow{2}*{entropy} 
            & \multirow{1}*{\textsc{st}} & 84.29$\pm{2.3}$    & 12.51$\pm{2.0}$    & 81.68$\pm{6.1}$    & 33.94$\pm{3.5}$    & 42.34$\pm{1.0}$    & 65.35$\pm{1.9}$    & 82.35$\pm{1.6}$    & 85.33$\pm{1.4}$    & 76.83$\pm{1.2}$    & 43.90$\pm{0.6}$    & 60.85 $\color{blue}\uparrow$\color{blue}\scriptsize{1.84}  & 58.18\\
\bottomrule
\end{tabular}

    }
\end{table*}

\begin{table*}[ht!]
\centering
    \caption{Ablation \textbf{FAC} results on \textsc{ActiveGLAE} with \protect\st and \protect\lt, two budget sizes (500, 1600), and 5 repetitions ($\pm$ standard deviation). \textbf{Best} and \underline{second} best results are highlighted for each dataset. $\color{blue}\uparrow$ and $\color{red}\downarrow$ demonstrate improvements over \textsc{random}. This ablation deals with the \textbf{query size}.}
    \label{tab:finalacc_bert_querybatch}
    \resizebox{1.0\textwidth}{!}{%
    \begin{tabular}{l | c | c | c | c | c | c | c | c | c | c | c || c || c}
        \toprule
        \multicolumn{14}{c} {\cellcolor{yescolor!10}
        \textbf{Low Data Budget: $100+400$}}\\
        \midrule
               &  & \textsc{AGN}     & \textsc{B77}  & \textbf{\textsc{DBP}} & \textsc{FNC1}&  \textsc{MNLI}& \textsc{QNLI}& \textsc{SST2}& \textsc{TREC6}& \textsc{WIK} & \textsc{YELP5} & Average & Baseline \\ 
        \specialrule{0.2pt}{0.5pt}{0.5pt}
        
        \multirow{2}*{random} 
            & \multirow{1}*{\textsc{st}} & 89.00$\pm{0.2}$    & 27.96$\pm{9.5}$    & 98.58$\pm{0.1}$    & 40.97$\pm{12.3}$   & 51.65$\pm{3.0}$    & 70.32$\pm{6.2}$    & 86.17$\pm{0.6}$    & 93.04$\pm{0.6}$    & 77.98$\pm{4.3}$    & \textbf{54.64}$\pm{0.7}$    & 69.03   $\leftarrow$& 69.03\\ [0.15em] 
                                 
            & \multirow{1}*{\textsc{lt}}& 88.96$\pm{0.2}$    & \underline{63.42}$\pm{1.2}$    & 98.57$\pm{0.1}$    & 55.23$\pm{2.5}$    & \textbf{53.48}$\pm{2.9}$    & \underline{74.17}$\pm{1.3}$    & 86.51$\pm{0.8}$    & 94.48$\pm{1.0}$    & 78.15$\pm{2.9}$    & \underline{53.76}$\pm{0.8}$    & 74.67  $\leftarrow$& 74.67\\  [0.15em]  
        \cline{1-14}  
        
        \multirow{2}*{entropy} 
            & \multirow{1}*{\textsc{st}} & 88.93$\pm{0.5}$    & 26.95$\pm{9.1}$    & 98.57$\pm{0.3}$    & 45.70$\pm{4.9}$    & 50.61$\pm{1.2}$    & 70.72$\pm{6.6}$    & \underline{86.81}$\pm{1.1}$    & 94.68$\pm{0.6}$    & 83.01$\pm{1.5}$    & 49.79$\pm{1.7}$    & 69.58 $\color{blue}\uparrow$\color{blue}\scriptsize{0.55} & 69.76\\
\bottomrule
\end{tabular}

    }
\end{table*}

\begin{table*}[ht!]
\centering
    \caption{Ablation \textbf{BERT AUC} results on a subset of \textsc{ActiveGLAE} with \protect\st and \protect\lt, two budget sizes (500, 1600), and 5 repetitions ($\pm$ standard deviation). \textbf{Best} and \underline{second} best results are highlighted for each dataset. $\color{blue}\uparrow$ and $\color{red}\downarrow$ demonstrate improvements over \textsc{random}. This ablation deals with the \textbf{pool subset}.}
    \label{tab:auc_bert_subset}
    \resizebox{0.85\textwidth}{!}{%
    \begin{tabular}{l | c | c | c | c | c | c | c || c || c}
        \toprule
        \multicolumn{10}{c} {\cellcolor{yescolor!10}
        \textbf{Low Data Budget: $100+400$}}\\
        \midrule
               &  & \textsc{AGN}  & \textsc{FNC1}&  \textsc{MNLI}& \textsc{QNLI}& \textsc{SST2} & \textsc{WIK} & Average & Baseline\\ 
        \specialrule{0.2pt}{0.5pt}{0.5pt}
        
        \multirow{2}*{random} 
            & \multirow{1}*{\textsc{st}}& 85.39$\pm{1.8}$    & 31.66$\pm{4.6}$    & 41.24$\pm{0.9}$    & 63.59$\pm{2.0}$    & 81.96$\pm{1.9}$    & 68.11$\pm{1.4}$    & 61.99  $\leftarrow$& 61.99\\ [0.15em]
                                 
            & \multirow{1}*{\textsc{lt}}& 87.68$\pm{0.3}$    & 43.92$\pm{4.8}$    & 46.06$\pm{1.6}$    & \textbf{67.64}$\pm{1.9}$    & 82.10$\pm{2.3}$    & 72.23$\pm{3.6}$    & 66.60  $\leftarrow$& 66.60\\  [0.15em]  
        \cline{1-10}  
        
        \multirow{2}*{entropy} 
            & \multirow{1}*{\textsc{st}} & 79.84$\pm{6.3}$    & 31.46$\pm{2.5}$    & 41.83$\pm{1.5}$    & 62.48$\pm{2.3}$    & 82.00$\pm{1.3}$    & 72.47$\pm{2.2}$    & 61.68 $\color{red}\downarrow$\color{red}\scriptsize{0.31} & 62.71 \\ [0.15em] 
                                 
            & \multirow{1}*{\textsc{lt}}& 87.64$\pm{0.3}$    & 41.92$\pm{3.2}$    & \underline{46.39}$\pm{1.5}$    & 66.13$\pm{3.5}$    & 82.46$\pm{0.8}$    & 76.99$\pm{3.1}$    & 66.92 $\color{blue}\uparrow$\color{blue}\scriptsize{0.32} & 68.08\\
\bottomrule
\end{tabular}

    }
\end{table*}

\begin{table*}[ht!]
\centering
    \caption{Ablation \textbf{BERT FAC} results on a subset of \textsc{ActiveGLAE} with \protect\st and \protect\lt, two budget sizes (500, 1600), and 5 repetitions ($\pm$ standard deviation). \textbf{Best} and \underline{second} best results are highlighted for each dataset. $\color{blue}\uparrow$ and $\color{red}\downarrow$ demonstrate improvements over \textsc{random}. This ablation deals with the \textbf{pool subset}.}
    \label{tab:finalacc_bert_subset}
    \resizebox{0.8\textwidth}{!}{%
    \begin{tabular}{l | c | c | c | c | c | c | c || c || c}
        \toprule
        \multicolumn{10}{c} {\cellcolor{yescolor!10}
        \textbf{Low Data Budget: $100+400$}}\\
        \midrule
               & & \textsc{AGN}  & \textsc{FNC1}&  \textsc{MNLI}& \textsc{QNLI}& \textsc{SST2} & \textsc{WIK} & Average & Baseline\\ 
        \specialrule{0.2pt}{0.5pt}{0.5pt}
        
        \multirow{2}*{random} 
            & \multirow{1}*{\textsc{st}}& 89.00$\pm{0.2}$    & 40.97$\pm{12.3}$    & 51.65$\pm{3.0}$    & 70.32$\pm{6.2}$    & 86.17$\pm{0.6}$    & 77.98$\pm{4.3}$    & 69.35   $\leftarrow$& 69.35\\ [0.15em] 
                                 
            & \multirow{1}*{\textsc{lt}}& 88.96$\pm{0.2}$    & 55.23$\pm{2.5}$    & \textbf{53.48}$\pm{2.9}$    & 74.17$\pm{1.3}$    & 86.51$\pm{0.8}$    & 78.15$\pm{2.9}$    & 72.75  $\leftarrow$ & 72.75\\  [0.15em]  
        \cline{1-10}  
        
        \multirow{2}*{entropy} 
            & \multirow{1}*{\textsc{st}} & 86.27$\pm{3.3}$    & 44.06$\pm{7.6}$    & 52.04$\pm{2.8}$    & 74.80$\pm{2.4}$    & 84.43$\pm{4.7}$    & 82.74$\pm{1.7}$    & 70.72 $\color{blue}\uparrow$\color{blue}\scriptsize{1.37} & 71.85\\ [0.15em] 
                                 
            & \CC\multirow{1}*{\textsc{lt}}& \CC89.41$\pm{0.2}$    & \CC\underline{58.55}$\pm{3.0}$    & \CC\underline{53.40}$\pm{2.0}$    & \CC74.76$\pm{1.9}$    & \CC\underline{86.58}$\pm{0.5}$    & \CC82.19$\pm{1.7}$    & \CC\textbf{74.15} $\color{blue}\uparrow$\color{blue}\scriptsize{1.40} & \CC\underline{74.34}\\
\bottomrule
\end{tabular}

    }
\end{table*}

\begin{table*}[ht!]
\centering
    \caption{Ablation \textbf{BERT AUC} results on \textsc{ActiveGLAE} with \protect\st and \protect\lt, two budget sizes (500, 1600), and 5 repetitions ($\pm$ standard deviation). \textbf{Best} and \underline{second} best results are highlighted for each dataset. $\color{blue}\uparrow$ and $\color{red}\downarrow$ demonstrate improvements over \textsc{random}. This ablation deals with the \textbf{model warm-start}.}
    \label{tab:auc_bert_warm}
    \resizebox{1.0\textwidth}{!}{
    \begin{tabular}{l | c | c | c | c | c | c | c | c | c | c | c || c || c}
        \toprule
        \multicolumn{14}{c} {\cellcolor{yescolor!10}
        \textbf{Low Data Budget: $100+400$}}\\
        \midrule
               &  & \textsc{AGN}     & \textsc{B77}  & \textbf{\textsc{DBP}} & \textsc{FNC1}&  \textsc{MNLI}& \textsc{QNLI}& \textsc{SST2}& \textsc{TREC6}& \textsc{WIK} & \textsc{YELP5} & Average & Baseline \\ 
        \specialrule{0.2pt}{0.5pt}{0.5pt}
        
        \multirow{2}*{random} 
            & \multirow{1}*{\textsc{st}} & 85.39$\pm{1.8}$    & 11.02$\pm{3.1}$    & 79.52$\pm{4.2}$    & 31.66$\pm{4.6}$    & 41.24$\pm{0.9}$    & 63.59$\pm{2.0}$    & 81.96$\pm{1.9}$    & 84.19$\pm{3.8}$    & 68.11$\pm{1.4}$    & 43.46$\pm{2.0}$    & 59.01  $\leftarrow$& 59.01 \\ [0.15em] 
                                 
            & \multirow{1}*{\textsc{lt}}& 87.68$\pm{0.3}$    & 34.52$\pm{1.6}$    & \textbf{96.59}$\pm{0.4}$    & 43.92$\pm{4.8}$    & 46.06$\pm{1.6}$    & 67.64$\pm{1.9}$    & 82.10$\pm{2.3}$    & 90.72$\pm{0.9}$    & 72.23$\pm{3.6}$    & \textbf{48.80}$\pm{1.6}$    & 67.02  $\leftarrow$& 67.02\\  [0.15em]  
        \cline{1-14}  
        
        \multirow{2}*{entropy} 
            & \multirow{1}*{\textsc{st}} & 85.91$\pm{2.2}$    & 19.68$\pm{2.9}$    & 82.23$\pm{5.2}$    & 40.70$\pm{4.2}$    & 43.78$\pm{0.7}$    & 67.11$\pm{2.2}$    & 82.90$\pm{1.4}$    & 85.86$\pm{3.7}$    & 75.86$\pm{1.7}$    & 45.21$\pm{1.3}$    & 62.92 $\color{blue}\uparrow$\color{blue}\scriptsize{3.91}  & 58.18\\
\bottomrule
\end{tabular}

    }
\end{table*}

\begin{table*}[ht!]
\centering
    \caption{Ablation \textbf{BERT FAC} results on \textsc{ActiveGLAE} with \protect\st and \protect\lt, two budget sizes (500, 1600), and 5 repetitions ($\pm$ standard deviation). \textbf{Best} and \underline{second} best results are highlighted for each dataset. $\color{blue}\uparrow$ and $\color{red}\downarrow$ demonstrate improvements over \textsc{random}. This ablation deals with the \textbf{model warm-start}.}
    \label{tab:finalacc_bert_warm}
    \resizebox{1.0\textwidth}{!}{
    %
\caption{Ablation learning curves for BERT on \textsc{ActiveGLAE} reporting the test accuracy with \protect\st, a learning rate of 5e-5 and 5 seeds. The shaded area marks the standard error. This ablation deals with the \textbf{query size}.}
\label{fig:query_size_complete_5ep}
\end{figure}

\begin{figure}[!ht]
%
\caption{Ablation learning curves for BERT on \textsc{ActiveGLAE} reporting the test accuracy with \protect\lt, a learning rate of 5e-5 and 5 seeds. The shaded area marks the standard error. This ablation deals with the \textbf{query size}.}
\label{fig:query_size_complete_15ep}
\end{figure}

\begin{figure}[!ht]
%
\caption{Ablation learning curves for BERT on \textsc{ActiveGLAE} reporting the test accuracy with \protect\st, a learning rate of 5e-5 and 5 seeds. The shaded area marks the standard error. This ablation deals with the \textbf{pool subset}.}
\label{fig:nosub_complete_5ep}
\end{figure}

\begin{figure}[!ht]
%
\caption{Ablation learning curves for BERT on \textsc{ActiveGLAE} reporting the test accuracy with \protect\lt, a learning rate of 5e-5 and 5 seeds. The shaded area marks the standard error. This ablation deals with the \textbf{pool subset}.}
\label{fig:nosub_complete_15ep}
\end{figure}

\begin{figure}[!ht]
%
\caption{Ablation learning curves for BERT on \textsc{ActiveGLAE} reporting the test accuracy with \protect\st, a learning rate of 5e-5 and 5 seeds. The shaded area marks the standard error. This ablation deals with the \textbf{model warm-start}.}
\label{fig:warm_complete_5ep}
\end{figure}

\begin{figure}[!ht]
%
\caption{Ablation learning curves for BERT on \textsc{ActiveGLAE} reporting the test accuracy with \protect\lt, a learning rate of 5e-5 and 5 seeds. The shaded area marks the standard error. This ablation deals with the \textbf{model warm-start}.}
\label{fig:warm_complete_15ep}
\end{figure}
\end{document}